\pdfoutput=1
\documentclass[11pt]{article}

\usepackage[preprint]{acl}

\usepackage{times}
\usepackage{latexsym}

\usepackage[T1]{fontenc}

\usepackage[utf8]{inputenc}

\usepackage{microtype}

\usepackage{inconsolata}

\usepackage{graphicx}
\graphicspath{{Figures/}{./}}

\usepackage{algorithm}
\usepackage{algorithmic}

\usepackage{listings}

\usepackage{amsmath}
\usepackage{amssymb}
\usepackage{booktabs}
\usepackage{multirow}
\usepackage{colortbl}
\usepackage[most]{tcolorbox}
\usepackage{placeins}

\newtcolorbox{promptbox}[1]{colback=teal!6, colframe=teal!50!black, title=#1,
  title after break={#1 (continued)},
  fonttitle=\bfseries\footnotesize, breakable, enhanced, boxrule=0.5pt,
  left=2pt, right=2pt, top=2pt, bottom=2pt, arc=2pt}
\newtcolorbox{codebox}[1]{colback=blue!4, colframe=blue!45!black, title=#1,
  fonttitle=\bfseries\footnotesize, breakable, enhanced, boxrule=0.5pt,
  left=2pt, right=2pt, top=2pt, bottom=2pt, arc=2pt}

\definecolor{pykw}{HTML}{0033B3}
\definecolor{pycmt}{HTML}{6A737D}
\definecolor{pystr}{HTML}{A31515}
\lstdefinestyle{pyseed}{%
  language=Python, basicstyle=\scriptsize\ttfamily,
  keywordstyle=\color{pykw}\bfseries, commentstyle=\color{pycmt}\itshape,
  stringstyle=\color{pystr}, numbers=none, showstringspaces=false,
  breaklines=true, tabsize=4, columns=fullflexible, keepspaces=true,
  aboveskip=2pt, belowskip=2pt}

\tcbuselibrary{listings}
\newtcblisting{seedbox}[1]{%
  listing only, listing options={style=pyseed},
  colback=blue!4, colframe=blue!45!black, title=#1,
  fonttitle=\bfseries\footnotesize, enhanced, boxrule=0.5pt,
  left=2pt, right=2pt, top=2pt, bottom=2pt, arc=2pt}

\newcommand{\gapvec}{\boldsymbol{\gamma}}
\newcommand{\Xspace}{\mathcal{X}}
\newcommand{\Sspace}{\mathcal{S}}
\newcommand{\Hspace}{\mathcal{H}}
\newcommand{\pol}{\pi_{\theta}}
\newcommand{\polref}{\pi_{\mathrm{ref}}}
\newcommand{\Dtr}{\mathcal{D}_{\mathrm{train}}}
\newcommand{\pool}{\mathcal{P}}
\newcommand{\opset}{\mathcal{O}}
\newcommand{\teacher}{\pi_{\mathrm{T}}}

\definecolor{shadegray}{HTML}{D9D9D9}
\newcommand{\sdpm}[1]{\,\raisebox{0.18ex}{\scalebox{0.72}{$\pm#1$}}}
\newcommand{\val}[2]{#1\sdpm{#2}}
\newcommand{\valb}[2]{\textbf{#1}\sdpm{#2}}
\newcommand{\vals}[2]{\cellcolor{shadegray}#1\sdpm{#2}}
\newcommand{\valsb}[2]{\cellcolor{shadegray}\textbf{#1}\sdpm{#2}}
\newcommand{\rowsep}[1]{\arrayrulecolor{black!25}\cmidrule(lr){1-#1}\arrayrulecolor{black}}

\title{EPOCH: Coverage-Aware Evolution with Experiential Preference Optimization for LLM-based Automated Heuristic Design\thanks{Preprint. Under review.}}

\author{
  \textbf{Oguzhan Gungordu},
  \textbf{Ayse Tursucular},
  \textbf{Siheng Xiong},
  \textbf{Faramarz Fekri}
\\
  Georgia Institute of Technology
\\
  \texttt{\{ogungordu3, atursucular3, sxiong45\}@gatech.edu}
\\
  \texttt{faramarz.fekri@ece.gatech.edu}
}

\begin{document}

\maketitle

\begin{abstract}
LLM-based automated heuristic design (AHD) discovers heuristics for combinatorial optimization problems (COPs), where an LLM generates candidates in an evolutionary search. Existing frameworks freeze the generator or fine-tune it on a scalar fitness signal hiding which instances each heuristic covers. Reflection-based frameworks improve heuristics with natural-language feedback, but it stays in the prompt and never reaches the generator's weights. We propose \textbf{EPOCH}, which makes instance coverage the objective of the evolutionary search and fine-tunes the generator to internalize reflection. Parent selection, operator allocation, and stagnation recovery all optimize how well the heuristics jointly cover the training instances, so the search returns a complementary heuristic ensemble. The generator learns from its own search by experiential preference optimization. A teacher LLM returns a natural-language insight on the generated heuristics, and the generator produces new heuristics with this insight. These improvements join the preference pairs for direct preference optimization (DPO). Since DPO only teaches the generator to rank these improvements, a reflective self-distillation (RSD) term makes it more likely to generate them on its own, and the gains accumulate over the long horizon of the search. Across COPs and open-source LLM backbones, \textbf{EPOCH}'s heuristic ensembles outperform inference-only and fine-tuned baselines and generalize across problem scales.
\end{abstract}

\section{Introduction}
\label{sec:intro}

Many operational decisions, from vehicle routing to scheduling, are COPs~\cite{nguyen2024ieee}. Because these problems are typically NP-hard, practitioners rely on heuristics that trade optimality for speed~\cite{Kirkpatrick1983,Glover1990}. Designing a strong heuristic by hand demands expertise, and a design for one problem rarely carries over to its variants~\cite{Choong2018,pillay2018}. AHD makes the heuristic itself the object of search~\cite{Burke2013}, classically with genetic programming, which cannot express heuristics beyond its human-defined function set~\cite{Langdon2013,gpahd}.

LLM-based AHD replaces that function set with a large language model (LLM) that generates heuristics in the evolutionary loop~\cite{liu2023algorithm}, demonstrated by FunSearch~\cite{funsearch} and EoH~\cite{eoh}. ReEvo~\cite{ye2024reevo} adds reflection~\cite{shinn2023reflexion}, HSEvo~\cite{hsevo} maintains population diversity, MCTS-AHD~\cite{zheng2025montecarlotreesearch} organizes heuristics in a Monte Carlo tree, and EoH-S~\cite{eohs} evolves a complementary set of heuristics. All keep the generator frozen. CALM~\cite{huang2025calmcoevolutionalgorithmslanguage} and EvoTune~\cite{surina2025evotune} instead fine-tune the generator on the heuristics it generates, using group-relative policy optimization (GRPO)~\cite{shao2024deepseekmath} with a manually designed, graded scalar reward and DPO~\cite{rafailov2023dpo} with fitness-ranked pairs, respectively. In both, a heuristic is reduced to a single number, so the update drives the generator toward one best-on-average heuristic, with no reflection on the generated heuristics. Appendix~\ref{app:related} gives an extended review of related work.

\begin{figure*}[t]
    \centering
    \includegraphics[width=\textwidth]{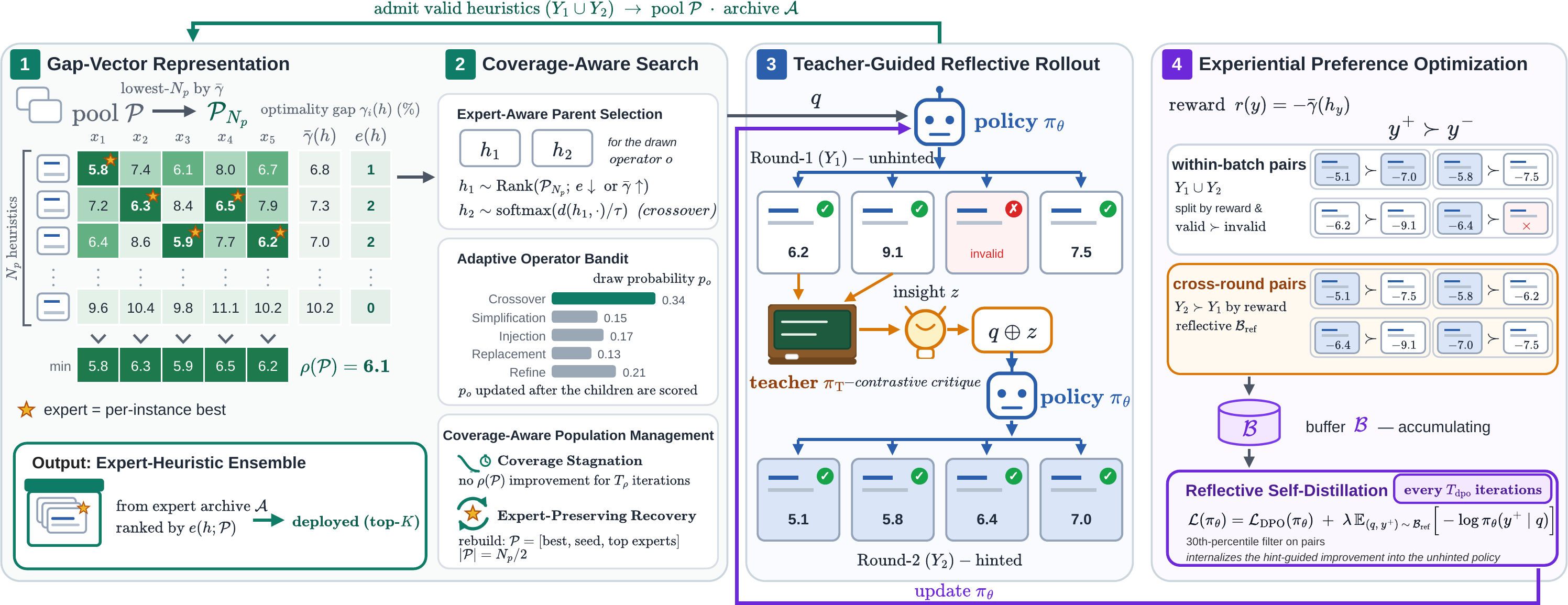}
    \caption{Overview of \textbf{EPOCH}, illustrating one training iteration.
    \textbf{(1)} Heuristics are represented by per-instance gap vectors, whose minima define expert heuristics and the coverage gap $\rho(\pool)$.
    \textbf{(2)} These drive parent selection, operator allocation, and coverage-aware population management.
    \textbf{(3)} The policy $\pol$ samples rollouts from the operator prompt $q$, the teacher $\teacher$ returns an insight $z$ on the best and worst valid ones, and $\pol$ regenerates from $q \oplus z$.
    \textbf{(4)} Responses form within-batch and cross-round preference pairs; periodic DPO bursts add an RSD term internalizing the hint-guided improvement into the unhinted policy.}
    \label{fig:overview}
\end{figure*}

Despite this progress, two limitations remain. First, no heuristic is best on every instance of a distribution~\cite{eohs}, and a scalar fitness hides which instances each heuristic wins, so the coverage of the training instances cannot guide the evolutionary search beyond survivor selection. Second, the natural-language feedback of the inference-only methods stays inside the prompt and never reaches the generator's weights, and the fine-tuned methods employ no reflection. We therefore propose \textbf{E}xperiential \textbf{P}reference \textbf{O}ptimization for \textbf{C}overage-aware \textbf{H}euristic Design (\textbf{EPOCH}), a framework that fine-tunes the generator during a coverage-aware evolutionary search and turns teacher-guided reflection into a training signal.

Our main contributions are as follows. \textbf{(1)} We develop a coverage-aware LLM-based AHD framework that fine-tunes the generator and threads each heuristic's per-instance optimality gaps through parent selection, operator allocation, stagnation recovery, and the selection of the deployed heuristic ensemble, so the search optimizes instance coverage rather than average fitness. \textbf{(2)} We convert teacher-guided reflection into a training signal, where a teacher LLM serves only as a critic, the improvements it guides form preference pairs, and a \emph{reflective self-distillation} term internalizes them so the gains accumulate over the search. \textbf{(3)} We design an adaptive operator bandit that favors the evolutionary operators improving instance coverage, and a coverage-based stagnation recovery with a persistent archive to preserve the best heuristics found on each instance. \textbf{(4)} Experiments across COPs and LLM backbones show that the heuristic ensembles discovered by EPOCH outperform inference-only and fine-tuned LLM-based AHD baselines and generalize across problem scales.

\section{Preliminaries}
\label{sec:prelim}

\paragraph{Automated Heuristic Design Setup.}
We consider a COP with an instance space $\Xspace$ and a solution space $\Sspace$. Each \emph{heuristic} $h \in \Hspace$ is an executable program that constructs a feasible solution, $h:\Xspace\to\Sspace$, and a cost function $f:\Sspace\to\mathbb{R}$ scores it. Performance is assessed on a training set $\Dtr=\{x_1,\dots,x_m\}$. Conventional AHD summarizes a heuristic by the scalar fitness
\begin{equation}
    P(h) \;=\; \frac{1}{m}\sum_{i=1}^{m}\big[-f(h(x_i))\big]
    \label{eq:fitness}
\end{equation}
and, under an evaluation budget, returns the single best-on-average heuristic $h^\star = \operatorname{arg\,max}_{h\in\Hspace} P(h)$. EPOCH instead searches for a set of complementary heuristics covering $\Dtr$ and retains each heuristic's full per-instance performance profile (Section~\ref{sec:method-primitive}) rather than the scalar of Eq.~\eqref{eq:fitness}.

\paragraph{Direct Preference Optimization.}
EPOCH updates the generator with DPO~\cite{rafailov2023dpo}, which trains a policy $\pol$ on triplets of a prompt $q$, a preferred response $y^+$, and a dispreferred response $y^-$, relative to a reference policy $\polref$. With the policy-side \emph{implicit reward} $\hat r_\theta(y)=\beta\log\!\big(\pol(y\mid q)/\polref(y\mid q)\big)$, the loss
\begin{equation}
    \mbox{\small$\displaystyle
    \begin{aligned}
        &\mathcal{L}_{\mathrm{DPO}}\big(\pol\big) = \\
        &\quad {-}\,\mathbb{E}_{(q,y^+,y^-)}\Big[\log\sigma\big(\hat r_\theta(y^+) - \hat r_\theta(y^-)\big)\Big]
    \end{aligned}$}
    \label{eq:dpo}
\end{equation}
widens the \emph{margin} $\hat r_\theta(y^+)-\hat r_\theta(y^-)$, where $\sigma$ is the logistic function and $\beta$ regularizes toward $\polref$. As the evolving policy itself generates our preference data, DPO runs in periodic bursts.

\section{Methodology}
\label{sec:method}

EPOCH augments an LLM-based evolutionary loop with two coupled ideas. The search maintains a heuristic pool $\pool$, initialized from a seed heuristic; its $N_p{=}10$ heuristics with the lowest average gap form the reproduction population $\pool_{N_p}$. Each iteration draws an operator, builds a prompt from parents selected from this population, samples candidates from the policy $\pol$, and admits the valid ones into $\pool$. A \emph{per-instance gap-vector representation} makes the search coverage-aware, and an \emph{experiential} preference-optimization update, fed by teacher-guided reflection and an RSD term that internalizes its gains, fine-tunes the generator on its own evolutionary search. Sections~\ref{sec:method-primitive}--\ref{sec:method-dpo} present the representation, parent selection, the evolutionary operators, the operator bandit, coverage stagnation and recovery, the reflective rollout, and experiential preference optimization, respectively. Figure~\ref{fig:overview} illustrates one training iteration and Algorithm~\ref{alg:main} gives its pseudocode, with the complete pseudocode of its subalgorithms in Appendix~\ref{app:algorithms}.

\subsection{Per-Instance Gap Vectors, Experts, and Coverage}
\label{sec:method-primitive}
Let $f^\star_i$ denote a reference optimum for training instance $x_i$, e.g., an optimal tour length for the Traveling Salesman Problem (TSP) or the $L_1$ lower bound on bin count for online Bin Packing (OBP). The \emph{per-instance optimality gap} of $h$ on $x_i$ is
\begin{equation}
    \gamma_i(h) \;=\; \frac{f(h(x_i)) - f^\star_i}{\lvert f^\star_i\rvert},
    \label{eq:gap}
\end{equation}
which is $0$ when $h$ is optimal on $x_i$ and grows with sub-optimality. We represent $h$ by its \emph{gap vector}
\begin{equation}
    \mbox{\small$\displaystyle
    \begin{aligned}
        \gapvec(h) &= \big(\gamma_1(h),\dots,\gamma_m(h)\big)\in\mathbb{R}^{m},\\
        \bar\gamma(h) &= \tfrac{1}{m}\textstyle\sum_{i}\gamma_i(h),
    \end{aligned}$}
    \label{eq:gapvec}
\end{equation}
where $\bar\gamma(h)$ is the average gap over the $m$ instances.

Given the current pool of heuristics $\pool$, define the pool's \emph{best per-instance gap} as $\gamma^\star_i(\pool)=\min_{h\in\pool}\gamma_i(h)$, the lowest gap any pool member attains on instance $x_i$. The heuristic achieving this minimum is the \emph{expert heuristic} for instance $x_i$, and the \emph{expert count}
\begin{equation}
    e(h;\pool)=\big\lvert\{\,i : h=\operatorname*{arg\,min}\nolimits_{h'\in\pool}\gamma_i(h')\,\}\big\rvert
    \label{eq:expert}
\end{equation}
counts the instances on which $h$ is uniquely best in the pool; a heuristic with $e(h;\pool)>0$ is an expert heuristic. The quality of the pool \emph{as a portfolio} is captured by the \emph{coverage gap}
\begin{equation}
    \mbox{\small$\displaystyle
    \rho(\pool) \;=\; \frac{1}{m}\sum_{i=1}^{m}\gamma^\star_i(\pool) \;=\; \frac{1}{m}\sum_{i=1}^{m}\min_{h\in\pool}\gamma_i(h),$}
    \label{eq:rho}
\end{equation}
obtained by routing each instance to its best heuristic in $\pool$. Unlike the single-best average gap $\min_{h}\bar\gamma(h)$, the coverage gap $\rho(\pool)$ improves whenever any heuristic becomes a better expert on any instance, even while the single best heuristic is unchanged. This set-level objective has previously guided survivor selection within an inference-only search~\cite{eohs}. EPOCH instead makes it the control signal of the entire search. Finally, the \emph{complementarity} of two heuristics is the $\ell_1$ distance between their gap vectors~\cite{eohs},
\begin{equation}
    \mbox{\small$\displaystyle
    \begin{aligned}
        d(h,h') &= \lVert\gapvec(h)-\gapvec(h')\rVert_1\\
        &= \sum_{i=1}^{m}\lvert\gamma_i(h)-\gamma_i(h')\rvert,
    \end{aligned}$}
    \label{eq:compl}
\end{equation}
which is large when two heuristics behave differently across instances.

\subsection{Expert-Aware Parent Selection}
\label{sec:method-parents}
The evolutionary operator prompts are seeded with parent heuristics drawn by rank-based sampling from the reproduction population $\pool_{N_p}$, the $N_p$ heuristics of $\pool$ with the lowest average gap; expert counts in this section are computed within that subset. A draw $h \sim \mathrm{Rank}(\pool_{N_p};\zeta)$ selects $h$ with probability proportional to $1/\mathrm{rank}$ under the ordering key $\zeta$, with $\downarrow$ and $\uparrow$ denoting descending and ascending order. The single-parent mutation operators draw
\begin{equation}
    \mbox{\small$\displaystyle
    h_1 \sim
    \begin{cases}
        \mathrm{Rank}\big(\pool_{N_p};\ e(\cdot;\pool_{N_p})\!\downarrow\big) & \text{w.p.\ } p_{\mathrm{exp}}{=}0.5,\\[2pt]
        \mathrm{Rank}\big(\pool_{N_p};\ \bar\gamma(\cdot)\!\uparrow\big) & \text{otherwise,}
    \end{cases}$}
    \label{eq:sel-single}
\end{equation}
where the expert-count branch, taken with probability $p_{\mathrm{exp}}$, recruits expert heuristics for under-covered instances, and the average-gap branch refines strong all-rounders. The crossover operator draws $h_1 \sim \mathrm{Rank}\big(\pool_{N_p};\bar\gamma(\cdot)\!\uparrow\big)$ and
\begin{equation}
    \mbox{\small$\displaystyle
    h_2 \sim
    \begin{cases}
        \mathrm{softmax}\big(d(h_1,\cdot)/\tau\big),\ \ \tau{=}0.5 & \text{w.p.\ } p_{\mathrm{comp}}{=}0.5,\\[2pt]
        \mathrm{Rank}\big(\pool_{N_p};\ \bar\gamma(\cdot)\!\uparrow\big) & \text{otherwise,}
    \end{cases}$}
    \label{eq:sel-cross}
\end{equation}
where the complementarity branch, taken with probability $p_{\mathrm{comp}}$, pairs the first parent with its most complementary counterpart under Eq.~\eqref{eq:compl}, and the average-gap branch pairs it with a strong all-rounder. The branch taken in Eq.~\eqref{eq:sel-cross} also determines the crossover variant of Section~\ref{sec:method-ops}. Algorithm~\ref{alg:parents} implements this as \textsc{SelectParents}.

\subsection{Evolutionary Operators}
\label{sec:method-ops}
EPOCH evolves heuristics with five evolutionary operators, a two-parent crossover operator and four single-parent mutation operators, all sharing a common prompt skeleton that presents each parent's design idea, code, and average gap; the full templates and their sources are given in Appendix~\ref{app:prompts}.
\begin{itemize}
    \item \textbf{Crossover} recombines two parents, with the variant set by the branch of Eq.~\eqref{eq:sel-cross}:
    \begin{itemize}
        \item \emph{Exploitative crossover}, used when the second parent was chosen by average-gap rank, asks for a heuristic that outperforms both parents on the same instances.
        \item \emph{Complementary crossover}, used when the second parent was chosen by complementarity, asks for a heuristic that unites the parents' strengths across all instances either of them already handles well.
    \end{itemize}
    \item \textbf{Simplification} takes one parent and distills it into a more concise, elegant version, countering the build-up of redundant components.
    \item \textbf{Injection} takes one parent and adds a novel component to it while preserving its core mechanism.
    \item \textbf{Replacement} takes one parent, locates a component that scores competing choices nearly equally, and rewrites it to differentiate them.
    \item \textbf{Refine} takes one parent, identifies its single most important weakness, and rewrites it to fix that weakness while preserving the parts that already work well.
\end{itemize}

\subsection{Adaptive Operator Bandit}
\label{sec:method-bandit}
EPOCH allocates operator probabilities adaptively, favoring operators that advance instance coverage. For each operator $o\in\opset$ we keep a quality estimate $Q_o\in[0,1]$, an exponential moving average of a per-draw reward. After $o$ is drawn and its children are scored, let $g_o$ be the fraction of the $m$ instances on which any of its children improves the pool's best per-instance gap measured before admission, $\nu_p$ the fraction of valid children whose average gap beats the best parent's, and $\nu_v$ the valid fraction. The reward combines these as
\begin{equation}
    R_o \;=\; w_{e}\,g_o + w_{p}\,\nu_{p} + w_{v}\,\nu_{v},
    \label{eq:bandit-reward}
\end{equation}
with weights $(w_e,w_p,w_v)=(0.6,0.25,0.15)$, so the coverage gain $g_o$ dominates. The quality is updated as $Q_o\leftarrow(1-\eta)Q_o+\eta R_o$ with $\eta{=}0.3$, and operators are drawn with probability
\begin{equation}
    \begin{aligned}
        \omega_o &= w^{0}_o\,\mathrm{clip}\!\big(e^{\kappa(Q_o - \bar Q)},\tfrac12,2\big),\\
        p_o &= \phi + (1 - k\phi)\,\omega_o\big/\!\textstyle\sum_{o'}\omega_{o'},
    \end{aligned}
    \label{eq:bandit-prob}
\end{equation}
where $\bar Q$ is the mean quality over the $k$ eligible operators, $w^{0}_o$ a fixed prior, set to $3$ for crossover and $1$ otherwise, offsetting the split of the crossover draw between its two variants, $\kappa{=}6$ a steepness, and $\phi{=}0.05$ a probability floor. Centering on $\bar Q$ promotes above-average operators, while the clip and the floor keep every probability in a bounded band, so the bandit concentrates budget without starving any operator. After the recovery of Section~\ref{sec:method-collapse} the quality estimates are stale, so all $Q_o$ are reset to $0$. Algorithms~\ref{alg:draw} and~\ref{alg:update} implement the draw and the update as \textsc{DrawOperator} and \textsc{UpdateBandit}.

\subsection{Coverage Stagnation and Expert-Preserving Recovery}
\label{sec:method-collapse}
Self-reinforcing prompting can drive the pool onto minor variants of the current best heuristic, so EPOCH performs coverage-aware population management, detecting \emph{coverage stagnation} and countering it with an \emph{expert-preserving recovery} that shrinks the pool to $N_p/2$ heuristics. The stagnation criterion is keyed to the coverage gap rather than the single best heuristic, which can stay unchanged while the pool keeps gaining expert heuristics. Each iteration recomputes $\rho(\pool)$ and its best value $\rho^\star$, registering an improvement when
\begin{equation}
    \rho(\pool) < \rho^\star - \epsilon_\rho ,
    \label{eq:collapse}
\end{equation}
for an absolute margin $\epsilon_\rho{=}10^{-4}$. The recovery fires after $T_\rho{=}10$ iterations without improvement, or $T_{\mathrm{cap}}{=}40$ iterations after the last recovery, the latter forcing a periodic reset in long improvement stretches. It rebuilds the pool from the best-by-average-gap heuristic, the seed, and the highest-expert-count heuristics, resets the bandit's quality estimates, and restarts the stagnation window by setting $\rho^\star$ to the recovered pool's coverage gap. A persistent \emph{expert archive} $\mathcal{A}$, untouched by recovery, retains the best heuristics found on each training instance, so the pool can shrink aggressively without losing discovered expert heuristics. Algorithm~\ref{alg:collapse} implements the recovery as \textsc{Recover}.

\begin{algorithm}[t]
\caption{\textsc{EPOCH}: one training iteration}
\label{alg:main}
\small
\begin{algorithmic}[1]
\REQUIRE policy $\pol$; teacher $\teacher$; pool $\pool$; expert archive $\mathcal{A}$; preference buffer $\mathcal{B}$; iteration index $t$
\ENSURE updated $\pool$, $\mathcal{A}$, $\mathcal{B}$, $\pol$
\STATE {\color{gray}\textit{$\backslash\backslash$ Coverage-aware population management}}
\STATE $\pool \leftarrow \textsc{Recover}(\pool)$ \COMMENT{stagnation and rebuild, Alg.~\ref{alg:collapse}}
\STATE {\color{gray}\textit{$\backslash\backslash$ Reflective rollout}}
\STATE $o \leftarrow \textsc{DrawOperator}()$ \COMMENT{bandit draw, Alg.~\ref{alg:draw}}
\STATE $S \leftarrow \textsc{SelectParents}(\pool, o)$ \COMMENT{expert-aware, Alg.~\ref{alg:parents}}
\STATE $q \leftarrow \mathrm{Prompt}(o, S)$ \COMMENT{operator prompt}
\STATE $Y_1 \leftarrow \{\,y_j \sim \pol(\cdot\mid q)\,\}_{j=1}^{N_1}$ \COMMENT{Round 1, unhinted; evaluate $\gapvec$}
\STATE $z \leftarrow \teacher(q, Y_1)$ \COMMENT{natural-language insight}
\STATE $Y_2 \leftarrow \{\,y_j \sim \pol(\cdot\mid q\oplus z)\,\}_{j=1}^{N_2}$ \COMMENT{Round 2, hinted; evaluate $\gapvec$}
\STATE {\color{gray}\textit{$\backslash\backslash$ Admission, pairing, and bookkeeping}}
\STATE $F \leftarrow \big(\gamma^\star_1(\pool),\dots,\gamma^\star_m(\pool)\big)$ \COMMENT{pre-admission snapshot}
\STATE $\pool \leftarrow \pool \cup \{\,h_y \mid y \in Y_1 \cup Y_2,\ y\ \text{valid}\,\}$ \COMMENT{admit}
\STATE $\mathcal{B} \leftarrow \mathcal{B} \cup \textsc{BuildPairs}(q, Y_1, Y_2)$ \COMMENT{pairs, Alg.~\ref{alg:pairs}}
\STATE $\textsc{UpdateBandit}(o, Y_1, Y_2, F)$ \COMMENT{reward, Alg.~\ref{alg:update}}
\STATE $\mathcal{A} \leftarrow \mathrm{UpdateArchive}(\mathcal{A}, \pool)$ \COMMENT{per-instance experts}
\STATE {\color{gray}\textit{$\backslash\backslash$ Policy update}}
\IF{$t \bmod T_{\mathrm{dpo}} = 0$}
    \STATE $\pol \leftarrow \textsc{DpoBurst}(\pol, \mathcal{B})$ \COMMENT{DPO $+$ RSD, Eq.~\eqref{eq:rsd}}
\ENDIF
\RETURN $\pool,\ \mathcal{A},\ \mathcal{B},\ \pol$
\end{algorithmic}
\end{algorithm}

\subsection{Teacher-Guided Reflective Rollout}
\label{sec:method-reflection}
Each iteration, the policy first generates the Round-1 rollouts $Y_1$, $N_1{=}4$ responses to the operator prompt $q$. Each response states a design idea and implements it as an executable heuristic. An external \emph{teacher} LLM $\teacher$ reads $q$ and $Y_1$ and returns a natural-language \emph{insight} $z$, and the policy generates the Round-2 rollouts $Y_2$, $N_2{=}4$ responses to the insight-augmented prompt $\tilde q = q\oplus z$; we refer to $\tilde q$ and $q$ as the \emph{hinted} and \emph{unhinted} prompts. The teacher contributes only this critique, queried once per iteration, and generates neither heuristics nor training data, so every heuristic and every preference pair comes from the policy itself. The teacher's full prompt appears in Appendix~\ref{app:prompts}. The critique adapts to how many \emph{valid} responses Round~1 produced, where a response $y$ is valid when it follows the output format, parses into an executable heuristic $h_y$, and runs on the training instances without failure.
\begin{itemize}
    \item \emph{Contrastive critique.} With two or more valid heuristics, the teacher contrasts the best and worst, reporting the design choices that reduced the gap and an instance-grounded summary of the training instances each wins.
    \item \emph{Repair critique.} With one valid heuristic, the teacher pairs it with a failed one and combines what worked with what to repair.
    \item \emph{Diagnostic critique.} With no valid heuristic, the teacher inspects two distinct failure modes and proposes a runnable direction.
\end{itemize}

\begin{table*}[t]
\centering
\scriptsize
\setlength{\tabcolsep}{6.6mm}
\setlength{\extrarowheight}{-1.0pt}
\renewcommand{\arraystretch}{0.82}
\resizebox{\textwidth}{!}{
\begin{tabular}{ll cccc}
\toprule
Method & & TSP-20 & \underline{TSP-50} & TSP-100 & TSP-200 \\
\midrule
\multicolumn{6}{c}{\textit{Classical heuristics}} \\
Nearest Neighbor  & & 17.64 & 21.96 & 24.58 & 26.10 \\
\midrule
\multicolumn{6}{c}{\textit{LLM-based AHD: GPT-4o-mini (closed-source)}} \\
\multirow{3}{*}{ReEvo}    & Top-1  & \val{10.55}{3.58} & \val{13.89}{4.48} & \val{15.43}{4.77} & \val{16.64}{5.00} \\
                          & Top-3  & \val{9.18}{2.55}  & \val{12.12}{2.66} & \val{13.66}{3.29} & \val{17.01}{4.28} \\
                          & Top-10 & \val{8.49}{2.52}  & \val{10.92}{2.29} & \val{12.48}{2.70} & \val{14.29}{2.98} \\
\rowsep{6}
\multirow{3}{*}{HSEvo}    & Top-1  & \val{12.58}{4.18} & \val{16.37}{4.62} & \val{18.56}{4.94} & \val{20.86}{5.30} \\
                          & Top-3  & \val{11.38}{5.16} & \val{15.12}{5.64} & \val{17.39}{5.91} & \val{20.14}{5.98} \\
                          & Top-10 & \val{10.59}{5.81} & \val{14.28}{6.32} & \val{16.66}{6.50} & \val{19.65}{6.48} \\
\rowsep{6}
\multirow{3}{*}{MCTS-AHD} & Top-1  & \vals{7.28}{0.71} & \vals{10.77}{0.71} & \vals{12.82}{0.67} & \vals{14.75}{1.03} \\
                          & Top-3  & \val{6.25}{1.07} & \val{9.50}{1.08}  & \vals{11.77}{1.14} & \val{14.00}{1.18} \\
                          & Top-10 & \val{4.54}{1.20} & \val{7.94}{1.17}  & \val{10.49}{1.25} & \val{12.85}{1.41} \\
\rowsep{6}
\multirow{3}{*}{EoH-S}    & Top-1  & \val{8.56}{0.91} & \val{11.69}{0.71} & \val{13.80}{0.91} & \val{15.05}{0.98} \\
                          & Top-3  & \vals{5.05}{0.43} & \vals{9.14}{0.61} & \val{11.80}{0.83} & \vals{13.75}{0.85} \\
                          & Top-10 & \vals{3.22}{0.41} & \vals{7.30}{0.38} & \vals{10.06}{0.38} & \vals{12.39}{0.42} \\
\midrule
\multicolumn{6}{c}{\textit{LLM-based AHD: Llama-3.1-8B}} \\
\multirow{3}{*}{ReEvo}    & Top-1  & \val{10.58}{0.58} & \val{14.87}{1.36} & \val{18.38}{2.10} & \val{26.06}{12.93} \\
                          & Top-3  & \val{10.28}{0.59} & \val{14.15}{2.15} & \val{16.19}{1.65} & \val{19.05}{1.37} \\
                          & Top-10 & \val{10.25}{0.62} & \val{14.10}{2.24} & \val{15.96}{2.01} & \val{18.69}{1.54} \\
\rowsep{6}
\multirow{3}{*}{HSEvo}    & Top-1  & \val{11.46}{1.69} & \val{15.85}{2.63} & \val{18.48}{4.51} & \val{21.34}{7.00} \\
                          & Top-3  & \val{10.61}{1.28} & \val{14.83}{1.76} & \val{17.27}{2.82} & \val{19.47}{3.89} \\
                          & Top-10 & \val{10.29}{1.54} & \val{14.45}{2.10} & \val{16.80}{3.24} & \val{18.99}{4.31} \\
\rowsep{6}
\multirow{3}{*}{MCTS-AHD} & Top-1  & \val{10.27}{0.37} & \val{14.23}{0.65} & \val{16.39}{1.48} & \val{17.87}{1.45} \\
                          & Top-3  & \val{9.76}{0.40}  & \val{13.40}{0.39} & \val{15.40}{0.81} & \val{16.92}{1.04} \\
                          & Top-10 & \val{7.58}{0.63}  & \val{11.73}{0.61} & \val{13.77}{0.91} & \val{15.51}{1.15} \\
\rowsep{6}
\multirow{3}{*}{EoH-S}    & Top-1  & \val{10.69}{0.93} & \val{15.43}{1.58} & \val{17.86}{1.38} & \val{20.04}{1.46} \\
                          & Top-3  & \val{6.05}{0.72} & \val{11.68}{1.06} & \val{14.59}{0.93} & \val{17.31}{1.05} \\
                          & Top-10 & \val{4.70}{0.81} & \val{10.46}{1.02} & \val{13.60}{0.96} & \val{16.45}{1.07} \\
\rowsep{6}
\multirow{3}{*}{CALM}     & Top-1  & \val{6.76}{0.26} & \val{10.02}{0.26} & \val{11.75}{0.27} & \vals{13.47}{0.31} \\
                          & Top-3  & \val{6.55}{0.23} & \val{9.68}{0.55}  & \val{11.34}{0.51} & \val{13.33}{0.47} \\
                          & Top-10 & \val{6.19}{0.54} & \val{9.31}{0.80}  & \val{11.04}{0.67} & \val{13.30}{0.91} \\
\rowsep{6}
\multirow{3}{*}{\shortstack[l]{\textbf{EPOCH}\\(Ours)}} & Top-1  & \vals{6.19}{0.30} & \vals{9.61}{0.25} & \vals{11.63}{0.21} & \val{13.57}{0.36} \\
                                    & Top-3  & \vals{4.31}{0.30} & \vals{7.62}{0.23} & \vals{9.85}{0.36}  & \vals{11.98}{0.49} \\
                                    & Top-10 & \valsb{2.81}{0.22} & \valsb{5.73}{0.15}  & \valsb{8.15}{0.19}  & \valsb{10.74}{0.38} \\
\bottomrule
\end{tabular}
}
\caption{TSP optimality gap (\%) to LKH-3 optima; each cell is the mean over runs with standard deviation, lower is better. Size $n{=}50$ is ID (underlined); $n\in\{20,100,200\}$ are OOD. Within each backbone block the best method per metric and test set is \colorbox{shadegray}{shaded}; each test set's overall best result is shown in \textbf{bold}.}
\label{tab:tsp}
\end{table*}

\subsection{Experiential Preference Optimization}
\label{sec:method-dpo}
An iteration produces a pooled batch $Y_1 \cup Y_2$ of valid and invalid responses. Each valid response receives the reward $r(y)=-\bar\gamma(h_y)$, the negative average gap of its heuristic. Invalid responses require no reward, only a failure-type label that selects the repair and diagnostic critiques of Section~\ref{sec:method-reflection}. Two families of preference pairs are appended to a growing buffer $\mathcal{B}$ by \textsc{BuildPairs} in Algorithm~\ref{alg:pairs}, all conditioned on the \emph{unhinted} prompt $q$:
\begin{itemize}
    \item \textbf{Within-batch pairs.} Following~\citet{surina2025evotune}, the valid responses are split by reward into halves and randomly matched across them, each used at most once. Each invalid response is paired as dispreferred against a random valid response, teaching the policy to avoid unparseable or crashing code.
    \item \textbf{Cross-round pairs.} Each valid Round-2 hinted response $y^+\sim\pol(\cdot\mid\tilde q)$ is preferred over a strictly worse valid Round-1 unhinted response $y^-\in Y_1$ with $r(y^-)<r(y^+)$. These pairs encode the lift from reflection and form the \emph{reflective subset} $\mathcal{B}_{\mathrm{ref}}$.
\end{itemize}
The insight $z$ biases Round~2 toward better heuristics but is only a sampling aid; every pair conditions on the unhinted $q$, so the policy is trained on the prompt distribution it generates from, and the optimality gap certifies every pair's ordering.

Every $T_{\mathrm{dpo}}{=}20$ iterations we run a preference-optimization burst, the \textsc{DpoBurst} step of Algorithm~\ref{alg:main}, over the accumulated buffer. A threshold filter drops pairs whose preferred reward falls below the $30$th percentile of the valid-output rewards observed since the previous burst~\cite{surina2025evotune}, and the buffer accumulates across all iterations, which preserves diversity. The reference $\polref$ is the current model with its low-rank adaptation (LoRA) disabled~\cite{hu2022lora}.

\begin{table*}[t]
\centering
\scriptsize
\setlength{\tabcolsep}{6.5pt}
\setlength{\extrarowheight}{-1.5pt}
\renewcommand{\arraystretch}{0.86}
\resizebox{\textwidth}{!}{
\begin{tabular}{ll cccccc|c}
\toprule
Method & & \underline{1k\_100} & \underline{1k\_500} & \underline{5k\_100} & \underline{5k\_500} & 10k\_100 & 10k\_500 & Avg. \\
\midrule
\multicolumn{8}{c|}{\textit{Classical heuristics}} & \multicolumn{1}{c}{} \\
First-Fit & & 5.02 & 0.25 & 4.65 & 0.55 & 4.36 & 0.50 & 2.55 \\
Best-Fit  & & 4.77 & 0.25 & 4.31 & 0.55 & 4.05 & 0.47 & 2.40 \\
\midrule
\multicolumn{8}{c|}{\textit{LLM-based AHD: GPT-4o-mini (closed-source)}} & \multicolumn{1}{c}{} \\
\multirow{2}{*}{ReEvo}    & Top-1 & \val{3.74}{0.94} & \val{0.40}{0.30} & \val{3.09}{1.12} & \val{0.69}{0.30} & \val{2.84}{1.11} & \val{0.53}{0.12} & \val{1.88}{0.46} \\
                          & Top-4 & \val{3.61}{1.01} & \val{0.40}{0.30} & \val{2.98}{1.21} & \val{0.56}{0.13} & \val{2.74}{1.22} & \val{0.44}{0.03} & \val{1.79}{0.53} \\
\rowsep{9}
\multirow{2}{*}{HSEvo}    & Top-1 & \val{2.85}{1.01} & \val{0.40}{0.30} & \val{2.19}{1.15} & \val{0.56}{0.02} & \val{2.01}{1.15} & \val{0.50}{0.04} & \val{1.42}{0.52} \\
                          & Top-4 & \val{2.82}{1.01} & \val{0.20}{0.10} & \val{2.17}{1.17} & \val{0.47}{0.12} & \val{2.01}{1.15} & \val{0.42}{0.10} & \val{1.35}{0.56} \\
\rowsep{9}
\multirow{2}{*}{MCTS-AHD} & Top-1 & \val{2.59}{0.45} & \val{0.15}{0.12} & \val{1.88}{0.47} & \val{0.42}{0.17} & \val{1.71}{0.41} & \val{0.38}{0.12} & \val{1.19}{0.27} \\
                          & Top-4 & \val{2.48}{0.51} & \val{0.10}{0.12} & \val{1.79}{0.53} & \val{0.36}{0.17} & \val{1.61}{0.48} & \val{0.33}{0.13} & \val{1.11}{0.31} \\
\rowsep{9}
\multirow{2}{*}{EoH-S}    & Top-1 & \val{2.62}{0.33} & \val{0.25}{0.16} & \val{1.74}{0.73} & \val{0.42}{0.15} & \val{1.46}{0.83} & \val{0.35}{0.13} & \vals{1.14}{0.33} \\
                          & Top-4 & \val{2.40}{0.38} & \val{0.15}{0.12} & \val{1.41}{0.56} & \val{0.24}{0.16} & \val{1.08}{0.58} & \val{0.18}{0.15} & \vals{0.91}{0.30} \\
\midrule
\multicolumn{8}{c|}{\textit{LLM-based AHD: Llama-3.1-8B}} & \multicolumn{1}{c}{} \\
\multirow{2}{*}{ReEvo}    & Top-1 & \val{4.34}{0.70} & \val{0.25}{0.00} & \val{3.27}{1.25} & \val{0.41}{0.17} & \val{3.04}{1.36} & \val{0.36}{0.16} & \val{1.95}{0.60} \\
                          & Top-4 & \val{4.10}{0.87} & \val{0.20}{0.10} & \val{3.23}{1.24} & \val{0.41}{0.17} & \val{2.98}{1.34} & \val{0.35}{0.16} & \val{1.88}{0.64} \\
\rowsep{9}
\multirow{2}{*}{HSEvo}    & Top-1 & \val{3.39}{0.93} & \val{0.25}{0.00} & \val{2.42}{1.26} & \val{0.42}{0.18} & \val{2.11}{1.33} & \val{0.36}{0.15} & \val{1.49}{0.58} \\
                          & Top-4 & \val{3.36}{0.97} & \val{0.25}{0.00} & \val{2.42}{1.26} & \val{0.40}{0.16} & \val{2.10}{1.33} & \val{0.36}{0.15} & \val{1.48}{0.59} \\
\rowsep{9}
\multirow{2}{*}{MCTS-AHD} & Top-1 & \val{4.14}{0.89} & \val{0.17}{0.12} & \val{3.54}{1.09} & \val{0.46}{0.12} & \val{3.30}{1.07} & \val{0.40}{0.11} & \val{2.00}{0.57} \\
                          & Top-4 & \val{4.11}{0.94} & \val{0.17}{0.12} & \val{3.54}{1.09} & \val{0.46}{0.12} & \val{3.30}{1.07} & \val{0.40}{0.11} & \val{1.99}{0.57} \\
\rowsep{9}
\multirow{2}{*}{EoH-S}    & Top-1 & \val{3.92}{0.78} & \val{0.21}{0.09} & \val{2.89}{0.87} & \val{0.51}{0.07} & \val{2.59}{1.00} & \val{0.45}{0.05} & \val{1.76}{0.38} \\
                          & Top-4 & \val{3.56}{0.73} & \valb{0.08}{0.12} & \val{1.46}{0.57} & \valb{0.16}{0.09} & \val{1.14}{0.69} & \val{0.14}{0.10} & \val{1.09}{0.27} \\
\rowsep{9}
\multirow{2}{*}{CALM}     & Top-1 & \val{3.71}{0.52} & \val{0.59}{0.82} & \val{1.32}{0.40} & \val{0.30}{0.14} & \val{0.82}{0.41} & \val{0.27}{0.14} & \val{1.17}{0.22} \\
                          & Top-4 & \val{3.69}{0.51} & \val{0.59}{0.82} & \val{1.29}{0.43} & \val{0.29}{0.15} & \val{0.80}{0.43} & \val{0.26}{0.13} & \val{1.15}{0.23} \\
\rowsep{9}
\multirow{2}{*}{\textbf{EPOCH (Ours)}} & Top-1 & \val{2.99}{0.70} & \val{0.30}{0.36} & \val{1.66}{0.90} & \val{0.32}{0.21} & \val{1.33}{0.97} & \val{0.28}{0.24} & \vals{1.15}{0.39} \\
                          & Top-4 & \valb{2.34}{0.43} & \val{0.10}{0.12} & \valb{0.74}{0.06} & \val{0.20}{0.07} & \valb{0.36}{0.06} & \valb{0.13}{0.04} & \valsb{0.64}{0.07} \\
\bottomrule
\end{tabular}
}
\caption{OBP optimality gap (\%) to the $L_1$ lower bound on the six held-out Weibull test sets; each cell is the mean over runs with standard deviation, lower is better. Columns are $(\text{items},\text{capacity})$, e.g.\ \texttt{1k\_100} is $1{,}000$ items at capacity $100$; \texttt{Avg.}\ is the mean over the six. Item scales $1\mathrm{k},5\mathrm{k}$ are ID (underlined); $10\mathrm{k}$ is OOD. Within each backbone block the best \texttt{Avg.}\ per metric is \colorbox{shadegray}{shaded}; each test set's overall best result is shown in \textbf{bold}.}
\label{tab:obp}
\end{table*}

\paragraph{Reflective Self-Distillation.}
\label{sec:method-rsd}
Optimizing the buffer with Eq.~\eqref{eq:dpo} teaches the policy to \emph{rank} a reflective winner $y^+$ above the worse unhinted candidate $y^-$, but ranking is not generation. The loss enters a pair only through the margin $\hat r_\theta(y^+)-\hat r_\theta(y^-)$ and need not raise $\log\pol(y^+\mid q)$; it is met largely by lowering $\hat r_\theta(y^-)$, and the reflective pairs are also a minority of the buffer with small margins, since both responses are valid heuristics. The unhinted policy thus learns to prefer the reflective winner without becoming more likely to generate it, although every Round-1 rollout is drawn from that policy.

To close this gap we distill the hinted policy into the unhinted one. Following~\citet{zhao2025rpo}, their discrepancy is the forward KL divergence
\begin{equation}
    \mbox{\small$\displaystyle
    \mathcal{L}_{\mathrm{RSD}}(\pol) = \mathbb{E}_{q,z}\,\mathrm{KL}\!\big(\pol(\cdot\mid\tilde q)\,\big\|\,\pol(\cdot\mid q)\big),$}
    \label{eq:rsd-kl}
\end{equation}
which we estimate, as a reward-filtered surrogate, from the Round-2 winner $y^+$, since $y^+$ is a draw from $\pol(\cdot\mid\tilde q)$ and freezing that hinted distribution reduces the gradient of Eq.~\eqref{eq:rsd-kl} to that of the negative log-likelihood $-\log\pol(y^+\mid q)$. This RSD term, applied to $\mathcal{B}_{\mathrm{ref}}$ alone, gives the burst objective
\begin{equation}
    \mbox{\small$\displaystyle
    \begin{aligned}
        &\mathcal{L}(\pol) = \mathcal{L}_{\mathrm{DPO}}(\pol)\\
        &\quad {}+\; \lambda\,\mathbb{E}_{(q,\,y^+)\,\sim\,\mathcal{B}_{\mathrm{ref}}}\big[-\log\pol(y^+\mid q)\big],
    \end{aligned}$}
    \label{eq:rsd}
\end{equation}
with $\lambda{=}0.1$ in all main runs and Appendix~\ref{app:rsd} giving the derivation. Where DPO is relative, RSD is absolute, directly raising the probability that the unhinted policy generates the reflective winner. Restricting the term to $\mathcal{B}_{\mathrm{ref}}$, whose pairs contain by construction a valid and better hint-guided heuristic, distills genuine reflective gains. Because the Round-1 rollouts are drawn from $\pol(\cdot\mid q)$, the internalized improvement makes them better on their own, and the gain accumulates over the search.

\section{Experiments}
\label{sec:experiments}

We evaluate EPOCH on three COPs, the TSP, OBP, and Knapsack Problem (KP), solved within a step-by-step construction framework. We assess the single best heuristic, the expert-heuristic ensemble, and generalization from the in-domain (ID) training size to out-of-distribution (OOD) sizes. Problem definitions, datasets, and evaluation protocol are in Appendix~\ref{app:problems}, and the KP results in Appendix~\ref{app:full-results}.

\textbf{Benchmarks and metrics.}
We report mean optimality gaps in percent, where lower is better. For TSP this is the gap of the tour length to the LKH-3~\cite{lkh3} optimum averaged over $300$ test instances per size, and for OBP the gap of the bin count to the $L_1$ lower bound averaged over the Weibull test instances. Each method saves a set of heuristics, scored at two granularities. \emph{Top-1} evaluates the saved heuristic with the best training fitness, while \emph{top-$K$} evaluates the saved set as a portfolio, taking on each test instance the lowest cost among the $K$ heuristics. We use $K\in\{1,3,10\}$ for TSP and $K\in\{1,4\}$ for OBP. EPOCH's saved set is the expert-heuristic ensemble from its archive $\mathcal{A}$, ranked by expert count; Appendix~\ref{app:problems} states how each baseline ranks its saved set.

\textbf{Backbones and settings.}
We run EPOCH with two open-source backbones, Qwen2.5-7B and Llama-3.1-8B, fine-tuned online during the search, with \texttt{GPT-5-nano} as the teacher (Appendix~\ref{app:teacher} shows the gains persist with other teachers, including its own backbone). Every method uses a budget of $1{,}000$ heuristic generations, which for EPOCH corresponds to $125$ iterations. Full hyperparameters are in Appendix~\ref{app:impl}.

\textbf{Baselines.}
We compare against five LLM-based AHD methods, CALM~\cite{huang2025calmcoevolutionalgorithmslanguage}, which fine-tunes the generator, and the inference-only ReEvo~\cite{ye2024reevo}, HSEvo~\cite{hsevo}, MCTS-AHD~\cite{zheng2025montecarlotreesearch}, and EoH-S~\cite{eohs}. The inference-only methods are also run with \texttt{GPT-4o-mini} and, in Appendix~\ref{app:full-results}, with \texttt{GPT-5-nano}; CALM and EPOCH require open backbones. For a fair comparison, where an open-backbone baseline makes auxiliary LLM calls that do not generate heuristics, such as reflection, they are served by \texttt{GPT-5-nano}, the teacher model of EPOCH (Appendix~\ref{app:cost}). We further include the classical heuristics nearest-neighbor~\cite{rosenkrantz1977analysis} for TSP, and First-Fit and Best-Fit~\cite{johnson1974worst} for OBP.

\subsection{Main Results}
\label{sec:exp-main}

\textbf{TSP.}
As reported in Table~\ref{tab:tsp} (Qwen2.5-7B in Table~\ref{tab:tsp-full}), EPOCH attains the best or a comparable top-1 gap in every size--backbone setting, improving on CALM by up to $8\%$. The advantage widens at the ensemble level, where the top-10 ensemble cuts the gap at the training size $n{=}50$ by $38$--$40\%$ relative to CALM and $19$--$45\%$ relative to the strongest inference-only baseline, and the top-3 ensemble already outperforms the top-10 ensemble of almost every baseline. The gains persist OOD, with the top-10 ensemble about $20\%$ below CALM at $n{=}200$. Despite far smaller open backbones, EPOCH outperforms the inference-only baselines run with \texttt{GPT-4o-mini} at every metric and test size, and nearest-neighbor by a wide margin.

\textbf{OBP.}
Across the six test sets of Table~\ref{tab:obp} (Qwen2.5-7B in Table~\ref{tab:obp-full}), EPOCH attains the lowest average gap of the open-backbone methods at both top-1 and top-4, with a top-4 average $44$--$55\%$ below CALM, over $40\%$ below the strongest open-backbone inference-only baseline, and $13$--$30\%$ below the strongest \texttt{GPT-4o-mini} baseline.

\textbf{Single-best versus ensemble.}
TSP, OBP, and KP show the same pattern, the best or a comparable top-1 heuristic and far larger gains at the ensemble level. Because parent selection, operator allocation, stagnation recovery, and the deployed ensemble all optimize instance coverage rather than average fitness, the search yields a heuristic set whose per-instance best cost falls far below that of any single heuristic. Full results with Qwen2.5-7B and all classical heuristics, seed-robustness studies, evolutionary curves, and TSPLIB results appear in Appendix~\ref{app:extended-exp}, and significance tests in Appendix~\ref{app:significance}.

\begin{figure}[t]
\centering
\includegraphics[width=\columnwidth]{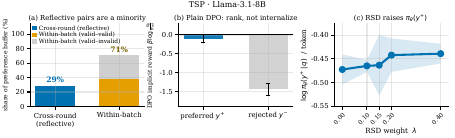}

\vspace{2pt}
\includegraphics[width=\columnwidth]{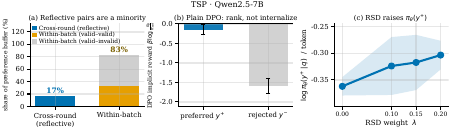}
\caption{Motivation for RSD, on TSP with Llama-3.1-8B (top) and Qwen2.5-7B (bottom): (a)~reflective cross-round pairs are a minority of the preference buffer; (b)~plain DPO at $\lambda{=}0$ satisfies the reflective pairs' margin by lowering the implicit reward of the rejected response, not raising the winner's; (c)~increasing the RSD weight $\lambda$ raises the unhinted policy's per-token log-likelihood of the reflective winners.}
\label{fig:rsd-motivation}
\end{figure}

\begin{figure}[t]
\centering
\includegraphics[width=0.49\columnwidth]{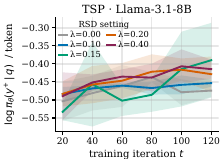}
\hfill
\includegraphics[width=0.49\columnwidth]{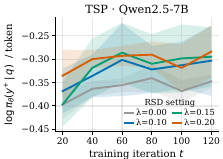}
\caption{Internalization over training, on TSP with Llama-3.1-8B (left) and Qwen2.5-7B (right). The unhinted policy's per-token log-likelihood of the reflective winners rises across DPO bursts for $\lambda>0$ and stays flat at $\lambda{=}0$.}
\label{fig:rsd-internalization}
\end{figure}

\subsection{Ablation Study}
\label{sec:exp-ablation}
We ablate EPOCH one component at a time, on TSP and OBP with Llama-3.1-8B over five runs, scored on $100$ instances per test set. We report in Table~\ref{tab:ablation} the ensemble gap, top-10 for TSP and top-4 for OBP, averaged over the test sets. Appendix~\ref{app:ablation} details each row's removal and replacement, along with the hyperparameter ablation.

\begin{table}[t]
\centering
\footnotesize
\setlength{\tabcolsep}{8.5pt}
\setlength{\extrarowheight}{-0.3pt}
\renewcommand{\arraystretch}{0.95}
\resizebox{\columnwidth}{!}{
\begin{tabular}{l cc}
\toprule
 & TSP & OBP \\
\midrule
\textbf{EPOCH (full)} & \valb{6.79}{0.19} & \valb{0.64}{0.07} \\
\midrule
\multicolumn{3}{l}{\textit{Reflective learning}} \\
\quad{}w/o RSD & \val{7.17}{0.27} & \val{0.79}{0.06} \\
\quad{}w/o cross-round pairs            & \val{6.92}{0.16} & \val{0.75}{0.10} \\
\quad{}w/o instance-grounded insight    & \val{6.90}{0.21} & \val{0.68}{0.07} \\
\quad{}w/o teacher reflection           & \val{6.99}{0.14} & \val{0.83}{0.16} \\
\quad{}w/o preference optimization      & \val{7.01}{0.13} & \val{0.81}{0.15} \\
\midrule
\multicolumn{3}{l}{\textit{Coverage-aware search}} \\
\quad{}w/o expert-aware selection       & \val{6.93}{0.15} & \val{0.81}{0.09} \\
\quad{}w/o adaptive operator allocation & \val{7.15}{0.39} & \val{0.76}{0.12} \\
\quad{}coverage-blind bandit            & \val{7.07}{0.16} & \val{0.77}{0.06} \\
\quad{}w/o coverage-stagnation signal   & \val{7.13}{0.29} & \val{0.69}{0.08} \\
\quad{}w/o expert-preserving recovery   & \val{6.99}{0.17} & \val{0.76}{0.06} \\
\midrule
\multicolumn{3}{l}{\textit{Gap-vector representation}} \\
\quad{}scalar fitness (no gap vector)   & \val{8.77}{0.63} & \val{0.86}{0.18} \\
\bottomrule
\end{tabular}
}
\caption{Component ablations on the Llama-3.1-8B backbone, reported as the ensemble optimality gap (\%), lower is better. The full method is in \textbf{bold}.}
\label{tab:ablation}
\end{table}

\textbf{Gap-vector representation.} Replacing the per-instance gap vector with its scalar average everywhere causes the largest degradation, widening the ensemble gap by $29\%$ on TSP and $34\%$ on OBP. Scoring heuristics by one number removes the per-instance structure that parent selection, the operator reward, recovery, and the deployed ensemble use.

\textbf{Coverage-aware search.} Every coverage mechanism contributes. Removing expert-aware parent selection, the operator bandit, or the coverage term of its reward widens the ensemble gap by $2$--$5\%$ on TSP and $19$--$27\%$ on OBP, with the uniform operator draw the largest on TSP. Replacing the coverage-stagnation signal with a fixed-period reset and the expert-preserving recovery with a single-best rebuild degrade the ensemble by up to $5\%$ on TSP and $19\%$ on OBP, confirming that preserving expert heuristics through recovery is essential.

\textbf{Reflective learning.} The reflective and preference components contribute most on OBP, whose four training instances make lost expert heuristics harder to recover. Removing RSD, the instance-grounded insight, teacher reflection, or preference optimization widens the OBP gap by $6$--$30\%$ and the TSP gap by $2$--$6\%$. The reflective cross-round pairs are a minority of the buffer, $29\%$ for Llama and $17\%$ for Qwen (Figure~\ref{fig:rsd-motivation}a), and on those pairs the DPO loss is met almost entirely by lowering the rejected response's implicit reward rather than raising the winner's (Figure~\ref{fig:rsd-motivation}b), so the update ranks the winner above the worse candidate without making the policy more likely to generate it. The RSD term supplies the missing absolute signal, as a larger weight $\lambda$ raises the unhinted policy's per-token log-likelihood of the reflective winners (Figure~\ref{fig:rsd-motivation}c), and this log-likelihood rises across training for $\lambda>0$ and stays flat at $\lambda{=}0$ (Figure~\ref{fig:rsd-internalization}). The unhinted policy thus internalizes the hint-guided improvement and generates stronger Round-1 heuristics, the gain Table~\ref{tab:ablation} attributes to RSD.

\section{Conclusion}
We propose EPOCH, an LLM-based AHD framework that fine-tunes the generator during the search and represents every heuristic by its per-instance optimality-gap vector rather than a scalar fitness. This representation makes the entire search loop coverage-aware, driving expert-aware parent selection, an adaptive operator bandit, and coverage-based stagnation recovery. Teacher-guided reflection becomes a training signal, and RSD internalizes these improvements into the generator. Across COPs and LLM backbones, EPOCH's complementary expert-heuristic ensembles outperform AHD methods and generalize across problem scales.

\section*{Limitations}
EPOCH fine-tunes only the generator. The teacher LLM that returns the natural-language insight stays fixed throughout the search, so reflection guides the training of the generator without being trained itself. Fine-tuning the reflection jointly with the generator, toward an end-to-end fine-tuned system, is left for future work. EPOCH is evaluated within the step-by-step construction framework; Appendix~\ref{app:problems} explains this choice, and solver-driven frameworks such as ant colony optimization and guided local search remain to be explored.

\bibliography{custom}

@article{Desale2015,
  author  = {Desale, Sachin and Rasool, Akhtar and Andhale, Sushil and Priti, Rane},
  title   = {Heuristic and Meta-Heuristic Algorithms and Their Relevance to the Real World: A Survey},
  journal = {International Journal of Computer Engineering in Research Trends},
  volume  = {2},
  number  = {5},
  pages   = {296--304},
  year    = {2015}
}

@INPROCEEDINGS{nguyen2024ieee,
  author={Tam, Nguyen Thi and Khanh Ly, Tran Ho and Duc, Bui Trong and Hung, Tran Huy and Thanh Binh, Huynh Thi},
  booktitle={2024 IEEE Congress on Evolutionary Computation (CEC)}, 
  title={Multi-Objective Virtual Network Functions Placement and Traffic Routing Problem}, 
  year={2024},
  pages={01-08}
}

@article{
Kirkpatrick1983,
author = {S. Kirkpatrick  and C. D. Gelatt  and M. P. Vecchi },
title = {Optimization by Simulated Annealing},
journal = {Science},
volume = {220},
number = {4598},
pages = {671-680},
year = {1983},
}

@article{Glover1990,
author = {Glover, Fred},
title = {Tabu Search---Part {II}},
journal = {ORSA Journal on Computing},
volume = {2},
number = {1},
pages = {4-32},
year = {1990}
}

@article{Choong2018,
title = {Automatic design of hyper-heuristic based on reinforcement learning},
journal = {Information Sciences},
volume = {436-437},
pages = {89-107},
year = {2018},
author = {Shin Siang Choong and Li-Pei Wong and Chee Peng Lim}
}

@incollection{pillay2018,
author="Pillay, Nelishia
and Qu, Rong",
title="Introduction to Hyper-Heuristics",
bookTitle="Hyper-Heuristics: Theory and Applications",
year="2018",
publisher="Springer",
}

@article{Burke2013,
author = {Burke, Edmund and Gendreau, Michel and Hyde, Matthew and Kendall, Graham and Ochoa, Gabriela and {\"O}zcan, Ender and Qu, Rong},
year = {2013},
month = {07},
pages = {1695-1724},
title = {Hyper-heuristics: A survey of the state of the art},
volume = {64},
journal = {Journal of the Operational Research Society}
}

@book{Langdon2013,
  title={Foundations of Genetic Programming},
  author={Langdon, W.B. and Poli, R.},
  year={2013},
  publisher = {Springer-Verlag},
}

@INPROCEEDINGS{gpahd,
  author={Duflo, Gabriel and Kieffer, Emmanuel and Brust, Matthias R. and Danoy, Gr{\'e}goire and Bouvry, Pascal},
  booktitle={2019 IEEE International Parallel and Distributed Processing Symposium Workshops (IPDPSW)}, 
  title={A {GP} Hyper-Heuristic Approach for Generating {TSP} Heuristics}, 
  year={2019},
  volume={},
  number={},
  pages={521-529},
}

@article{reinelt1991tsplib,
  title={{TSPLIB}---{A} traveling salesman problem library},
  author={Reinelt, Gerhard},
  journal={ORSA journal on computing},
  volume={3},
  number={4},
  pages={376--384},
  year={1991},
}

@article{liu2023algorithm,
  title={Algorithm evolution using large language model},
  author={Liu, Fei and Tong, Xialiang and Yuan, Mingxuan and Zhang, Qingfu},
  journal={arXiv preprint arXiv:2311.15249},
  year={2023}
}

@Article{funsearch,
  author  = {Romera-Paredes, Bernardino and Barekatain, Mohammadamin and Novikov, Alexander and Balog, Matej and Kumar, M. Pawan and Dupont, Emilien and Ruiz, Francisco J. R. and Ellenberg, Jordan and Wang, Pengming and Fawzi, Omar and Kohli, Pushmeet and Fawzi, Alhussein},
  journal = {Nature},
  title   = {Mathematical discoveries from program search with large language models},
  year    = {2023}
}

@inproceedings{eoh,
author = {Liu, Fei and Tong, Xialiang and Yuan, Mingxuan and Lin, Xi and Luo, Fu and Wang, Zhenkun and Lu, Zhichao and Zhang, Qingfu},
title = {Evolution of heuristics: towards efficient automatic algorithm design using large language model},
year = {2024},
booktitle = {Proceedings of the 41st International Conference on Machine Learning},
series = {ICML'24}
}

@inproceedings{ye2024reevo,
author = {Ye, Haoran and Wang, Jiarui and Cao, Zhiguang and Berto, Federico and Hua, Chuanbo and Kim, Haeyeon and Park, Jinkyoo and Song, Guojie},
title = {{ReEvo}: large language models as hyper-heuristics with reflective evolution},
year = {2024},
booktitle = {Proceedings of the 38th International Conference on Neural Information Processing Systems}
}

@inproceedings{
shinn2023reflexion,
title={Reflexion: language agents with verbal reinforcement learning},
author={Noah Shinn and Federico Cassano and Ashwin Gopinath and Karthik R Narasimhan and Shunyu Yao},
booktitle={37th Conference on Neural Information Processing Systems},
year={2023}
}

@inproceedings{hsevo,
  title={Hsevo: Elevating automatic heuristic design with diversity-driven harmony search and genetic algorithm using llms},
  author={Dat, Pham Vu Tuan and Doan, Long and Binh, Huynh Thi Thanh},
  booktitle={Proceedings of the AAAI Conference on Artificial Intelligence},
  volume={39},
  number={25},
  pages={26931--26938},
  year={2025}
}

@inproceedings{
    zheng2025montecarlotreesearch,
    title={Monte {C}arlo Tree Search for Comprehensive Exploration in {LLM}-Based Automatic Heuristic Design},
    author={Zhi Zheng and Zhuoliang Xie and Zhenkun Wang and Bryan Hooi},
    booktitle={Forty-second International Conference on Machine Learning},
    year={2025}
}

@inproceedings{huang2025calmcoevolutionalgorithmslanguage,
  title={{CALM}: Co-evolution of Algorithms and Language Model for Automatic Heuristic Design},
  author={Ziyao Huang and Weiwei Wu and Kui Wu and Wei-Bin Lee and Jianping Wang},
  booktitle={The Fourteenth International Conference on Learning Representations},
  year={2026}
}

@inproceedings{eohs,
  title={{EoH-S}: Evolution of Heuristic Set using {LLM}s for Automated Heuristic Design},
  author={Liu, Fei and Liu, Yilu and Zhang, Qingfu and Tong, Xialiang and Yuan, Mingxuan},
  booktitle={The Fortieth AAAI Conference on Artificial Intelligence},
  year={2026}
}

@inproceedings{pryzant-etal-2023-automatic,
    title = "Automatic Prompt Optimization with ``Gradient Descent'' and Beam Search",
    author = "Pryzant, Reid  and
      Iter, Dan  and
      Li, Jerry  and
      Lee, Yin  and
      Zhu, Chenguang  and
      Zeng, Michael",
    booktitle = "Proceedings of the 2023 Conference on Empirical Methods in Natural Language Processing",
    year = "2023",
    pages = "7957--7968",
    doi = "10.18653/v1/2023.emnlp-main.494",
    url = "https://aclanthology.org/2023.emnlp-main.494",
}

@inproceedings{
madaan2023selfrefine,
title={{Self-Refine}: Iterative Refinement with Self-Feedback},
author={Aman Madaan and Niket Tandon and Prakhar Gupta and Skyler Hallinan and Luyu Gao and Sarah Wiegreffe and Uri Alon and Nouha Dziri and Shrimai Prabhumoye and Yiming Yang and others},
booktitle={Thirty-seventh Conference on Neural Information Processing Systems},
year={2023}
}

@inproceedings{expellm,
author = {Zhao, Andrew and Huang, Daniel and Xu, Quentin and Lin, Matthieu and Liu, Yong-Jin and Huang, Gao},
title = {{ExpeL}: {LLM} agents are experiential learners},
year = {2024},
publisher = {AAAI Press},
booktitle = {Proceedings of the AAAI Conference on Artificial Intelligence},
articleno = {2188},
numpages = {11}
}

@article{wang2025planningheuristicsstrategicplanning,
  title={Planning of Heuristics: Strategic Planning on Large Language Models with {M}onte {C}arlo Tree Search for Automating Heuristic Optimization},
  author={Wang, Hui and Zhang, Xufeng and Mu, Chaoxu},
  journal={arXiv preprint arXiv:2502.11422},
  year={2025}
}

@article{liu2025fitnesslandscapelargelanguage,
      title={Fitness Landscape of Large Language Model-Assisted Automated Algorithm Search}, 
      author={Fei Liu and Qingfu Zhang and Jialong Shi and Xialiang Tong and Kun Mao and Mingxuan Yuan},
      journal={arXiv preprint arXiv:2504.19636},
      year={2025}
}

@inproceedings{vsteincodeevolutiongraphs,
author = {van Stein, Niki and V. Kononova, Anna and Kotthoff, Lars and B\"{a}ck, Thomas},
title = {Code Evolution Graphs: Understanding Large Language Model Driven Design of Algorithms},
year = {2025},
booktitle = {Proceedings of the Genetic and Evolutionary Computation Conference},
pages = {943--951},
numpages = {9},
}

@article{meyerson,
author = {Meyerson, Elliot and Nelson, Mark J. and Bradley, Herbie and Gaier, Adam and Moradi, Arash and Hoover, Amy K. and Lehman, Joel},
title = {Language Model Crossover: Variation through Few-Shot Prompting},
year = {2024},
volume = {4},
number = {4},
journal = {ACM Trans. Evol. Learn. Optim.},
articleno = {27},
numpages = {40},
}

@inproceedings{evoprompting,
 author = {Chen, Angelica and Dohan, David and So, David},
 booktitle = {Advances in Neural Information Processing Systems},
 pages = {7787--7817},
 title = {{EvoPrompting}: Language Models for Code-Level Neural Architecture Search},
 volume = {36},
 year = {2023}
}

@inproceedings{rafailov2023dpo,
  title={Direct Preference Optimization: Your Language Model is Secretly a Reward Model},
  author={Rafailov, Rafael and Sharma, Archit and Mitchell, Eric and Manning, Christopher D. and Ermon, Stefano and Finn, Chelsea},
  booktitle={Advances in Neural Information Processing Systems},
  volume={36},
  year={2023}
}

@inproceedings{surina2025evotune,
  title={Algorithm Discovery With {LLM}s: Evolutionary Search Meets Reinforcement Learning},
  author={{\v{S}}urina, Anja and Mansouri, Amin and Quaedvlieg, Lars C.P.M. and Seddas, Amal and Viazovska, Maryna and Abbe, Emmanuel and Gulcehre, Caglar},
  booktitle={Second Conference on Language Modeling},
  year={2025}
}

@article{shao2024deepseekmath,
  title={{DeepSeekMath}: Pushing the Limits of Mathematical Reasoning in Open Language Models},
  author={Shao, Zhihong and Wang, Peiyi and Zhu, Qihao and Xu, Runxin and Song, Junxiao and Bi, Xiao and Zhang, Haowei and Zhang, Mingchuan and Li, Y. K. and Wu, Y. and Guo, Daya},
  journal={arXiv preprint arXiv:2402.03300},
  year={2024}
}

@article{qwen2025qwen25,
  title={Qwen2.5 Technical Report},
  author={{Qwen Team}},
  journal={arXiv preprint arXiv:2412.15115},
  year={2024}
}

@article{grattafiori2024llama3,
  title={The {L}lama 3 Herd of Models},
  author={Grattafiori, Aaron and others},
  journal={arXiv preprint arXiv:2407.21783},
  year={2024}
}

@inproceedings{hu2022lora,
  title={{LoRA}: Low-Rank Adaptation of Large Language Models},
  author={Hu, Edward J. and Shen, Yelong and Wallis, Phillip and Allen-Zhu, Zeyuan and Li, Yuanzhi and Wang, Shean and Wang, Lu and Chen, Weizhu},
  booktitle={International Conference on Learning Representations},
  year={2022}
}

@article{zhao2025rpo,
  title={Reflective Preference Optimization ({RPO}): Enhancing On-Policy Alignment via Hint-Guided Reflection},
  author={Zhao, Zihui and Li, Zechang},
  journal={arXiv preprint arXiv:2512.13240},
  year={2025}
}

@inproceedings{gungordu2026pathwise,
  title={{PathWise}: Planning through World Model for Automated Heuristic Design via Self-Evolving {LLM}s},
  author={Gungordu, Oguzhan and Xiong, Siheng and Fekri, Faramarz},
  booktitle={International Conference on Machine Learning},
  year={2026}
}

@inproceedings{song2026textfeedback,
  title={Expanding the Capabilities of Reinforcement Learning via Text Feedback},
  author={Song, Yuda and Chen, Lili and Tajwar, Fahim and Munos, R{\'e}mi and Pathak, Deepak and Bagnell, Drew and Singh, Aarti and Zanette, Andrea},
  booktitle={International Conference on Machine Learning},
  year={2026}
}

@inproceedings{shi2026experiential,
  title={Experiential Reinforcement Learning},
  author={Shi, Taiwei and Chen, Sihao and Jiang, Bowen and Song, Linxin and Yang, Longqi and Zhao, Jieyu},
  booktitle={ICLR 2026 Workshop on Lifelong Agents: Learning, Aligning, Evolving},
  year={2026}
}

@inproceedings{ouyang2022instructgpt,
  title={Training Language Models to Follow Instructions with Human Feedback},
  author={Ouyang, Long and Wu, Jeffrey and Jiang, Xu and Almeida, Diogo and Wainwright, Carroll L. and Mishkin, Pamela and Zhang, Chong and Agarwal, Sandhini and Slama, Katarina and Ray, Alex and Schulman, John and Hilton, Jacob and Kelton, Fraser and Miller, Luke and Simens, Maddie and Askell, Amanda and Welinder, Peter and Christiano, Paul and Leike, Jan and Lowe, Ryan},
  booktitle={Advances in Neural Information Processing Systems},
  volume={35},
  year={2022}
}

@incollection{lehman2023elm,
  title={Evolution through Large Models},
  author={Lehman, Joel and Gordon, Jonathan and Jain, Shawn and Ndousse, Kamal and Yeh, Cathy and Stanley, Kenneth O.},
  booktitle={Handbook of Evolutionary Machine Learning},
  pages={331--366},
  year={2024},
  publisher={Springer}
}

@article{rosenkrantz1977analysis,
  title={An Analysis of Several Heuristics for the Traveling Salesman Problem},
  author={Rosenkrantz, Daniel J. and Stearns, Richard E. and Lewis, II, Philip M.},
  journal={SIAM Journal on Computing},
  volume={6},
  number={3},
  pages={563--581},
  year={1977},
  doi={10.1137/0206041}
}

@article{bentley1992fast,
  title={Fast Algorithms for Geometric Traveling Salesman Problems},
  author={Bentley, Jon Louis},
  journal={ORSA Journal on Computing},
  volume={4},
  number={4},
  pages={387--411},
  year={1992},
  publisher={INFORMS}
}

@techreport{christofides1976worst,
  title={Worst-Case Analysis of a New Heuristic for the Travelling Salesman Problem},
  author={Christofides, Nicos},
  year={1976},
  number={388},
  institution={Graduate School of Industrial Administration, Carnegie Mellon University}
}

@article{johnson1974worst,
  title={Worst-Case Performance Bounds for Simple One-Dimensional Packing Algorithms},
  author={Johnson, David S. and Demers, Alan and Ullman, Jeffrey D. and Garey, Michael R. and Graham, Ronald L.},
  journal={SIAM Journal on Computing},
  volume={3},
  number={4},
  pages={299--325},
  year={1974},
  doi={10.1137/0203025}
}

@article{welch1947,
  title={The Generalization of `{S}tudent's' Problem when Several Different Population Variances are Involved},
  author={Welch, Bernard L.},
  journal={Biometrika},
  volume={34},
  number={1/2},
  pages={28--35},
  year={1947},
  publisher={Oxford University Press}
}

@article{wilcoxon1945,
  title={Individual Comparisons by Ranking Methods},
  author={Wilcoxon, Frank},
  journal={Biometrics Bulletin},
  volume={1},
  number={6},
  pages={80--83},
  year={1945},
  publisher={JSTOR}
}

@techreport{lkh3,
  author      = {Helsgaun, Keld},
  title       = {An Extension of the {L}in--{K}ernighan--{H}elsgaun {TSP} Solver for Constrained Traveling Salesman and Vehicle Routing Problems},
  institution = {Roskilde University},
  year        = {2017}
}

\appendix

\makeatletter
\setlength{\@fptop}{0pt}
\setlength{\@fpsep}{8pt plus 1fil}
\setlength{\@fpbot}{0pt plus 4fil}
\setlength{\@dblfptop}{0pt}
\setlength{\@dblfpsep}{8pt plus 1fil}
\setlength{\@dblfpbot}{0pt plus 4fil}
\makeatother

\section{Extended Related Work}
\label{app:related}

\paragraph{Heuristic design and hyper-heuristics.}
Exact algorithms for NP-hard COPs rarely scale to practical instance sizes, so routing, scheduling, and design problems are commonly solved with heuristics~\cite{Desale2015,nguyen2024ieee} and with metaheuristics such as simulated annealing and tabu search~\cite{Kirkpatrick1983,Glover1990}. Such solvers encode expert knowledge, and adapting one to a new problem variant largely restarts the design effort~\cite{pillay2018,Choong2018}. Hyper-heuristics and AHD instead automate the design itself, selecting or constructing heuristics for a problem class~\cite{Burke2013}. Genetic programming is the classical realization, evolving heuristic programs built from a human-defined function set~\cite{Langdon2013,gpahd}, and free-form LLM code generation widens the search space beyond what that function set expresses.

\paragraph{LLMs in evolutionary search.}
A line of work runs a standard evolutionary algorithm but lets an LLM generate the offspring, prompting it with the selected parents to produce new code or text, while fitness evaluation and survivor selection stay outside the model~\cite{lehman2023elm,meyerson,evoprompting}. LLM-based AHD builds on this setup, and EPOCH keeps the same outer loop but additionally \emph{trains} the operator rather than treating it as fixed.

\paragraph{LLM-based AHD frameworks.}
FunSearch~\cite{funsearch} pioneered evolutionary program search with an LLM proposer, EoH~\cite{eoh} co-evolves thoughts and code, ReEvo~\cite{ye2024reevo} adds reflection~\cite{shinn2023reflexion,pryzant-etal-2023-automatic}, HSEvo~\cite{hsevo} couples diversity metrics with harmony search, and MCTS-AHD~\cite{zheng2025montecarlotreesearch} organizes heuristics in a Monte Carlo tree. Analyses of the AHD fitness landscape and code-evolution dynamics further motivate diversity~\cite{liu2025fitnesslandscapelargelanguage,vsteincodeevolutiongraphs}, and planning-based approaches organize the search over a tree~\cite{wang2025planningheuristicsstrategicplanning} or an entailment graph that carries derivation history~\cite{gungordu2026pathwise}. Most relevant to our coverage view, EoH-S~\cite{eohs} evolves a \emph{set} of complementary heuristics, using per-instance performance vectors to select survivors that specialize on different parts of the instance distribution. All of these methods keep the LLM frozen, whereas EPOCH trains the generator and uses the per-instance best gaps as an online control signal throughout the run.

\paragraph{Training the LLM generator for AHD.}
CALM~\cite{huang2025calmcoevolutionalgorithmslanguage} co-evolves the heuristic pool and the LLM, fine-tuning with GRPO~\cite{shao2024deepseekmath} on a progressive scalar reward and using a stagnation-driven collapse that resets the pool to the seed and current best. EvoTune~\cite{surina2025evotune} interleaves FunSearch-style search with periodic DPO over a program database treated as a replay buffer, using a forward-KL variant that preserves output diversity and a dynamic reward-percentile filter. EPOCH likewise fine-tunes the generator, but governs selection, operator allocation, and recovery by per-instance gap vectors rather than scalar fitness, converts teacher-guided reflection into preference pairs and distills the hint-guided improvement into the unhinted policy with an RSD term, and deploys a complementary ensemble of expert heuristics.

\paragraph{Reflection and natural-language feedback.}
Reflexion~\cite{shinn2023reflexion} and Self-Refine~\cite{madaan2023selfrefine} improve LLM outputs through in-context natural-language feedback, ExpeL~\cite{expellm} reuses accumulated experience across tasks, and prompt-optimization methods treat textual critiques as gradients~\cite{pryzant-etal-2023-automatic}; all act at inference time without updating weights. A recent thread instead folds text feedback into training. Text feedback can be distilled into first-attempt behavior so the model needs no feedback at test time~\cite{song2026textfeedback}, and an experience--reflection--consolidation loop can internalize corrective revisions into the policy~\cite{shi2026experiential}, though both target domains other than heuristic design. EPOCH brings this idea to AHD, where the executable, verifiable reward orders reflection-conditioned generations without a learned judge, and its RSD term internalizes the hint-guided improvement into the policy's unhinted behavior so it carries into later generations of the search.

\paragraph{Preference optimization and learning from feedback.}
Aligning LLMs from preferences was popularized by reinforcement learning from human feedback (RLHF)~\cite{ouyang2022instructgpt}; DPO~\cite{rafailov2023dpo} removes the learned reward model by optimizing preferences directly, and Reflective Preference Optimization~\cite{zhao2025rpo} constructs the pairs by conditioning regeneration on a critic's hint and contrasting hinted with unhinted responses, keeping the update on-policy. EPOCH adopts DPO for the generator together with this reflection-as-preference construction, but the preference ordering is certified by the optimality gap on executed code rather than by a learned or model-based judge, and the contrasted responses are a hinted second batch versus a strictly worse unhinted first batch from the same evolutionary iteration. That work also distills the hint-conditioned behavior back into the unconditioned policy to remove a test-time dependence on the hint. EPOCH's RSD term shares this mechanism but is motivated within training, makes each iteration's unhinted rollouts better and accumulates across generations, and is gated to the pairs whose verifiable reward certifies a genuine reflective gain.

\section{Problem Definitions and Datasets}
\label{app:problems}

We instantiate EPOCH on three COPs, solved within a step-by-step construction framework; the function signatures match the conventions of prior LLM-based AHD work~\cite{zheng2025montecarlotreesearch,ye2024reevo}.

\paragraph{Solution framework.}
We evaluate within the step-by-step construction framework because there the heuristic alone builds the solution, so the optimality gap measures the heuristic an AHD framework discovered. In solver-driven frameworks such as ant colony optimization and guided local search~\cite{ye2024reevo,zheng2025montecarlotreesearch}, the heuristic only biases a solver that continues the search on its own, through pheromone updates or local search, so the solver narrows the differences between heuristics and the gap reflects the solver as much as the discovery method. Each candidate evaluation also runs the solver on every training instance, which makes it far costlier than a construction run and the $1{,}000$-generation search of every method correspondingly slower. Extending EPOCH to these frameworks is left for future work.

\paragraph{Traveling Salesman Problem (TSP).}
A TSP instance is a complete graph on $n$ nodes with coordinates sampled uniformly from $[0,1]^2$ and Euclidean edge costs; a solution is a Hamiltonian tour scored by its total length. Within the step-by-step construction framework, the LLM designs the key function below, which chooses the next city from the current one given the unvisited set and the distance matrix, and the framework calls it repeatedly until the tour is complete.
\begin{codebox}{TSP key function}
\scriptsize
\begin{verbatim}
select_next_node(current_node, destination_node,
                 unvisited_nodes, distance_matrix)
    -> next_node
\end{verbatim}
\end{codebox} Per-instance gaps in Eq.~\eqref{eq:gap} are computed against LKH-3~\cite{lkh3} optimal tour lengths. Training uses $m{=}64$ instances at $n{=}50$; held-out evaluation uses a range of sizes ($n\in\{20,50,100,200\}$) to test OOD generalization.

\paragraph{Online Bin Packing (OBP).}
In OBP, items arrive sequentially with sizes drawn from a Weibull distribution and must be placed immediately into a bin that can hold them, minimizing the number of bins used. The LLM designs the priority function below, which scores each feasible bin for the arriving item, and the item is placed in the highest-priority bin.
\begin{codebox}{OBP key function}
\scriptsize
\begin{verbatim}
step(item_size, remaining_capacity) -> priority
\end{verbatim}
\end{codebox} Per-instance gaps are computed against the $L_1$ lower bound $\lceil(\sum_i s_i)/C\rceil$ on the number of bins~\cite{eoh}. Because the training set comprises few instances, the inference-time ensemble is reported up to the available number of experts.

\paragraph{0-1 Knapsack Problem (KP).}
Following the KP setting of MCTS-AHD and ReEvo~\cite{zheng2025montecarlotreesearch,ye2024reevo}, an instance has $n$ items with weights and values sampled uniformly from $[0,1]$ and a knapsack of capacity $25$ ($12.5$ at $n{=}50$); a solution is a subset of items whose total weight fits the capacity, scored by its total value. Within the step-by-step construction framework, the LLM designs the key function below, which chooses the next item to pack given the remaining capacity and the weights and values of the items not yet packed, and the framework calls it repeatedly until no remaining item fits.
\begin{codebox}{KP key function}
\scriptsize
\begin{verbatim}
select_next_item(remaining_capacity, weights,
                 values) -> next_item
\end{verbatim}
\end{codebox} Per-instance gaps are computed against the optimal value from dynamic programming, with the negated value as the cost in Eq.~\eqref{eq:gap}. Training uses $m{=}64$ instances at $n{=}100$; held-out evaluation uses $n\in\{50,100,200\}$.

\paragraph{Datasets and evaluation protocol.}
As summarized in Table~\ref{tab:datasets}, for TSP we train on $64$ instances at $n{=}50$ and evaluate on held-out sets of $300$ instances at each of $n\in\{20,50,100,200\}$, where $n{=}50$ is ID and the other sizes are OOD. For OBP we train on four Weibull instances spanning $\{1\mathrm{k},5\mathrm{k}\}$ items at capacity $100$, and evaluate on the six held-out Weibull sets $\{1\mathrm{k},5\mathrm{k},10\mathrm{k}\}\times\{100,500\}$ of five instances each; the $10\mathrm{k}$-item sets are OOD in scale. For KP we train on $64$ instances at $n{=}100$ and evaluate on held-out sets of $1{,}000$ instances at each of $n\in\{50,100,200\}$, where $n{=}100$ is ID, with $K\in\{1,3,10\}$ as for TSP. Following standard practice, both training and test instances are sampled under fixed settings used by prior LLM-based AHD work~\cite{ye2024reevo,eoh,zheng2025montecarlotreesearch}, and every method sees only the training set during search and is scored on the held-out sets afterward. We report the mean over five independent runs. During search, each heuristic evaluates the full training set under a $60$\,s wall-clock budget and is treated as a runtime failure if it exceeds it. At test time, TSP heuristics run under a per-instance wall-clock budget of $15$\,s at $n{=}50$, $30$\,s at $n{=}100$, and $60$\,s at $n{=}200$, KP heuristics under $10$\,s at every size, and a heuristic exceeding its budget is recorded as failing those instances.

\begin{table}[t]
\centering
\small
\setlength{\tabcolsep}{4pt}
\renewcommand{\arraystretch}{1.1}
\resizebox{\columnwidth}{!}{
\begin{tabular}{l cc cc}
\toprule
& \multicolumn{2}{c}{Train} & \multicolumn{2}{c}{Test} \\
\cmidrule(lr){2-3}\cmidrule(lr){4-5}
Problem & Scale & \#Inst. & Scale & \#Inst. \\
\midrule
TSP & $n{=}\underline{50}$ & 64 & $n\in\{20,\underline{50},100,200\}$ & 300 \\
OBP & $\{\underline{1\mathrm{k}},\underline{5\mathrm{k}}\}$, $C{=}100$ & 4 & $\{\underline{1\mathrm{k}},\underline{5\mathrm{k}},10\mathrm{k}\}{\times}\{100,500\}$ & 5 \\
KP  & $n{=}\underline{100}$ & 64 & $n\in\{50,\underline{100},200\}$ & 1,000 \\
\bottomrule
\end{tabular}
}
\caption{Training and test datasets. ID scales are underlined; the rest are OOD. \#Inst.\ is per size (TSP, KP) or per $(\text{items},\text{capacity})$ set (OBP). TSP gaps are to LKH-3 optima, OBP gaps to the $L_1$ lower bound, and KP gaps to the optimum.}
\label{tab:datasets}
\end{table}

\paragraph{Top-$K$ snapshots of the methods.}
Every method ends a run with a saved set of heuristics carrying its own ranking. \emph{Top-1} evaluates the member of that set with the best training fitness, identically for every method, and \emph{top-$K$} routes each test instance to the lowest cost among $K$ of its members; for OBP, $K{=}4$ matches the four training instances. For EPOCH and EoH-S, whose saved sets are ranked for coverage, the ensemble takes the first $K$ of that ranking, and for every other method it takes the $K$ with the best training fitness. At $K$ equal to the full set size the two rules coincide. Top-$K$ thus scores the set each method itself returns, with no re-selection at test time; it is the deployment a set-based AHD method targets, and top-1 gives the single-heuristic view alongside. Since EoH-S is scored by the same procedure on its own coverage-ranked set, the top-$K$ margin of EPOCH over it in Tables~\ref{tab:tsp-full} and~\ref{tab:obp-full} reflects the search rather than the selection of the deployed set. EPOCH deploys its expert-heuristic ensemble ranked by expert count, Eq.~\eqref{eq:expert}, with the training-fitness-best heuristic always included. EoH-S saves its final complementary set in its own order, the best individual first, followed by the members added for coverage. CALM saves its all-time-best buffer ranked by training fitness. For HSEvo and ReEvo we take the ten best heuristics evaluated during each run, ranked by training fitness, so that strong candidates displaced by generational replacement are retained. MCTS-AHD saves the ten tree nodes with the highest back-propagated $Q$ value, the quantity its tree search ranks nodes by. On OBP every method, including HSEvo, is trained on the same four-instance Weibull set, and we omit the Worst-Fit rule as it is orders of magnitude from competitive. Among methods that fine-tune the generator, we omit EvoTune~\cite{surina2025evotune}: CALM outperforms it under matched comparisons that include a DPO-trained CALM variant against the DPO-based EvoTune~\cite{huang2025calmcoevolutionalgorithmslanguage}, and within our framework the scalar-fitness and reflective-learning ablations of Appendix~\ref{app:ablation} isolate, under a matched budget, the gains of the coverage-aware search and of the reflective preference construction over preference optimization on scalar fitness alone. The classical heuristics are deterministic single programs, so their top-1 and top-$K$ scores coincide.

\section{Algorithm Pseudocodes}
\label{app:algorithms}

This section provides the complete pseudocode of the subalgorithms invoked by Algorithm~\ref{alg:main}. Algorithm~\ref{alg:parents} is the expert-aware parent selector that supplies operator parents from the reproduction population. Algorithms~\ref{alg:draw} and~\ref{alg:update} are the operator bandit's draw and update steps. Algorithm~\ref{alg:collapse} is the expert-preserving recovery that restores diversity when the coverage gap stagnates, and Algorithm~\ref{alg:pairs} is the preference-pair construction that turns one iteration's generations into training pairs.

\paragraph{Parent selection.}
Algorithm~\ref{alg:parents} mixes exploitation with coverage over the reproduction population $\pool_{N_p}$, the $N_p$ heuristics of the pool with the lowest average gap. A single-parent mutation operator samples either by expert count, which recruits heuristics for under-covered instances, or by average gap, which refines strong all-rounders. The crossover operator samples its first parent by average gap and its second parent either by complementarity or by average gap, selecting the matching crossover variant of Section~\ref{sec:method-ops}.

\begin{algorithm}[t]
\caption{\textsc{SelectParents}: expert-aware parent selection}
\label{alg:parents}
\small
\begin{algorithmic}[1]
\REQUIRE pool $\pool$, population size $N_p$, operator $o$, probabilities $p_{\mathrm{exp}},p_{\mathrm{comp}}$, temperature $\tau$
\ENSURE parent set $S$
\STATE $\pool_{N_p} \leftarrow \operatorname{Top}_{N_p}\big(\pool;\ \bar\gamma(\cdot)\!\uparrow\big)$ \COMMENT{reproduction population}
\STATE {\color{gray}\textit{$\backslash\backslash$ Single-parent mutation operators}}
\IF{$o$ is a single-parent mutation operator}
    \IF{$\mathrm{rand}() < p_{\mathrm{exp}}$ \textbf{and} some $h$ has $e(h;\pool_{N_p})>0$}
        \STATE sample $h_1 \in \pool_{N_p}$ with prob.\ $\propto 1/\mathrm{rank}$, descending $e(\cdot;\pool_{N_p})$ \COMMENT{target under-covered instances}
    \ELSE
        \STATE sample $h_1 \in \pool_{N_p}$ with prob.\ $\propto 1/\mathrm{rank}$, ascending $\bar\gamma(\cdot)$ \COMMENT{refine strong}
    \ENDIF
    \RETURN $\{h_1\}$
\ENDIF
\STATE {\color{gray}\textit{$\backslash\backslash$ Crossover}}
\STATE sample $h_1 \in \pool_{N_p}$ with prob.\ $\propto 1/\mathrm{rank}$, ascending $\bar\gamma(\cdot)$ \COMMENT{parent 1}
\IF{$\mathrm{rand}() < p_{\mathrm{comp}}$}
    \STATE sample $h_2 \sim \mathrm{softmax}\big(d(h_1,\cdot)/\tau\big)$ over $\pool_{N_p}\setminus\{h_1\}$ \COMMENT{complementary variant}
\ELSE
    \STATE sample $h_2 \in \pool_{N_p}$ with prob.\ $\propto 1/\mathrm{rank}$, ascending $\bar\gamma(\cdot)$, $h_2 \neq h_1$ \COMMENT{exploitative variant}
\ENDIF
\RETURN $\{h_1, h_2\}$
\end{algorithmic}
\end{algorithm}

\paragraph{Operator bandit.}
Algorithm~\ref{alg:draw} draws operators in proportion to a bounded weight that is centered on the mean quality across eligible operators, under the probability floor $\phi$. Algorithm~\ref{alg:update} then revises the quality $Q_o$ from the expert-coverage gain $g_o$ together with the parent-beating and validity fractions of the iteration's children, and resets $Q_o$ to $0$ on recovery.

\begin{algorithm}[t]
\caption{\textsc{DrawOperator}: bandit operator draw}
\label{alg:draw}
\small
\begin{algorithmic}[1]
\REQUIRE pool $\pool$, quality $\{Q_o\}$, priors $\{w^0_o\}$, steepness $\kappa$, floor $\phi$
\ENSURE sampled operator $o$
\STATE $\opset' \leftarrow \opset$
\IF{$|\pool| < 2$}
    \STATE remove crossover from $\opset'$ \COMMENT{needs two parents}
\ENDIF
\STATE $\bar Q \leftarrow \mathrm{mean}_{o\in\opset'}\,Q_o$
\STATE $\omega_o \leftarrow w^0_o\,\mathrm{clip}(e^{\kappa(Q_o-\bar Q)},\tfrac12,2)$ for $o \in \opset'$ \COMMENT{$\omega_o\in[\tfrac12 w^0_o,2w^0_o]$}
\STATE $p_o \leftarrow \phi + (1-|\opset'|\phi)\,\omega_o\big/\!\sum_{o'\in\opset'}\omega_{o'}$ \COMMENT{probability floor}
\RETURN $o \sim p$
\end{algorithmic}
\end{algorithm}

\begin{algorithm}[t]
\caption{\textsc{UpdateBandit}: bandit quality update}
\label{alg:update}
\small
\begin{algorithmic}[1]
\REQUIRE operator $o$, Round-1/Round-2 rollouts $Y_1, Y_2$, best-gap snapshot $F$, margin $\epsilon_\rho$, rate $\eta$, weights $w_e,w_p,w_v$
\ENSURE updated quality $Q_o$
\STATE $V \leftarrow \{\, y \in Y_1 \cup Y_2 : y\ \text{valid}\,\}$
\STATE $g_o \leftarrow \tfrac{1}{m}\big|\{\, i : \min_{y\in V}\gamma_i(h_y) < F_i - \epsilon_\rho \,\}\big|$ \COMMENT{instances won vs.\ $F$}
\STATE $\nu_p \leftarrow \tfrac{1}{|V|}\big|\{\, y \in V : \bar\gamma(h_y) < \bar\gamma_{\mathrm{par}} \,\}\big|$ \COMMENT{$\bar\gamma_{\mathrm{par}}$: best parent's average gap}
\STATE $\nu_v \leftarrow |V| \,/\, |Y_1 \cup Y_2|$ \COMMENT{valid fraction}
\STATE $R_o \leftarrow \mathrm{clip}\big(w_e g_o + w_p \nu_p + w_v \nu_v,\ 0,\ 1\big)$
\STATE $Q_o \leftarrow (1-\eta)Q_o + \eta R_o$ \COMMENT{EMA; $Q_o\!\leftarrow\!0$ on recovery}
\RETURN $Q_o$
\end{algorithmic}
\end{algorithm}

\paragraph{Expert-preserving recovery.}
Algorithm~\ref{alg:collapse} tracks the coverage gap $\rho(\pool)$; after $T_\rho$ iterations without an $\epsilon_\rho$ improvement, or $T_{\mathrm{cap}}$ iterations since the last recovery, it rebuilds a pool of $N_p/2$ heuristics from the best heuristic, the seed, and the highest-expert-count heuristics, resets the bandit's quality estimates, and restarts the stagnation window by setting $\rho^\star$ to the recovered pool's coverage gap. The persistent expert archive of Section~\ref{sec:method-collapse} is updated each iteration outside this procedure and remains intact across recoveries.

\begin{algorithm}[t]
\caption{\textsc{Recover}: expert-preserving recovery}
\label{alg:collapse}
\small
\begin{algorithmic}[1]
\REQUIRE pool $\pool$, best gap $\rho^\star$, age $a$, patience $T_\rho$, iterations-since-recovery $c$, cap $T_{\mathrm{cap}}$, margin $\epsilon_\rho$, population size $N_p$, seed heuristic $h_{\mathrm{seed}}$
\ENSURE updated pool
\STATE {\color{gray}\textit{$\backslash\backslash$ Stagnation criterion}}
\STATE $\rho \leftarrow \tfrac{1}{m}\sum_i \min_{h\in\pool}\gamma_i(h)$ \COMMENT{coverage gap, Eq.~\eqref{eq:rho}}
\STATE $c \leftarrow c + 1$
\IF{$\rho < \rho^\star - \epsilon_\rho$}
    \STATE $\rho^\star \leftarrow \rho$;\ \ $a \leftarrow 0$ \COMMENT{coverage gap improved}
\ELSE
    \STATE $a \leftarrow a + 1$
\ENDIF
\IF{$a < T_\rho$ \textbf{and} $c < T_{\mathrm{cap}}$}
    \RETURN $\pool$ \COMMENT{within patience and cap}
\ENDIF
\STATE {\color{gray}\textit{$\backslash\backslash$ Expert-preserving rebuild}}
\STATE $\pool_{\mathrm{rec}} \leftarrow \{\arg\min_{h\in\pool} \bar\gamma(h)\} \cup \{h_{\mathrm{seed}}\}$ \COMMENT{best-by-average-gap and seed}
\STATE $\pool_{\mathrm{rec}} \leftarrow \pool_{\mathrm{rec}} \cup \operatorname{Top}_{\,N_p/2-|\pool_{\mathrm{rec}}|}\big(\pool \setminus \pool_{\mathrm{rec}};\ e(\cdot;\pool)\!\downarrow\big)$ \COMMENT{ties: lower $\bar\gamma$}
\STATE reset bandit;\ \ $\rho^\star \leftarrow \tfrac{1}{m}\sum_i \min_{h\in\pool_{\mathrm{rec}}}\gamma_i(h)$;\ \ $a \leftarrow 0$;\ \ $c \leftarrow 0$
\RETURN $\pool_{\mathrm{rec}}$
\end{algorithmic}
\end{algorithm}

\paragraph{Preference-pair construction.}
Algorithm~\ref{alg:pairs} forms within-batch pairs by splitting valid responses at the reward median and matching across the halves, pairs each invalid response as dispreferred against a randomly chosen valid one, and adds cross-round pairs that prefer a hinted Round-2 response over a strictly worse unhinted Round-1 response. Every pair is conditioned on the unhinted prompt $q$.

\begin{algorithm}[t]
\caption{\textsc{BuildPairs}: preference-pair construction}
\label{alg:pairs}
\small
\begin{algorithmic}[1]
\REQUIRE unhinted prompt $q$, Round-1 rollouts $Y_1$, Round-2 rollouts $Y_2$
\ENSURE preference pairs $\mathcal{B}_q$
\STATE $V \leftarrow$ valid responses in $Y_1\cup Y_2$;\ \ $I \leftarrow$ invalid responses;\ \ $\mathcal{B}_q \leftarrow \emptyset$
\STATE {\color{gray}\textit{$\backslash\backslash$ Within-batch pairs}}
\STATE $(V^+, V^-) \leftarrow$ higher- and lower-scoring halves of $V$ by $r(\cdot)$
\STATE add $(q, y^+, y^-)$ for a random matching of $V^+$ with $V^-$ \COMMENT{each response used at most once}
\FORALL{$y^- \in I$}
    \STATE add $(q,\, y,\, y^-)$ with $y \sim \mathrm{Unif}(V)$ \COMMENT{valid $\succ$ invalid}
\ENDFOR
\STATE {\color{gray}\textit{$\backslash\backslash$ Cross-round pairs}}
\FORALL{valid $y^+ \in Y_2$}
    \STATE add $(q, y^+, y^-)$ with $y^- \sim \mathrm{Unif}\big(\{\, y \in V \cap Y_1 : r(y) < r(y^+) \,\}\big)$ \COMMENT{skip if empty}
\ENDFOR
\RETURN $\mathcal{B}_q$
\end{algorithmic}
\end{algorithm}

\section{Reflective Self-Distillation: Derivation and Margin Analysis}
\label{app:rsd}

This appendix expands Section~\ref{sec:method-rsd}. We make precise the sense in which preference optimization over the search buffer can \emph{rank} the reflective winners without making the policy more likely to \emph{generate} them, and then derive the RSD term as the sampled, stop-gradient form of a divergence between the hinted and unhinted policies. Throughout, $q$ is an unhinted operator prompt, $\tilde q=q\oplus z$ its insight-augmented counterpart (Section~\ref{sec:method-reflection}), and a reflective pair $(q,y^+,y^-)\in\mathcal{B}_{\mathrm{ref}}$ has a hinted winner $y^+\sim\pol(\cdot\mid\tilde q)$ and a strictly worse unhinted candidate $y^-\sim\pol(\cdot\mid q)$.

\paragraph{Ranking is not generation.}
Recall the implicit reward $\hat r_\theta(y)=\beta\log\frac{\pol(y\mid q)}{\polref(y\mid q)}$. The per-pair loss of Eq.~\eqref{eq:dpo} is a decreasing function of the margin
\begin{equation}
    \begin{aligned}
        &\Delta_\theta(q,y^+,y^-) = \hat r_\theta(y^+)-\hat r_\theta(y^-)\\
        &\quad = \beta\Big[\log\tfrac{\pol(y^+\mid q)}{\polref(y^+\mid q)} - \log\tfrac{\pol(y^-\mid q)}{\polref(y^-\mid q)}\Big]
    \end{aligned}
    \label{eq:app-margin}
\end{equation}
and enters the pair only through $\Delta_\theta$, so nothing in it rewards a high \emph{absolute} value of $\log\pol(y^+\mid q)$; a given increase of $\Delta_\theta$ is credited equally whether it raises $\log\pol(y^+\mid q)$ or lowers $\log\pol(y^-\mid q)$. Because a log-probability is bounded above by $0$ but unbounded below, the margin can be widened far more cheaply by driving $\log\pol(y^-\mid q)$ down than by lifting the already-high $\log\pol(y^+\mid q)$, and gradient descent does so preferentially, leaving $\log\pol(y^+\mid q)$ near its reference value $\log\polref(y^+\mid q)$. Two consequences specialize this to our setting. First, the reflective pairs are a minority of the buffer and have a \emph{small} margin by construction: $y^+$ and $y^-$ are both valid heuristics that differ only by the hint's refinement, so $\lvert\Delta_\theta\rvert$ is small relative to the within-batch and valid-versus-invalid pairs, whose rejected members are far weaker, and the reflective signal is easily dominated. Second, even when the reflective margin is realized it need not raise $\log\pol(y^+\mid q)$, the quantity that governs how likely the \emph{unhinted} policy is to generate $y^+$ in the next iteration. Ranking $y^+$ above $y^-$ and generating $y^+$ from $q$ are thus distinct objectives, and Eq.~\eqref{eq:dpo} targets only the former.

\paragraph{The hinted--unhinted divergence.}
The objective $\mathcal{L}_{\mathrm{RSD}}$ of Eq.~\eqref{eq:rsd-kl} is the forward Kullback--Leibler divergence between the hinted policy $\pol(\cdot\mid\tilde q)$, which the insight $z$ biases toward better heuristics, and the unhinted policy $\pol(\cdot\mid q)$ the search rolls out from. It vanishes when the hint changes nothing and grows as the hint moves the policy, so minimizing it pulls the unhinted policy toward the hinted one and the reflective improvement carries into the policy's own Round-1 rollouts.

\paragraph{From the divergence to a likelihood.}
The divergence averages over the open-ended space of heuristics and has no closed form, so we estimate it by sampling. Expanding it,
\begin{equation}
    \begin{aligned}
        &\mathrm{KL}\!\big(\pol(\cdot\mid\tilde q)\,\big\|\,\pol(\cdot\mid q)\big)\\
        &\quad = \mathbb{E}_{y\sim\pol(\cdot\mid\tilde q)}\big[\log\pol(y\mid\tilde q)\\
        &\quad \phantom{{}= \mathbb{E}_{y\sim\pol(\cdot\mid\tilde q)}\big[}{}-\log\pol(y\mid q)\big].
    \end{aligned}
    \label{eq:app-klexpand}
\end{equation}
Two properties of our pipeline reduce this to a single likelihood. First, the Round-2 winner $y^+$ is a draw from $\pol(\cdot\mid\tilde q)$, the sampling distribution of Eq.~\eqref{eq:app-klexpand}, so it serves as a Monte Carlo sample at no extra cost; because $\mathcal{B}_{\mathrm{ref}}$ retains only winners whose gap certifies a reflective gain, the resulting term is a reward-filtered surrogate of the divergence, deliberately biased toward the region the hint improved. Second, treating the hinted policy as a fixed teacher, a stop-gradient on $\pol(\cdot\mid\tilde q)$ that is standard in distillation, makes $\log\pol(y^+\mid\tilde q)$ a constant in $\theta$ that drops from the gradient. What remains is
\begin{equation}
    \nabla_\theta\,\widehat{\mathcal{L}}_{\mathrm{RSD}} = -\,\nabla_\theta\log\pol(y^+\mid q),
    \label{eq:app-rsdgrad}
\end{equation}
the gradient of the negative log-likelihood of the reflective winner under the unhinted prompt. The RSD term $-\lambda\log\pol(y^+\mid q)$ of Eq.~\eqref{eq:rsd} is therefore not a separate objective added to DPO but the reward-filtered, sampled, stop-gradient realization of the divergence in Eq.~\eqref{eq:rsd-kl}, evaluated on samples the search has already produced. The same forward pass DPO uses to score $y^+$ already yields $\log\pol(y^+\mid q)$, so the term adds essentially no compute.

\paragraph{Why this accumulates.}
A related distillation term has been motivated by inference-time hint independence, dropping the teacher at deployment~\cite{zhao2025rpo}. That reading does not apply here, since EPOCH deploys saved heuristic \emph{artifacts} rather than the policy. Our motivation is internal to training. Every iteration draws its Round-1 rollouts from the unhinted policy $\pol(\cdot\mid q)$, so raising $\log\pol(y^+\mid q)$ makes those rollouts more likely to recover the reflective improvement unprompted, which in turn seeds better parents, preference pairs, and experts in subsequent iterations. Because search and policy update interleave, an improvement internalized in one burst propagates through the population it generates, so the effect accumulates over the run rather than acting only at its end.

\paragraph{Implementation.}
We restrict RSD to $\mathcal{B}_{\mathrm{ref}}$ because each of its pairs contains, by construction, a valid Round-2 winner over a strictly worse valid Round-1 rollout, so the term distills genuine reflective gains rather than arbitrary accepted code. We length-normalize $\log\pol(y^+\mid q)$ by the token length of $y^+$ so the objective does not favor short heuristics, and we keep $\lambda$ small. With few reflective samples per burst, an unnormalized or large-$\lambda$ term behaves like maximum-likelihood concentration onto a handful of accepted heuristics and can over-sharpen the policy, eroding the output diversity that the coverage objective depends on. Setting $\lambda=0$ recovers the plain-DPO update of Eq.~\eqref{eq:dpo}, which makes the contribution of RSD a single-coefficient ablation.

\section{Extended Experiments}
\label{app:extended-exp}

\subsection{Extended Experiment Results}
\label{app:full-results}

Table~\ref{tab:tsp-full} extends Table~\ref{tab:tsp} with the greedy-edge~\cite{bentley1992fast}, nearest-insertion~\cite{rosenkrantz1977analysis}, and Christofides~\cite{christofides1976worst} classical heuristics and the Qwen2.5-7B backbone; Table~\ref{tab:obp-full} extends Table~\ref{tab:obp} with the Qwen2.5-7B backbone. Both tables also include the inference-only baselines run with the closed-source \texttt{GPT-5-nano} as the generator. On OBP, EPOCH attains a lower average gap than each of them on both backbones at both top-1 and top-4. On TSP, EPOCH keeps the lower top-1 gap, and only EoH-S matches its top-10 ensemble, with a lower gap at $n\in\{20,200\}$ and a slightly higher one at $n\in\{50,100\}$. These runs make every one of their $1{,}000$ heuristic generations with \texttt{GPT-5-nano}, which EPOCH queries only as a critic, and cost $14$--$29\times$ more per run in API calls (Appendix~\ref{app:cost}).

Table~\ref{tab:kp} reports the KP results with the Llama-3.1-8B backbone against ReEvo and MCTS-AHD, whose released implementations include this problem. EPOCH attains the best or a comparable top-1 gap at every size, its top-3 ensemble outperforms the top-10 ensemble of ReEvo at every size, and its top-10 ensemble cuts the gap by $26$--$47\%$ relative to MCTS-AHD and $55$--$59\%$ relative to ReEvo, with the gains persisting OOD at $n\in\{50,200\}$.

\subsection{Robustness to Seed Quality}
Several LLM-based AHD methods begin their search from a fixed \emph{seed heuristic}. ReEvo and HSEvo require one to initialize the population and to sustain variation among candidates, since a population that has converged to a single objective value cannot improve further and the search halts; CALM and EPOCH likewise start from a provided heuristic. To keep the comparison controlled, every method that requires a seed is initialized from the same heuristic, adopted from the MCTS-AHD experimental setup. The TSP seed, shown in the box below, is structurally detailed, scoring each candidate node by its distance to the current node, the mean and dispersion of its distances to the remaining nodes, and its distance to the destination.

\begin{seedbox}{Standard TSP seed heuristic}
import numpy as np

def select_next_node(current_node: int,
                     destination_node: int,
                     unvisited_nodes: set,
                     distance_matrix: np.ndarray) -> int:
    c1, c2, c3, c4 = 0.4, 0.3, 0.2, 0.1
    scores = {}
    for node in unvisited_nodes:
        all_distances = [distance_matrix[node][i]
                         for i in unvisited_nodes
                         if i != node]
        average_distance = np.mean(all_distances)
        std_dev_distance = np.std(all_distances)
        score = (
            c1 * distance_matrix[current_node][node]
            - c2 * average_distance
            + c3 * std_dev_distance
            - c4 * distance_matrix[destination_node][node]
        )
        scores[node] = score
    return min(scores, key=scores.get)
\end{seedbox}

The OBP seed, shown in the box below, is a single priority rule shared by every seed-requiring method; it is structurally detailed yet weak, scoring below the First-Fit and Best-Fit rules of Table~\ref{tab:obp-full}. It returns a priority score per open bin, and the framework places the item in the highest-scoring bin.

\begin{seedbox}{OBP seed heuristic}
import numpy as np

def step(item_size: float,
         remaining_capacity: np.ndarray) -> np.ndarray:
    max_bin_cap = max(remaining_capacity)
    score = ((remaining_capacity - max_bin_cap) ** 2
             / item_size
             + remaining_capacity ** 2 / item_size ** 2
             + remaining_capacity ** 2 / item_size ** 3)
    over = remaining_capacity > item_size
    score[over] = -score[over]
    score[1:] -= score[:-1]
    return score
\end{seedbox}

The KP seed, shown in the box below, packs the first remaining item that fits; it is the First-Fit row of Table~\ref{tab:kp}.

\begin{seedbox}{KP seed heuristic}
import numpy as np

def select_next_item(remaining_capacity: float,
                     weights: np.ndarray,
                     values: np.ndarray) -> int:
    fits = np.flatnonzero(weights <= remaining_capacity)
    return int(fits[0])
\end{seedbox}

A capable seed narrows the margin a method must close and can reward a search that exploits the seed over one that improves the generator. To test whether the advantage of EPOCH persists from a weaker starting point, we re-run the two methods that fine-tune the generator, CALM and EPOCH, from the substantially weaker TSP seed shown in the box below, which assigns every unvisited node the same constant score and therefore selects the next node without any problem-specific preference. All other settings match the TSP protocol of Section~\ref{sec:experiments}.

\begin{seedbox}{Weak TSP seed heuristic (robustness study)}
import numpy as np

def select_next_node(current_node: int,
                     destination_node: int,
                     unvisited_nodes: set,
                     distance_matrix: np.ndarray) -> int:
    scores = {node: 1 for node in unvisited_nodes}
    return min(scores, key=scores.get)
\end{seedbox}

\begin{table*}[tp]
\centering
\scriptsize
\setlength{\tabcolsep}{6.6mm}
\setlength{\extrarowheight}{0pt}
\renewcommand{\arraystretch}{1.0}
\resizebox{\textwidth}{!}{
\begin{tabular}{ll cccc}
\toprule
Method & & TSP-20 & \underline{TSP-50} & TSP-100 & TSP-200 \\
\midrule
\multicolumn{6}{c}{\textit{Classical heuristics}} \\
Nearest Neighbor  & & 17.64 & 21.96 & 24.58 & 26.10 \\
Greedy Edge       & & 11.96 & 17.41 & 18.69 & 17.89 \\
Nearest Insertion & & 13.14 & 19.38 & 21.55 & 23.69 \\
Christofides      & &  8.82 & 11.17 & 11.95 & 12.41 \\
\midrule
\multicolumn{6}{c}{\textit{LLM-based AHD: GPT-4o-mini (closed-source)}} \\
\multirow{3}{*}{ReEvo}    & Top-1  & \val{10.55}{3.58} & \val{13.89}{4.48} & \val{15.43}{4.77} & \val{16.64}{5.00} \\
                          & Top-3  & \val{9.18}{2.55}  & \val{12.12}{2.66} & \val{13.66}{3.29} & \val{17.01}{4.28} \\
                          & Top-10 & \val{8.49}{2.52}  & \val{10.92}{2.29} & \val{12.48}{2.70} & \val{14.29}{2.98} \\
\rowsep{6}
\multirow{3}{*}{HSEvo}    & Top-1  & \val{12.58}{4.18} & \val{16.37}{4.62} & \val{18.56}{4.94} & \val{20.86}{5.30} \\
                          & Top-3  & \val{11.38}{5.16} & \val{15.12}{5.64} & \val{17.39}{5.91} & \val{20.14}{5.98} \\
                          & Top-10 & \val{10.59}{5.81} & \val{14.28}{6.32} & \val{16.66}{6.50} & \val{19.65}{6.48} \\
\rowsep{6}
\multirow{3}{*}{MCTS-AHD} & Top-1  & \vals{7.28}{0.71} & \vals{10.77}{0.71} & \vals{12.82}{0.67} & \vals{14.75}{1.03} \\
                          & Top-3  & \val{6.25}{1.07} & \val{9.50}{1.08}  & \vals{11.77}{1.14} & \val{14.00}{1.18} \\
                          & Top-10 & \val{4.54}{1.20} & \val{7.94}{1.17}  & \val{10.49}{1.25} & \val{12.85}{1.41} \\
\rowsep{6}
\multirow{3}{*}{EoH-S}    & Top-1  & \val{8.56}{0.91} & \val{11.69}{0.71} & \val{13.80}{0.91} & \val{15.05}{0.98} \\
                          & Top-3  & \vals{5.05}{0.43} & \vals{9.14}{0.61} & \val{11.80}{0.83} & \vals{13.75}{0.85} \\
                          & Top-10 & \vals{3.22}{0.41} & \vals{7.30}{0.38} & \vals{10.06}{0.38} & \vals{12.39}{0.42} \\
\midrule
\multicolumn{6}{c}{\textit{LLM-based AHD: GPT-5-nano (closed-source)}} \\
\multirow{3}{*}{ReEvo}    & Top-1  & \val{12.32}{5.16} & \val{16.11}{5.41} & \val{18.10}{5.87} & \val{19.84}{6.08} \\
                          & Top-3  & \val{12.32}{5.16} & \val{16.11}{5.41} & \val{18.10}{5.87} & \val{19.84}{6.08} \\
                          & Top-10 & \val{12.32}{5.16} & \val{16.11}{5.41} & \val{18.10}{5.87} & \val{19.84}{6.08} \\
\rowsep{6}
\multirow{3}{*}{HSEvo}    & Top-1  & \val{8.44}{1.05} & \val{11.90}{1.20} & \val{13.54}{1.07} & \val{14.73}{1.17} \\
                          & Top-3  & \val{8.21}{1.08} & \val{11.64}{1.18} & \val{13.22}{1.09} & \val{14.39}{1.22} \\
                          & Top-10 & \val{8.10}{1.13} & \val{11.48}{1.24} & \val{13.05}{1.17} & \val{14.24}{1.30} \\
\rowsep{6}
\multirow{3}{*}{MCTS-AHD} & Top-1  & \val{7.23}{0.39} & \val{10.24}{0.38} & \val{12.13}{0.49} & \vals{13.77}{1.02} \\
                          & Top-3  & \val{5.64}{1.22} & \val{8.67}{0.84}  & \val{10.66}{0.85} & \val{12.51}{1.07} \\
                          & Top-10 & \val{3.78}{0.30} & \val{7.11}{0.34}  & \val{9.32}{0.58}  & \val{11.64}{0.88} \\
\rowsep{6}
\multirow{3}{*}{EoH-S}    & Top-1  & \vals{6.47}{0.71} & \vals{10.04}{0.37} & \vals{12.04}{0.48} & \val{13.94}{0.86} \\
                          & Top-3  & \vals{4.13}{0.21} & \vals{7.69}{0.35}  & \vals{10.07}{0.56} & \vals{12.25}{0.81} \\
                          & Top-10 & \valsb{2.48}{0.15} & \vals{5.78}{0.22}  & \vals{8.23}{0.33}  & \valsb{10.46}{0.33} \\
\midrule
\multicolumn{6}{c}{\textit{LLM-based AHD: Qwen2.5-7B}} \\
\multirow{3}{*}{ReEvo}    & Top-1  & \val{7.83}{2.07} & \val{11.10}{2.02} & \val{12.99}{2.10} & \val{14.43}{2.40} \\
                          & Top-3  & \val{7.25}{1.55} & \val{10.48}{1.59} & \val{12.32}{1.64} & \val{13.81}{2.06} \\
                          & Top-10 & \val{7.23}{1.58} & \val{10.47}{1.60} & \val{12.30}{1.66} & \val{13.78}{2.11} \\
\rowsep{6}
\multirow{3}{*}{HSEvo}    & Top-1  & \val{12.27}{3.27} & \val{16.41}{3.78} & \val{19.17}{3.78} & \val{22.50}{3.60} \\
                          & Top-3  & \val{12.12}{3.50} & \val{16.30}{3.92} & \val{19.08}{3.88} & \val{22.50}{3.60} \\
                          & Top-10 & \val{11.16}{3.00} & \val{15.42}{3.58} & \val{18.22}{3.55} & \val{21.53}{3.14} \\
\rowsep{6}
\multirow{3}{*}{MCTS-AHD} & Top-1  & \val{8.42}{0.66} & \val{12.00}{0.16} & \val{14.25}{0.49} & \val{15.93}{0.89} \\
                          & Top-3  & \val{7.83}{0.58} & \val{11.34}{0.54} & \val{13.63}{0.92} & \val{15.19}{1.32} \\
                          & Top-10 & \val{6.04}{1.40} & \val{9.75}{1.32}  & \val{12.18}{1.29} & \val{14.07}{1.31} \\
\rowsep{6}
\multirow{3}{*}{EoH-S}    & Top-1  & \val{8.64}{1.03} & \val{12.25}{1.26} & \val{14.44}{1.84} & \val{16.73}{2.50} \\
                          & Top-3  & \val{5.43}{1.14} & \val{9.12}{1.05} & \val{11.76}{1.45} & \val{14.43}{2.37} \\
                          & Top-10 & \val{3.46}{0.66} & \val{7.08}{0.76} & \val{9.97}{1.01} & \val{12.69}{1.34} \\
\rowsep{6}
\multirow{3}{*}{CALM}     & Top-1  & \val{6.75}{0.08} & \val{9.99}{0.22} & \val{11.96}{0.20} & \val{13.96}{0.20} \\
                          & Top-3  & \val{6.70}{0.14} & \val{9.92}{0.23} & \val{11.90}{0.21} & \val{13.91}{0.27} \\
                          & Top-10 & \val{6.30}{0.50} & \val{9.59}{0.55} & \val{11.59}{0.42} & \val{13.77}{0.34} \\
\rowsep{6}
\multirow{3}{*}{\shortstack[l]{\textbf{EPOCH}\\(Ours)}} & Top-1  & \vals{6.42}{0.08} & \vals{9.81}{0.22} & \vals{11.70}{0.46} & \vals{13.67}{0.91} \\
                                    & Top-3  & \vals{4.39}{0.20} & \vals{7.62}{0.44} & \vals{10.06}{0.45} & \vals{12.25}{0.50} \\
                                    & Top-10 & \vals{2.91}{0.14} & \valsb{5.71}{0.31} & \vals{8.22}{0.37}  & \vals{10.93}{0.47} \\
\midrule
\multicolumn{6}{c}{\textit{LLM-based AHD: Llama-3.1-8B}} \\
\multirow{3}{*}{ReEvo}    & Top-1  & \val{10.58}{0.58} & \val{14.87}{1.36} & \val{18.38}{2.10} & \val{26.06}{12.93} \\
                          & Top-3  & \val{10.28}{0.59} & \val{14.15}{2.15} & \val{16.19}{1.65} & \val{19.05}{1.37} \\
                          & Top-10 & \val{10.25}{0.62} & \val{14.10}{2.24} & \val{15.96}{2.01} & \val{18.69}{1.54} \\
\rowsep{6}
\multirow{3}{*}{HSEvo}    & Top-1  & \val{11.46}{1.69} & \val{15.85}{2.63} & \val{18.48}{4.51} & \val{21.34}{7.00} \\
                          & Top-3  & \val{10.61}{1.28} & \val{14.83}{1.76} & \val{17.27}{2.82} & \val{19.47}{3.89} \\
                          & Top-10 & \val{10.29}{1.54} & \val{14.45}{2.10} & \val{16.80}{3.24} & \val{18.99}{4.31} \\
\rowsep{6}
\multirow{3}{*}{MCTS-AHD} & Top-1  & \val{10.27}{0.37} & \val{14.23}{0.65} & \val{16.39}{1.48} & \val{17.87}{1.45} \\
                          & Top-3  & \val{9.76}{0.40}  & \val{13.40}{0.39} & \val{15.40}{0.81} & \val{16.92}{1.04} \\
                          & Top-10 & \val{7.58}{0.63}  & \val{11.73}{0.61} & \val{13.77}{0.91} & \val{15.51}{1.15} \\
\rowsep{6}
\multirow{3}{*}{EoH-S}    & Top-1  & \val{10.69}{0.93} & \val{15.43}{1.58} & \val{17.86}{1.38} & \val{20.04}{1.46} \\
                          & Top-3  & \val{6.05}{0.72} & \val{11.68}{1.06} & \val{14.59}{0.93} & \val{17.31}{1.05} \\
                          & Top-10 & \val{4.70}{0.81} & \val{10.46}{1.02} & \val{13.60}{0.96} & \val{16.45}{1.07} \\
\rowsep{6}
\multirow{3}{*}{CALM}     & Top-1  & \val{6.76}{0.26} & \val{10.02}{0.26} & \val{11.75}{0.27} & \vals{13.47}{0.31} \\
                          & Top-3  & \val{6.55}{0.23} & \val{9.68}{0.55}  & \val{11.34}{0.51} & \val{13.33}{0.47} \\
                          & Top-10 & \val{6.19}{0.54} & \val{9.31}{0.80}  & \val{11.04}{0.67} & \val{13.30}{0.91} \\
\rowsep{6}
\multirow{3}{*}{\shortstack[l]{\textbf{EPOCH}\\(Ours)}} & Top-1  & \vals{6.19}{0.30} & \vals{9.61}{0.25} & \vals{11.63}{0.21} & \val{13.57}{0.36} \\
                                    & Top-3  & \vals{4.31}{0.30} & \vals{7.62}{0.23} & \vals{9.85}{0.36}  & \vals{11.98}{0.49} \\
                                    & Top-10 & \vals{2.81}{0.22} & \vals{5.73}{0.15}  & \valsb{8.15}{0.19}  & \vals{10.74}{0.38} \\
\bottomrule
\end{tabular}
}
\caption{Complete version of Table~\ref{tab:tsp}. TSP optimality gap (\%) to LKH-3 optima; each cell is the mean over runs with standard deviation, lower is better. Size $n{=}50$ is ID (underlined); $n\in\{20,100,200\}$ are OOD. Within each backbone block the best method per metric and test set is \colorbox{shadegray}{shaded}; each test set's overall best result is shown in \textbf{bold}.}
\label{tab:tsp-full}
\end{table*}

\begin{table*}[tp]
\centering
\scriptsize
\setlength{\tabcolsep}{6pt}
\setlength{\extrarowheight}{0pt}
\renewcommand{\arraystretch}{1.2}
\resizebox{\textwidth}{!}{
\begin{tabular}{ll cccccc|c}
\toprule
Method & & \underline{1k\_100} & \underline{1k\_500} & \underline{5k\_100} & \underline{5k\_500} & 10k\_100 & 10k\_500 & Avg. \\
\midrule
\multicolumn{8}{c|}{\textit{Classical heuristics}} & \multicolumn{1}{c}{} \\
First-Fit & & 5.02 & 0.25 & 4.65 & 0.55 & 4.36 & 0.50 & 2.55 \\
Best-Fit  & & 4.77 & 0.25 & 4.31 & 0.55 & 4.05 & 0.47 & 2.40 \\
\midrule
\multicolumn{8}{c|}{\textit{LLM-based AHD: GPT-4o-mini (closed-source)}} & \multicolumn{1}{c}{} \\
\multirow{2}{*}{ReEvo}    & Top-1 & \val{3.74}{0.94} & \val{0.40}{0.30} & \val{3.09}{1.12} & \val{0.69}{0.30} & \val{2.84}{1.11} & \val{0.53}{0.12} & \val{1.88}{0.46} \\
                          & Top-4 & \val{3.61}{1.01} & \val{0.40}{0.30} & \val{2.98}{1.21} & \val{0.56}{0.13} & \val{2.74}{1.22} & \val{0.44}{0.03} & \val{1.79}{0.53} \\
\rowsep{9}
\multirow{2}{*}{HSEvo}    & Top-1 & \val{2.85}{1.01} & \val{0.40}{0.30} & \val{2.19}{1.15} & \val{0.56}{0.02} & \val{2.01}{1.15} & \val{0.50}{0.04} & \val{1.42}{0.52} \\
                          & Top-4 & \val{2.82}{1.01} & \val{0.20}{0.10} & \val{2.17}{1.17} & \val{0.47}{0.12} & \val{2.01}{1.15} & \val{0.42}{0.10} & \val{1.35}{0.56} \\
\rowsep{9}
\multirow{2}{*}{MCTS-AHD} & Top-1 & \val{2.59}{0.45} & \val{0.15}{0.12} & \val{1.88}{0.47} & \val{0.42}{0.17} & \val{1.71}{0.41} & \val{0.38}{0.12} & \val{1.19}{0.27} \\
                          & Top-4 & \val{2.48}{0.51} & \val{0.10}{0.12} & \val{1.79}{0.53} & \val{0.36}{0.17} & \val{1.61}{0.48} & \val{0.33}{0.13} & \val{1.11}{0.31} \\
\rowsep{9}
\multirow{2}{*}{EoH-S}    & Top-1 & \val{2.62}{0.33} & \val{0.25}{0.16} & \val{1.74}{0.73} & \val{0.42}{0.15} & \val{1.46}{0.83} & \val{0.35}{0.13} & \vals{1.14}{0.33} \\
                          & Top-4 & \val{2.40}{0.38} & \val{0.15}{0.12} & \val{1.41}{0.56} & \val{0.24}{0.16} & \val{1.08}{0.58} & \val{0.18}{0.15} & \vals{0.91}{0.30} \\
\midrule
\multicolumn{8}{c|}{\textit{LLM-based AHD: GPT-5-nano (closed-source)}} & \multicolumn{1}{c}{} \\
\multirow{2}{*}{ReEvo}    & Top-1 & \val{3.58}{1.24} & \val{0.50}{0.25} & \val{2.91}{1.32} & \val{0.62}{0.07} & \val{2.73}{1.37} & \val{0.54}{0.06} & \val{1.81}{0.72} \\
                          & Top-4 & \val{3.58}{1.24} & \val{0.37}{0.13} & \val{2.90}{1.31} & \val{0.57}{0.02} & \val{2.72}{1.35} & \val{0.47}{0.00} & \val{1.77}{0.68} \\
\rowsep{9}
\multirow{2}{*}{HSEvo}    & Top-1 & \val{4.56}{0.15} & \val{0.25}{0.00} & \val{4.04}{0.22} & \val{0.55}{0.00} & \val{3.89}{0.13} & \val{0.47}{0.01} & \val{2.29}{0.08} \\
                          & Top-4 & \val{4.56}{0.15} & \val{0.25}{0.00} & \val{4.04}{0.22} & \val{0.55}{0.00} & \val{3.88}{0.13} & \val{0.47}{0.01} & \val{2.29}{0.08} \\
\rowsep{9}
\multirow{2}{*}{MCTS-AHD} & Top-1 & \val{3.78}{1.15} & \val{0.25}{0.00} & \val{3.06}{1.44} & \val{0.55}{0.00} & \val{2.81}{1.46} & \val{0.48}{0.02} & \val{1.82}{0.67} \\
                          & Top-4 & \val{3.73}{1.19} & \val{0.20}{0.10} & \val{3.04}{1.43} & \val{0.48}{0.09} & \val{2.79}{1.45} & \val{0.43}{0.08} & \val{1.78}{0.71} \\
\rowsep{9}
\multirow{2}{*}{EoH-S}    & Top-1 & \val{3.22}{0.83} & \val{0.53}{0.81} & \val{1.84}{0.83} & \val{0.48}{0.18} & \val{1.53}{0.86} & \val{0.41}{0.09} & \vals{1.33}{0.51} \\
                          & Top-4 & \val{2.84}{0.85} & \val{0.28}{0.50} & \val{1.39}{0.70} & \val{0.36}{0.17} & \val{0.94}{0.70} & \val{0.31}{0.16} & \vals{1.02}{0.42} \\
\midrule
\multicolumn{8}{c|}{\textit{LLM-based AHD: Qwen2.5-7B}} & \multicolumn{1}{c}{} \\
\multirow{2}{*}{ReEvo}    & Top-1 & \val{4.53}{0.68} & \val{0.25}{0.00} & \val{3.29}{1.29} & \val{0.49}{0.12} & \val{3.00}{1.37} & \val{0.44}{0.13} & \val{2.00}{0.55} \\
                          & Top-4 & \val{4.47}{0.65} & \val{0.20}{0.10} & \val{3.25}{1.26} & \val{0.48}{0.12} & \val{2.98}{1.35} & \val{0.41}{0.12} & \val{1.97}{0.52} \\
\rowsep{9}
\multirow{2}{*}{HSEvo}    & Top-1 & \val{4.72}{0.10} & \val{0.25}{0.00} & \val{4.24}{0.10} & \val{0.55}{0.00} & \val{4.02}{0.05} & \val{0.47}{0.00} & \val{2.38}{0.04} \\
                          & Top-4 & \val{4.71}{0.12} & \val{0.25}{0.00} & \val{4.24}{0.10} & \val{0.55}{0.00} & \val{4.02}{0.06} & \val{0.47}{0.00} & \val{2.37}{0.05} \\
\rowsep{9}
\multirow{2}{*}{MCTS-AHD} & Top-1 & \val{4.71}{0.14} & \val{0.25}{0.00} & \val{4.13}{0.21} & \val{0.55}{0.00} & \val{3.95}{0.22} & \val{0.47}{0.00} & \val{2.34}{0.09} \\
                          & Top-4 & \val{4.64}{0.08} & \val{0.25}{0.00} & \val{4.12}{0.21} & \val{0.55}{0.00} & \val{3.93}{0.21} & \val{0.47}{0.00} & \val{2.33}{0.08} \\
\rowsep{9}
\multirow{2}{*}{EoH-S}    & Top-1 & \val{3.70}{1.10} & \val{0.49}{0.29} & \val{2.89}{1.21} & \val{0.52}{0.12} & \val{2.71}{1.21} & \val{0.44}{0.12} & \val{1.79}{0.57} \\
                          & Top-4 & \val{3.37}{1.10} & \val{0.25}{0.25} & \val{2.31}{1.35} & \val{0.39}{0.12} & \val{2.04}{1.43} & \val{0.39}{0.12} & \val{1.46}{0.64} \\
\rowsep{9}
\multirow{2}{*}{CALM}     & Top-1 & \val{4.17}{0.54} & \val{0.35}{0.12} & \val{2.89}{1.18} & \val{0.45}{0.15} & \val{2.55}{1.33} & \val{0.35}{0.15} & \val{1.79}{0.51} \\
                          & Top-4 & \val{4.09}{0.55} & \val{0.30}{0.10} & \val{2.82}{1.28} & \val{0.45}{0.15} & \val{2.50}{1.41} & \val{0.35}{0.16} & \val{1.75}{0.56} \\
\rowsep{9}
\multirow{2}{*}{\textbf{EPOCH (Ours)}} & Top-1 & \val{3.91}{0.66} & \val{0.25}{0.22} & \val{1.54}{0.46} & \val{0.17}{0.09} & \val{1.27}{0.66} & \val{0.19}{0.14} & \vals{1.22}{0.20} \\
                          & Top-4 & \val{3.19}{0.10} & \val{0.15}{0.20} & \val{0.79}{0.03} & \val{0.17}{0.09} & \valb{0.30}{0.02} & \val{0.15}{0.07} & \vals{0.79}{0.05} \\
\midrule
\multicolumn{8}{c|}{\textit{LLM-based AHD: Llama-3.1-8B}} & \multicolumn{1}{c}{} \\
\multirow{2}{*}{ReEvo}    & Top-1 & \val{4.34}{0.70} & \val{0.25}{0.00} & \val{3.27}{1.25} & \val{0.41}{0.17} & \val{3.04}{1.36} & \val{0.36}{0.16} & \val{1.95}{0.60} \\
                          & Top-4 & \val{4.10}{0.87} & \val{0.20}{0.10} & \val{3.23}{1.24} & \val{0.41}{0.17} & \val{2.98}{1.34} & \val{0.35}{0.16} & \val{1.88}{0.64} \\
\rowsep{9}
\multirow{2}{*}{HSEvo}    & Top-1 & \val{3.39}{0.93} & \val{0.25}{0.00} & \val{2.42}{1.26} & \val{0.42}{0.18} & \val{2.11}{1.33} & \val{0.36}{0.15} & \val{1.49}{0.58} \\
                          & Top-4 & \val{3.36}{0.97} & \val{0.25}{0.00} & \val{2.42}{1.26} & \val{0.40}{0.16} & \val{2.10}{1.33} & \val{0.36}{0.15} & \val{1.48}{0.59} \\
\rowsep{9}
\multirow{2}{*}{MCTS-AHD} & Top-1 & \val{4.14}{0.89} & \val{0.17}{0.12} & \val{3.54}{1.09} & \val{0.46}{0.12} & \val{3.30}{1.07} & \val{0.40}{0.11} & \val{2.00}{0.57} \\
                          & Top-4 & \val{4.11}{0.94} & \val{0.17}{0.12} & \val{3.54}{1.09} & \val{0.46}{0.12} & \val{3.30}{1.07} & \val{0.40}{0.11} & \val{1.99}{0.57} \\
\rowsep{9}
\multirow{2}{*}{EoH-S}    & Top-1 & \val{3.92}{0.78} & \val{0.21}{0.09} & \val{2.89}{0.87} & \val{0.51}{0.07} & \val{2.59}{1.00} & \val{0.45}{0.05} & \val{1.76}{0.38} \\
                          & Top-4 & \val{3.56}{0.73} & \valb{0.08}{0.12} & \val{1.46}{0.57} & \valb{0.16}{0.09} & \val{1.14}{0.69} & \val{0.14}{0.10} & \val{1.09}{0.27} \\
\rowsep{9}
\multirow{2}{*}{CALM}     & Top-1 & \val{3.71}{0.52} & \val{0.59}{0.82} & \val{1.32}{0.40} & \val{0.30}{0.14} & \val{0.82}{0.41} & \val{0.27}{0.14} & \val{1.17}{0.22} \\
                          & Top-4 & \val{3.69}{0.51} & \val{0.59}{0.82} & \val{1.29}{0.43} & \val{0.29}{0.15} & \val{0.80}{0.43} & \val{0.26}{0.13} & \val{1.15}{0.23} \\
\rowsep{9}
\multirow{2}{*}{\textbf{EPOCH (Ours)}} & Top-1 & \val{2.99}{0.70} & \val{0.30}{0.36} & \val{1.66}{0.90} & \val{0.32}{0.21} & \val{1.33}{0.97} & \val{0.28}{0.24} & \vals{1.15}{0.39} \\
                          & Top-4 & \valb{2.34}{0.43} & \val{0.10}{0.12} & \valb{0.74}{0.06} & \val{0.20}{0.07} & \val{0.36}{0.06} & \valb{0.13}{0.04} & \valsb{0.64}{0.07} \\
\bottomrule
\end{tabular}
}
\caption{Complete version of Table~\ref{tab:obp}. OBP optimality gap (\%) to the $L_1$ lower bound on the six held-out Weibull test sets; each cell is the mean over runs with standard deviation, lower is better. Columns are $(\text{items},\text{capacity})$, e.g.\ \texttt{1k\_100} is $1{,}000$ items at capacity $100$; \texttt{Avg.}\ is the mean over the six. Item scales $1\mathrm{k},5\mathrm{k}$ are ID (underlined); $10\mathrm{k}$ is OOD. Within each backbone block the best \texttt{Avg.}\ per metric is \colorbox{shadegray}{shaded}; each test set's overall best result is shown in \textbf{bold}.}
\label{tab:obp-full}
\end{table*}

\begin{table*}[tp]
\centering
\scriptsize
\setlength{\tabcolsep}{8mm}
\setlength{\extrarowheight}{0pt}
\renewcommand{\arraystretch}{1.05}
\resizebox{\textwidth}{!}{
\begin{tabular}{ll ccc}
\toprule
Method & & KP-50 & \underline{KP-100} & KP-200 \\
\midrule
\multicolumn{5}{c}{\textit{Classical heuristics}} \\
First-Fit (seed) & & 32.856 & 34.836 & 53.153 \\
\midrule
\multicolumn{5}{c}{\textit{LLM-based AHD: Llama-3.1-8B}} \\
\multirow{3}{*}{ReEvo}    & Top-1  & \val{0.224}{0.031} & \val{0.099}{0.016} & \val{0.107}{0.025} \\
                          & Top-3  & \val{0.198}{0.018} & \val{0.081}{0.006} & \val{0.089}{0.023} \\
                          & Top-10 & \val{0.170}{0.022} & \val{0.066}{0.005} & \val{0.071}{0.018} \\
\rowsep{5}
\multirow{3}{*}{MCTS-AHD} & Top-1  & \val{0.202}{0.013} & \val{0.090}{0.009} & \vals{0.077}{0.010} \\
                          & Top-3  & \val{0.187}{0.021} & \val{0.080}{0.016} & \val{0.063}{0.017} \\
                          & Top-10 & \val{0.131}{0.038} & \val{0.054}{0.012} & \val{0.039}{0.006} \\
\rowsep{5}
\multirow{3}{*}{\shortstack[l]{\textbf{EPOCH}\\(Ours)}} & Top-1  & \vals{0.188}{0.008} & \vals{0.089}{0.006} & \val{0.078}{0.010} \\
                                    & Top-3  & \vals{0.123}{0.027} & \vals{0.057}{0.007} & \vals{0.059}{0.015} \\
                                    & Top-10 & \valsb{0.070}{0.018} & \valsb{0.030}{0.005} & \valsb{0.029}{0.006} \\
\bottomrule
\end{tabular}
}
\caption{KP optimality gap (\%) to the optimum with the Llama-3.1-8B backbone; each cell is the mean over runs with standard deviation, lower is better. Size $n{=}100$ is ID (underlined); $n\in\{50,200\}$ are OOD. The best method per metric and test set is \colorbox{shadegray}{shaded}; each test set's overall best result is shown in \textbf{bold}.}
\label{tab:kp}
\end{table*}

As reported in Table~\ref{tab:seed-robust}, from the weak seed CALM degrades sharply on both backbones, and its top-10 gap at the training size $n{=}50$ rises to $14.51\%$ on Llama and $16.10\%$ on Qwen, against $9.31$ and $9.59\%$ from the standard seed in Table~\ref{tab:tsp-full}. EPOCH instead stays close to its standard-seed performance, with a top-10 gap of $6.71$ and $6.76\%$ at $n{=}50$ against $5.73$ and $5.71\%$ in Table~\ref{tab:tsp-full}, and it attains the lowest gap on every size, metric, and backbone. Its top-3 ensemble from the weak seed still outperforms the full top-10 ensemble of CALM throughout. Because the per-instance gap-vector objective steers selection, operator allocation, recovery, and the deployed ensemble toward coverage rather than the fitness of a single heuristic, EPOCH recovers a strong complementary set even from an uninformative seed, where a fitness-only method does not.

\begin{table*}[tp]
\centering
\scriptsize
\setlength{\tabcolsep}{6mm}
\setlength{\extrarowheight}{0pt}
\renewcommand{\arraystretch}{0.95}
\resizebox{\textwidth}{!}{
\begin{tabular}{ll cccc}
\toprule
Method & & TSP-20 & \underline{TSP-50} & TSP-100 & TSP-200 \\
\midrule
\multicolumn{6}{c}{\textit{LLM-based AHD: Qwen2.5-7B}} \\
\multirow{3}{*}{CALM} & Top-1  & \val{12.08}{3.56} & \val{16.70}{3.89} & \val{19.38}{4.21} & \val{21.39}{3.56} \\
                      & Top-3  & \val{12.08}{3.57} & \val{16.65}{3.89} & \val{19.32}{4.21} & \val{21.21}{3.56} \\
                      & Top-10 & \val{11.49}{3.56} & \val{16.10}{3.72} & \val{18.60}{3.93} & \val{20.48}{3.23} \\
\rowsep{6}
\multirow{3}{*}{\shortstack[l]{\textbf{EPOCH}\\(Ours)}} & Top-1  & \vals{7.97}{2.39} & \vals{10.85}{1.45} & \vals{12.82}{1.30} & \vals{14.64}{1.28} \\
                      & Top-3  & \vals{5.55}{2.01} & \vals{8.61}{1.59} & \vals{11.08}{1.71} & \vals{13.22}{1.53} \\
                      & Top-10 & \vals{3.88}{1.60} & \vals{6.76}{1.27} & \vals{9.23}{1.35} & \valsb{11.59}{1.39} \\
\midrule
\multicolumn{6}{c}{\textit{LLM-based AHD: Llama-3.1-8B}} \\
\multirow{3}{*}{CALM} & Top-1  & \val{11.30}{1.49} & \val{15.89}{2.65} & \val{17.81}{3.03} & \val{19.46}{3.12} \\
                      & Top-3  & \val{10.93}{1.66} & \val{15.41}{2.82} & \val{17.38}{3.18} & \val{19.05}{3.25} \\
                      & Top-10 & \val{9.91}{1.31} & \val{14.51}{2.63} & \val{16.40}{3.10} & \val{18.01}{3.15} \\
\rowsep{6}
\multirow{3}{*}{\shortstack[l]{\textbf{EPOCH}\\(Ours)}} & Top-1  & \vals{7.97}{1.44} & \vals{11.42}{1.40} & \vals{13.11}{1.17} & \vals{14.86}{1.00} \\
                      & Top-3  & \vals{5.31}{0.85} & \vals{8.82}{1.06} & \vals{11.24}{1.20} & \vals{13.53}{1.07} \\
                      & Top-10 & \valsb{3.33}{0.51} & \valsb{6.71}{0.62} & \valsb{9.12}{0.69} & \valsb{11.59}{0.73} \\
\bottomrule
\end{tabular}
}
\caption{TSP optimality gap (\%) to LKH-3 optima when the two methods that fine-tune the generator start from the weak seed; each cell is the mean over runs with standard deviation, lower is better. Size $n{=}50$ is ID (underlined); $n\in\{20,100,200\}$ are OOD. Within each backbone block the best method per metric and test set is \colorbox{shadegray}{shaded}; each test set's overall best result is shown in \textbf{bold}.}
\label{tab:seed-robust}
\end{table*}

\subsection{Search Dynamics \& Evolutionary Curves}
\begin{figure*}[tp]
\centering
\begin{minipage}{0.49\textwidth}\centering
  \includegraphics[width=\linewidth]{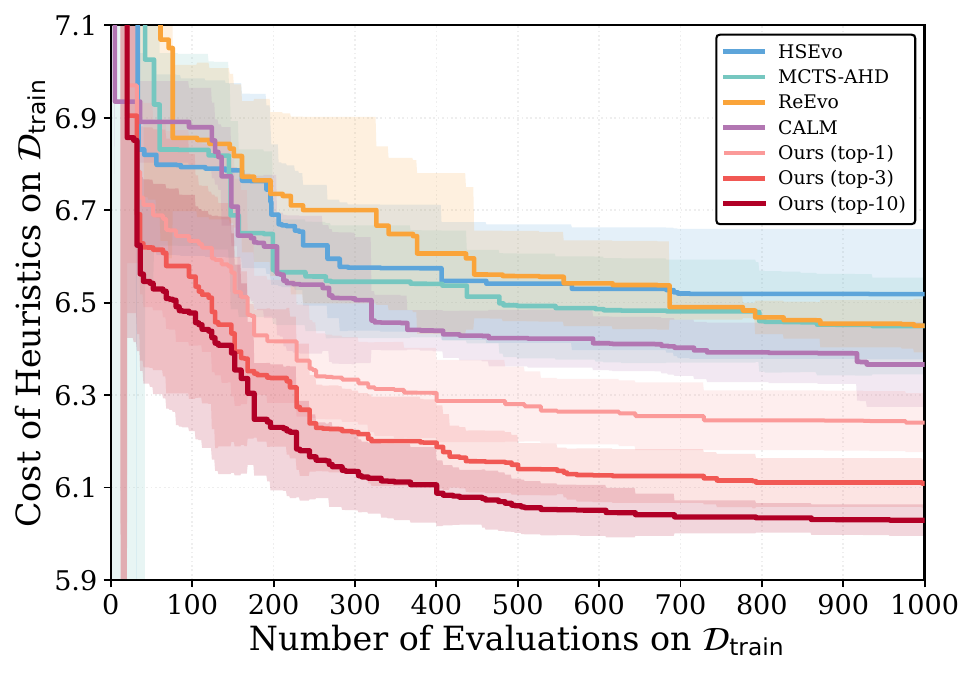}\\[2pt]
  (a) Llama-3.1-8B
\end{minipage}\hfill
\begin{minipage}{0.49\textwidth}\centering
  \includegraphics[width=\linewidth]{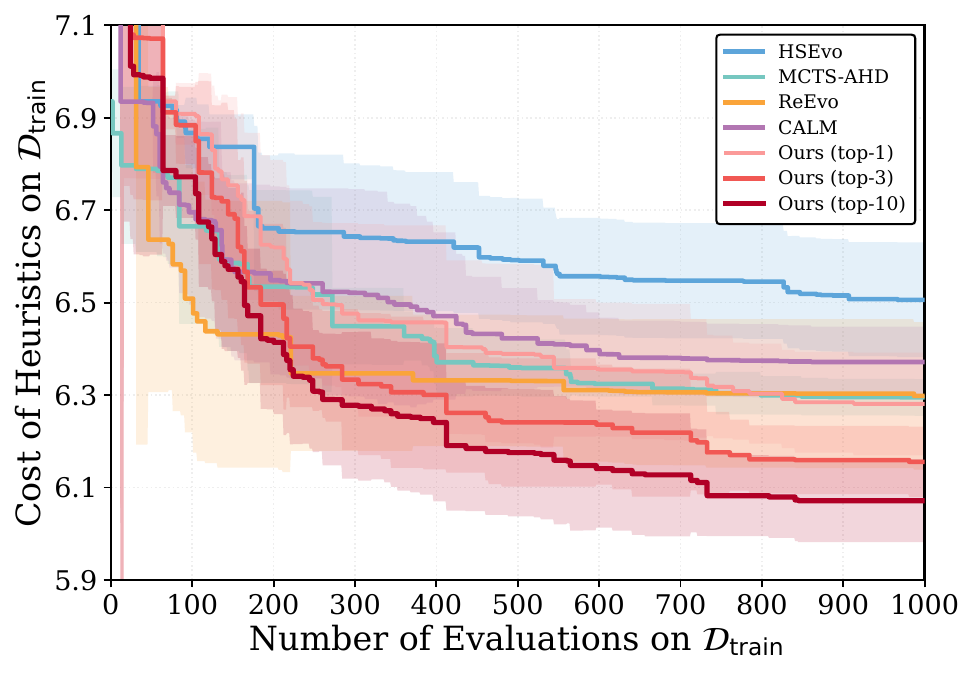}\\[2pt]
  (b) Qwen2.5-7B
\end{minipage}
\caption{Evolution curves on TSP from the weak seed for (a)~Llama-3.1-8B and (b)~Qwen2.5-7B. Each curve is the mean over runs with its standard deviation as the shaded band; EPOCH is drawn as its top-1, top-3, and top-10 ensembles.}
\label{fig:evolution}
\end{figure*}

To assess search efficiency and stability, Figure~\ref{fig:evolution} plots the best-so-far cost on $\mathcal{D}_{\mathrm{train}}$ against the number of heuristic evaluations, from the same weak seed and averaged over at least five runs of each method. EPOCH converges faster and to a lower cost than the LLM-based AHD baselines on both backbones, and its top-10 ensemble curve stays below every single-heuristic method across the entire budget, with low deviation across runs. This is the intended effect of optimizing per-instance coverage rather than the fitness of a single heuristic.

\subsection{Generalization to Real-World TSPLIB Instances}
\label{app:tsplib}
The experiments of Section~\ref{sec:experiments} train and test on random uniform instances. To test whether the discovered heuristics transfer to real, structured instances of varied size and scale, we evaluate on TSPLIB~\cite{reinelt1991tsplib}, using its $15$ instances with fewer than $500$ nodes. Coordinates are scaled to $[0,1]^2$ to match the training distribution, each tour is built from three start nodes and averaged, and the gap is taken against the known optimum. For every method we deploy the top-10 ensemble of its single best run, selected by the lowest mean top-10 gap over the four uniform test sizes. We compare against the manually designed greedy, nearest-insertion, and nearest-neighbor heuristics~\cite{rosenkrantz1977analysis} and the genetic-programming hyper-heuristic GPHH~\cite{gpahd}.

As reported in Table~\ref{tab:tsplib}, EPOCH attains the lowest average optimality gap and the best result on the majority of the TSPLIB instances.

\begin{table*}[tp]
\centering
\scriptsize
\setlength{\tabcolsep}{4pt}
\renewcommand{\arraystretch}{0.95}
\resizebox{\textwidth}{!}{
\begin{tabular}{l cccc cccccc}
\toprule
& \multicolumn{4}{c}{Manual \& GP heuristics} & \multicolumn{6}{c}{LLM-based AHD} \\
\cmidrule(lr){2-5}\cmidrule(lr){6-11}
Instance & Greedy & Near-ins. & Near-gr. & GPHH & ReEvo & HSEvo & MCTS-AHD & EoH-S & CALM & \textbf{EPOCH} \\
\midrule
ts225   & 5.38 & 19.93 & 16.82 & 7.71 & 7.36 & 6.18 & \cellcolor{shadegray}\textbf{3.56} & 3.69 & 5.78 & 5.68 \\
rat99   & 22.30 & 21.05 & 21.79 & 14.09 & 17.39 & 17.72 & 16.88 & 18.50 & 14.05 & \cellcolor{shadegray}\textbf{9.87} \\
bier127 & 19.50 & 23.05 & 23.25 & 15.64 & 20.94 & 17.26 & 17.53 & 11.24 & 12.18 & \cellcolor{shadegray}\textbf{6.33} \\
lin318  & 18.75 & 24.44 & 25.78 & 14.30 & 15.47 & 16.21 & 14.48 & 14.48 & 12.73 & \cellcolor{shadegray}\textbf{11.97} \\
eil51   & 13.03 & 16.14 & 31.96 & 10.20 & \cellcolor{shadegray}\textbf{6.65} & 8.42 & 7.26 & 7.56 & 7.53 & 6.83 \\
d493    & 16.68 & 20.39 & 24.00 & 15.58 & 17.81 & 14.37 & 15.33 & 15.54 & 12.96 & \cellcolor{shadegray}\textbf{11.37} \\
kroB100 & 16.59 & 21.53 & 26.26 & 14.06 & 16.88 & 12.52 & 12.51 & 14.20 & 12.40 & \cellcolor{shadegray}\textbf{10.38} \\
kroC100 & 12.94 & 24.25 & 25.76 & 16.22 & 16.14 & 15.55 & 15.75 & 13.82 & 15.05 & \cellcolor{shadegray}\textbf{10.97} \\
ch130   & 28.40 & 19.21 & 25.66 & 14.77 & 13.50 & 12.52 & 11.52 & 12.58 & 10.47 & \cellcolor{shadegray}\textbf{8.09} \\
pr299   & 31.42 & 25.05 & 31.42 & 18.24 & 20.68 & 20.95 & 19.55 & 19.66 & \cellcolor{shadegray}\textbf{17.17} & 18.07 \\
fl417   & \textbf{12.64} & 25.52 & 32.42 & 22.72 & 15.01 & 18.90 & 16.25 & 18.92 & 21.92 & \cellcolor{shadegray}13.67 \\
kroA150 & 20.24 & 19.09 & 26.08 & 15.59 & 15.33 & 19.87 & 15.12 & 16.19 & 12.75 & \cellcolor{shadegray}\textbf{9.10} \\
pr264   & \textbf{11.89} & 34.28 & 17.87 & 23.96 & 30.03 & 14.37 & 15.35 & \cellcolor{shadegray}12.78 & 19.67 & 15.87 \\
pr226   & 21.44 & 28.02 & 24.65 & 15.51 & 13.41 & \cellcolor{shadegray}\textbf{11.15} & 16.55 & 12.44 & 19.89 & 16.55 \\
pr439   & 20.08 & 24.67 & 27.36 & 21.36 & 27.07 & 23.80 & 18.91 & 19.53 & 17.01 & \cellcolor{shadegray}\textbf{13.01} \\
\midrule
Average & 18.09 & 23.11 & 25.41 & 16.00 & 16.91 & 15.32 & 14.44 & 14.08 & 14.11 & \cellcolor{shadegray}\textbf{11.18} \\
\bottomrule
\end{tabular}
}
\caption{Generalization to real-world TSPLIB instances~\cite{reinelt1991tsplib}, reported as the per-instance optimality gap (\%) to the known optimum, lower is better. The LLM-based methods use the Llama-3.1-8B backbone, and the manual and GP reference columns are single heuristics taken from~\citet{gpahd}. Per instance, the best LLM-based method is \colorbox{shadegray}{shaded} and the overall best is in \textbf{bold}.}
\label{tab:tsplib}
\end{table*}

\section{Significance Testing and Uncertainty Quantification}
\label{app:significance}

We test whether the advantage of EPOCH in Tables~\ref{tab:tsp-full} and~\ref{tab:obp-full} is statistically reliable along the two axes that matter for an LLM-driven search: the randomness of an entire run (LLM sampling and the stochastic search trajectory), and the spread of difficulty across test instances. The first is addressed by a significance test across runs, the second by a paired comparison across instances. Every method is run at least five times.

\subsection{Statistical Significance Across Runs}
We treat each independent run of a method as one observation, summarized by the mean optimality gap of its deployed ensemble on a test set: the top-10 ensemble for TSP and the top-4 ensemble for OBP. For each test set we test whether EPOCH reaches a lower mean gap than a given baseline with a one-sided two-sample $t$-test in the Welch form, which does not assume the two methods share a variance~\cite{welch1947}. The per-size $p$-values for TSP on both backbones are in Table~\ref{tab:sig-tsp}; those for OBP, where the test is run on the gap averaged over the six Weibull sets as each set holds only a few instances, are in Table~\ref{tab:sig-obp}. A $p$-value below $0.05$ indicates that the lower mean gap of EPOCH is unlikely to be an artifact of run-to-run noise. EPOCH attains a significantly lower mean gap than the baselines in nearly all settings.

\begin{table*}[tp]
\centering
\scriptsize
\setlength{\tabcolsep}{6pt}
\renewcommand{\arraystretch}{0.95}
\resizebox{\textwidth}{!}{
\begin{tabular}{l cccc cccc}
\toprule
& \multicolumn{4}{c}{Qwen2.5-7B} & \multicolumn{4}{c}{Llama-3.1-8B} \\
\cmidrule(lr){2-5}\cmidrule(lr){6-9}
Method & TSP-20 & TSP-50 & TSP-100 & TSP-200 & TSP-20 & TSP-50 & TSP-100 & TSP-200 \\
\midrule
ReEvo    & {\boldmath$0.003$} & {\boldmath$0.002$} & {\boldmath$0.003$} & {\boldmath$0.026$} & {\boldmath$2.7{\times}10^{-6}$} & {\boldmath$8.5{\times}10^{-4}$} & {\boldmath$7.1{\times}10^{-4}$} & {\boldmath$1.8{\times}10^{-4}$} \\
HSEvo    & {\boldmath$0.003$} & {\boldmath$0.003$} & {\boldmath$0.002$} & {\boldmath$0.020$} & {\boldmath$2.9{\times}10^{-4}$} & {\boldmath$5.6{\times}10^{-4}$} & {\boldmath$0.003$} & {\boldmath$0.009$} \\
MCTS-AHD & {\boldmath$0.005$} & {\boldmath$0.001$} & {\boldmath$0.001$} & {\boldmath$0.003$} & {\boldmath$2.4{\times}10^{-5}$} & {\boldmath$1.2{\times}10^{-5}$} & {\boldmath$9.5{\times}10^{-5}$} & {\boldmath$3.7{\times}10^{-4}$} \\
EoH-S    & $0.060$ & {\boldmath$0.004$} & {\boldmath$0.005$} & {\boldmath$0.016$} & {\boldmath$0.001$} & {\boldmath$6.2{\times}10^{-5}$} & {\boldmath$1.9{\times}10^{-5}$} & {\boldmath$1.4{\times}10^{-5}$} \\
CALM     & {\boldmath$4.0{\times}10^{-5}$} & {\boldmath$6.2{\times}10^{-6}$} & {\boldmath$1.3{\times}10^{-6}$} & {\boldmath$1.8{\times}10^{-5}$} & {\boldmath$4.2{\times}10^{-5}$} & {\boldmath$3.7{\times}10^{-4}$} & {\boldmath$3.3{\times}10^{-4}$} & {\boldmath$0.002$} \\
\bottomrule
\end{tabular}
}
\caption{Across-run significance on TSP. One-sided Welch $t$-test $p$-values for the EPOCH top-10 ensemble gap against each baseline, per test size and backbone; $p<0.05$ is shown in \textbf{bold}.}
\label{tab:sig-tsp}
\end{table*}

\begin{table}[t]
\centering
\footnotesize
\setlength{\tabcolsep}{12pt}
\renewcommand{\arraystretch}{1.0}
\resizebox{\columnwidth}{!}{
\begin{tabular}{l cc}
\toprule
Method & Qwen2.5-7B & Llama-3.1-8B \\
\midrule
ReEvo & {\boldmath$0.005$} & {\boldmath$0.009$} \\
HSEvo & {\boldmath$1.9{\times}10^{-11}$} & {\boldmath$0.023$} \\
EoH-S & {\boldmath$0.033$} & {\boldmath$0.007$} \\
CALM  & {\boldmath$0.013$} & {\boldmath$0.005$} \\
\bottomrule
\end{tabular}
}
\caption{Across-run significance on OBP. One-sided Welch $t$-test $p$-values for the EPOCH top-4 ensemble gap, averaged over the six Weibull test sets, against the baselines; $p<0.05$ is shown in \textbf{bold}.}
\label{tab:sig-obp}
\end{table}

\subsection{Cross-Instance Uncertainty Quantification}
The run-level test takes a run as its unit; we now ask whether the advantage holds across the individual test instances, that is, whether it is distributional rather than carried by a few easy cases. We restrict this analysis to TSP and KP, whose test sets contain $300$ and $1{,}000$ instances each; the OBP sets are too small for a per-instance comparison. For each method we average its per-instance optimality gap over runs to obtain one gap per instance for the deployed top-10 ensemble, and compare EPOCH against each baseline on the same instances. The paired per-instance gap differences are tested with the two-sided Wilcoxon signed-rank test~\cite{wilcoxon1945}, reported in Tables~\ref{tab:uq-tsp} and~\ref{tab:uq-kp} together with the number of instances on which EPOCH wins, ties, or loses against the baseline (a tie is a gap difference within $10^{-9}$). EPOCH wins the large majority of instances at every size and on both backbones on TSP, and on KP it wins several times more instances than it loses at every size, where most ties are instances both ensembles solve optimally; the test is significant throughout, so the gains in Tables~\ref{tab:tsp-full} and~\ref{tab:kp} reflect a population-wide advantage rather than a few outliers.

\begin{table*}[tp]
\centering
\scriptsize
\setlength{\tabcolsep}{12pt}
\renewcommand{\arraystretch}{1.1}
\resizebox{\textwidth}{!}{
\begin{tabular}{l c cccc cccc}
\toprule
& & \multicolumn{4}{c}{Qwen2.5-7B} & \multicolumn{4}{c}{Llama-3.1-8B} \\
\cmidrule(lr){3-6}\cmidrule(lr){7-10}
Method & $n$ & Wilcoxon $p$ & Win & Tie & Loss & Wilcoxon $p$ & Win & Tie & Loss \\
\midrule
\multirow{4}{*}{ReEvo}
 & 20  & $1.8{\times}10^{-47}$ & 278 & 0 & 22 & $7.9{\times}10^{-50}$ & 292 & 0 & 8 \\
 & 50  & $3.9{\times}10^{-50}$ & 289 & 0 & 11 & $6.1{\times}10^{-51}$ & 300 & 0 & 0 \\
 & 100 & $9.3{\times}10^{-51}$ & 296 & 0 & 4  & $6.1{\times}10^{-51}$ & 300 & 0 & 0 \\
 & 200 & $1.9{\times}10^{-50}$ & 289 & 0 & 11 & $6.1{\times}10^{-51}$ & 300 & 0 & 0 \\
\midrule
\multirow{4}{*}{HSEvo}
 & 20  & $3.7{\times}10^{-50}$ & 295 & 0 & 5 & $2.3{\times}10^{-50}$ & 292 & 0 & 8 \\
 & 50  & $6.1{\times}10^{-51}$ & 300 & 0 & 0 & $6.1{\times}10^{-51}$ & 300 & 0 & 0 \\
 & 100 & $6.1{\times}10^{-51}$ & 300 & 0 & 0 & $6.1{\times}10^{-51}$ & 300 & 0 & 0 \\
 & 200 & $6.1{\times}10^{-51}$ & 300 & 0 & 0 & $6.1{\times}10^{-51}$ & 300 & 0 & 0 \\
\midrule
\multirow{4}{*}{MCTS-AHD}
 & 20  & $1.4{\times}10^{-42}$ & 269 & 10 & 21 & $5.1{\times}10^{-45}$ & 275 & 0 & 25 \\
 & 50  & $1.9{\times}10^{-48}$ & 286 & 0 & 14 & $7.1{\times}10^{-51}$ & 298 & 0 & 2 \\
 & 100 & $7.9{\times}10^{-51}$ & 298 & 0 & 2  & $6.3{\times}10^{-51}$ & 298 & 0 & 2 \\
 & 200 & $7.7{\times}10^{-51}$ & 295 & 0 & 5  & $6.1{\times}10^{-51}$ & 300 & 0 & 0 \\
\midrule
\multirow{4}{*}{EoH-S}
 & 20  & $3.1{\times}10^{-11}$ & 196 & 10 & 94 & $3.0{\times}10^{-24}$ & 223 & 7 & 70 \\
 & 50  & $3.8{\times}10^{-32}$ & 246 & 0 & 54 & $5.2{\times}10^{-50}$ & 289 & 0 & 11 \\
 & 100 & $3.9{\times}10^{-46}$ & 279 & 0 & 21 & $9.5{\times}10^{-51}$ & 297 & 0 & 3 \\
 & 200 & $1.8{\times}10^{-48}$ & 283 & 0 & 17 & $7.1{\times}10^{-51}$ & 298 & 0 & 2 \\
\midrule
\multirow{4}{*}{CALM}
 & 20  & $2.5{\times}10^{-46}$ & 272 & 25 & 3 & $7.6{\times}10^{-49}$ & 277 & 17 & 6 \\
 & 50  & $8.0{\times}10^{-51}$ & 295 & 0 & 5 & $1.1{\times}10^{-50}$ & 296 & 1 & 3 \\
 & 100 & $7.1{\times}10^{-51}$ & 298 & 0 & 2 & $1.4{\times}10^{-50}$ & 294 & 0 & 6 \\
 & 200 & $1.4{\times}10^{-50}$ & 294 & 0 & 6 & $7.3{\times}10^{-51}$ & 296 & 0 & 4 \\
\bottomrule
\end{tabular}
}
\caption{Cross-instance uncertainty quantification on TSP for the deployed top-10 ensemble. Each entry is the Wilcoxon signed-rank $p$-value on the paired per-instance gap differences, with the Win/Tie/Loss counts of EPOCH against the baseline.}
\label{tab:uq-tsp}
\end{table*}

\begin{table}[t]
\centering
\footnotesize
\setlength{\tabcolsep}{7pt}
\renewcommand{\arraystretch}{1.05}
\resizebox{\columnwidth}{!}{
\begin{tabular}{l c c ccc}
\toprule
Method & $n$ & Wilcoxon $p$ & Win & Tie & Loss \\
\midrule
\multirow{3}{*}{ReEvo}
 & 50  & $2.3{\times}10^{-126}$ & 522 & 440 & 38 \\
 & 100 & $1.8{\times}10^{-121}$ & 657 & 273 & 70 \\
 & 200 & $4.4{\times}10^{-131}$ & 825 & 66 & 109 \\
\midrule
\multirow{3}{*}{MCTS-AHD}
 & 50  & $2.9{\times}10^{-77}$ & 453 & 493 & 54 \\
 & 100 & $3.6{\times}10^{-78}$ & 544 & 367 & 89 \\
 & 200 & $1.8{\times}10^{-36}$ & 515 & 309 & 176 \\
\bottomrule
\end{tabular}
}
\caption{Cross-instance uncertainty quantification on KP for the deployed top-10 ensemble with the Llama-3.1-8B backbone. Each entry is the Wilcoxon signed-rank $p$-value on the paired per-instance gap differences over the $1{,}000$ instances per size, with the Win/Tie/Loss counts of EPOCH against the baseline.}
\label{tab:uq-kp}
\end{table}

\section{Extended Ablations}
\label{app:ablation}
Each ablation removes exactly one component and replaces it with a simpler, untuned alternative, holding every other component and all hyperparameters at their full-method values. The full-method and \emph{w/o RSD} rows reuse the main EPOCH runs (the latter is the $\lambda{=}0$ setting), so all rows are directly comparable. Ablations use the Llama-3.1-8B backbone on TSP and OBP, the full training budget, and five runs per variant scored on $100$ instances per test set. Each row's removed component and its replacement are listed in Table~\ref{tab:ablation-defs}; the hyperparameter ablation of Section~\ref{sec:exp-ablation} is in Table~\ref{tab:ablation-sens}.

\begin{table*}[tp]
\centering
\small
\setlength{\tabcolsep}{15pt}
\setlength{\extrarowheight}{3pt}
\renewcommand{\arraystretch}{1.2}
\resizebox{\textwidth}{!}{
\begin{tabular}{p{0.205\textwidth} p{0.37\textwidth} p{0.35\textwidth}}
\toprule
Ablation & Removed & Replacement \\
\midrule
\multicolumn{3}{l}{\textit{Reflective learning}} \\
w/o RSD & the RSD distillation term & plain DPO on the same pairs ($\lambda{=}0$) \\
w/o cross-round pairs & the certified cross-round matching, a valid hinted Round-2 winner over a strictly worse valid Round-1 heuristic & each valid Round-2 heuristic paired with a randomly chosen valid Round-1 heuristic, reward-ordered ($\lambda{=}0$) \\
w/o instance-grounded insight & the gap-vector instance-split sentence in the teacher prompt & a generic critique from code and scalar gaps only \\
w/o teacher reflection & the insight channel & Round 2 regenerates unhinted ($\lambda{=}0$) \\
w/o preference optimization & all DPO bursts & a frozen generator, search and reflection only \\
\midrule
\multicolumn{3}{l}{\textit{Coverage-aware search}} \\
w/o expert-aware selection & expert-count ranking and the complementary second parent & parents ranked by average gap only \\
w/o adaptive operator allocation & the operator bandit & every eligible operator drawn with equal probability \\
coverage-blind bandit & the expert-coverage term of the bandit reward & a bandit rewarded by the parent-improvement and validity terms only \\
w/o coverage-stagnation signal & the coverage-stagnation criterion and soft cap & a fixed reset every $20$ iterations \\
w/o expert-preserving recovery & the expert-preserving survivor rule & reset to the single best heuristic by average gap \\
\midrule
\multicolumn{3}{l}{\textit{Gap-vector representation}} \\
scalar fitness (no gap vector) & every read of the per-instance gap vector & the scalar average gap throughout selection, the bandit reward, stall detection, rebuild, and archiving \\
\bottomrule
\end{tabular}
}
\caption{Component ablations: what each row removes and the simpler alternative that replaces it, grouped as in Table~\ref{tab:ablation}.}
\label{tab:ablation-defs}
\end{table*}

\subsection{Hyperparameter Ablation}
With the components fixed at the full method, we vary one hyperparameter at a time and report in Table~\ref{tab:ablation-sens} the same averaged ensemble gaps as the component study. The RSD weight $\lambda$ improves performance across a broad range of small values rather than at a single point, and the default $\lambda{=}0.1$ is best or tied on both problems; distilling only the certified cross-round winners outperforms distilling all pairs. The coverage-pressure blend $p$, which sets $p_{\mathrm{exp}}=p_{\mathrm{comp}}=p$, is best at the default $0.5$, since pure exploitation at $p{=}0$ removes expert-aware selection and pure coverage at $p{=}1$ over-explores. The preference-burst period $T_{\mathrm{dpo}}$ is best at its default of $20$ iterations; both shorter and longer periods are worse, and the longer period markedly destabilizes OBP.

\begin{table}[t]
\centering
\footnotesize
\setlength{\tabcolsep}{14pt}
\renewcommand{\arraystretch}{1.05}
\resizebox{\columnwidth}{!}{
\begin{tabular}{l cc}
\toprule
 & TSP & OBP \\
\midrule
\multicolumn{3}{l}{\textit{RSD weight} $\lambda$} \\
\quad$\lambda{=}0$ (no RSD)        & \val{7.17}{0.27} & \val{0.79}{0.06} \\
\quad$\lambda{=}0.1$~$\star$       & \valb{6.79}{0.19} & \valb{0.64}{0.07} \\
\quad$\lambda{=}0.15$              & \val{6.84}{0.35} & \val{0.73}{0.12} \\
\quad$\lambda{=}0.2$               & \val{6.96}{0.09} & \val{0.76}{0.09} \\
\midrule
\multicolumn{3}{l}{\textit{RSD scope}} \\
\quad{}cross-round only~$\star$      & \valb{6.79}{0.19} & \valb{0.64}{0.07} \\
\quad{}all pairs                     & \val{7.05}{0.05} & \val{0.73}{0.12} \\
\midrule
\multicolumn{3}{l}{\textit{Coverage pressure} $p$} \\
\quad$p{=}0$ (no expert selection) & \val{6.93}{0.15} & \val{0.81}{0.09} \\
\quad$p{=}0.5$~$\star$             & \valb{6.79}{0.19} & \valb{0.64}{0.07} \\
\quad$p{=}1$                       & \val{6.88}{0.18} & \val{0.78}{0.05} \\
\midrule
\multicolumn{3}{l}{\textit{Preference-burst period} $T_{\mathrm{dpo}}$} \\
\quad$T_{\mathrm{dpo}}{=}10$         & \val{7.05}{0.45} & \val{0.69}{0.11} \\
\quad$T_{\mathrm{dpo}}{=}20$~$\star$ & \valb{6.79}{0.19} & \valb{0.64}{0.07} \\
\quad$T_{\mathrm{dpo}}{=}40$         & \val{6.99}{0.14} & \val{1.00}{0.52} \\
\bottomrule
\end{tabular}
}
\caption{Hyperparameter ablation on the Llama-3.1-8B backbone, reported as the ensemble optimality gap (\%), lower is better; $\star$ marks the full-method default.}
\label{tab:ablation-sens}
\end{table}

\subsection{Teacher LLM Ablation}
\label{app:teacher}
The teacher is a critic rather than a generator: it reads two heuristics with their gaps and explains their difference in a few sentences. We fill this role with a reasoning model that reads code, as is now common for critic roles, and take it from a different model family than the generator, so that the critique is an independent view of the candidates rather than the generator's own. \texttt{GPT-5-nano} is a small model of this kind, which keeps a run's critiques at \$0.08 (Table~\ref{tab:cost}). To test whether the gains of EPOCH depend on the teacher model, we also run EPOCH with \texttt{GPT-4o-mini} and with the generator's own backbone as the teacher; in the latter, the frozen pretrained backbone writes the critique, so the run uses no external model. The teacher prompts of Appendix~\ref{app:prompts} are unchanged. Tables~\ref{tab:teacher-tsp} and~\ref{tab:teacher-obp} report the results under the protocol of Section~\ref{sec:experiments}; the EPOCH rows with \texttt{GPT-5-nano} and the EoH-S and CALM rows, the strongest baselines on these backbones at the ensemble level, are taken from Tables~\ref{tab:tsp-full} and~\ref{tab:obp-full}.

The teacher model has little effect on the ensemble. Across the three teachers, the top-10 TSP gap changes by less than $0.4$ points at every size and the top-4 OBP average by less than $0.2$ points, and every teacher keeps EPOCH's top-10 TSP ensemble and top-4 OBP average ahead of all baselines on the same backbone in Tables~\ref{tab:tsp-full} and~\ref{tab:obp-full}. With its own backbone as the teacher, EPOCH uses no model beyond its generator. Its top-10 gap at $n{=}50$ is still $36$--$37\%$ below CALM and $15$--$43\%$ below EoH-S, and its top-4 OBP average is $31$--$45\%$ below CALM and $28$--$34\%$ below EoH-S. \texttt{GPT-5-nano} gives the lowest ensemble gap in most settings, by a small margin. At top-1, where a single heuristic represents the run, the teachers differ more, and all three remain comparable to CALM while staying below every inference-only baseline on the same backbone, at each TSP size and in the OBP average. An independent teacher is therefore a design default rather than a dependency: the gains persist when the generator's own backbone writes the critique.

With \texttt{GPT-4o-mini} as the teacher, EPOCH on Llama-3.1-8B still outperforms the inference-only baselines run with \texttt{GPT-4o-mini} (Tables~\ref{tab:tsp-full} and~\ref{tab:obp-full}): on TSP at every metric and test size, with its top-10 ensemble $11$--$19\%$ below EoH-S and its top-1 heuristic $5$--$7\%$ below MCTS-AHD, and on OBP with a top-4 average $20\%$ below EoH-S and a top-1 average $7\%$ below. The same closed-source model serves EPOCH only as a critic, at $7$--$13\times$ lower API cost per run (Table~\ref{tab:cost}). With its own backbone as the teacher, EPOCH uses no external model and still outperforms the inference-only baselines run with \texttt{GPT-4o-mini} on TSP at every metric and test size on both backbones, with its top-10 ensemble $9$--$19\%$ below EoH-S and its top-1 heuristic $7$--$12\%$ below MCTS-AHD, and on OBP with Llama-3.1-8B at top-4, $13\%$ below EoH-S. Against the inference-only baselines run with \texttt{GPT-5-nano} (Table~\ref{tab:obp-full}), every teacher keeps EPOCH's top-4 OBP average lower on both backbones, and its top-1 average lower on Llama-3.1-8B.

\begin{table*}[tp]
\centering
\scriptsize
\setlength{\tabcolsep}{6.6mm}
\setlength{\extrarowheight}{0pt}
\renewcommand{\arraystretch}{0.95}
\resizebox{\textwidth}{!}{
\begin{tabular}{ll cccc}
\toprule
Method & & TSP-20 & \underline{TSP-50} & TSP-100 & TSP-200 \\
\midrule
\multicolumn{6}{c}{\textit{LLM-based AHD: Qwen2.5-7B}} \\
\multirow{3}{*}{EoH-S}    & Top-1  & \val{8.64}{1.03} & \val{12.25}{1.26} & \val{14.44}{1.84} & \val{16.73}{2.50} \\
                          & Top-3  & \val{5.43}{1.14} & \val{9.12}{1.05} & \val{11.76}{1.45} & \val{14.43}{2.37} \\
                          & Top-10 & \val{3.46}{0.66} & \val{7.08}{0.76} & \val{9.97}{1.01} & \val{12.69}{1.34} \\
\rowsep{6}
\multirow{3}{*}{CALM}     & Top-1  & \val{6.75}{0.08} & \val{9.99}{0.22} & \val{11.96}{0.20} & \val{13.96}{0.20} \\
                          & Top-3  & \val{6.70}{0.14} & \val{9.92}{0.23} & \val{11.90}{0.21} & \val{13.91}{0.27} \\
                          & Top-10 & \val{6.30}{0.50} & \val{9.59}{0.55} & \val{11.59}{0.42} & \val{13.77}{0.34} \\
\rowsep{6}
\multirow{3}{*}{\shortstack[l]{\textbf{EPOCH}\\(GPT-5-nano)}} & Top-1  & \vals{6.42}{0.08} & \val{9.81}{0.22} & \val{11.70}{0.46} & \val{13.67}{0.91} \\
                                    & Top-3  & \vals{4.39}{0.20} & \vals{7.62}{0.44} & \val{10.06}{0.45} & \val{12.25}{0.50} \\
                                    & Top-10 & \vals{2.91}{0.14} & \valsb{5.71}{0.31} & \vals{8.22}{0.37}  & \val{10.93}{0.47} \\
\rowsep{6}
\multirow{3}{*}{\shortstack[l]{\textbf{EPOCH}\\(GPT-4o-mini)}} & Top-1  & \val{6.46}{0.30} & \val{9.75}{0.25} & \val{11.75}{0.30} & \val{13.41}{0.90} \\
                                    & Top-3  & \val{4.73}{0.42} & \val{7.98}{0.48} & \val{10.29}{0.59} & \val{12.39}{0.60} \\
                                    & Top-10 & \val{3.15}{0.20} & \val{5.99}{0.27} & \val{8.30}{0.15}  & \vals{10.66}{0.22} \\
\rowsep{6}
\multirow{3}{*}{\shortstack[l]{\textbf{EPOCH}\\(own backbone)}} & Top-1  & \val{6.49}{0.46} & \vals{9.73}{0.34} & \vals{11.61}{0.46} & \vals{13.31}{0.59} \\
                                    & Top-3  & \val{4.58}{0.79} & \val{7.69}{0.32} & \vals{9.71}{0.17}  & \vals{11.83}{0.15} \\
                                    & Top-10 & \val{2.92}{0.26} & \val{6.04}{0.11} & \val{8.42}{0.14}  & \val{10.86}{0.29} \\
\midrule
\multicolumn{6}{c}{\textit{LLM-based AHD: Llama-3.1-8B}} \\
\multirow{3}{*}{EoH-S}    & Top-1  & \val{10.69}{0.93} & \val{15.43}{1.58} & \val{17.86}{1.38} & \val{20.04}{1.46} \\
                          & Top-3  & \val{6.05}{0.72} & \val{11.68}{1.06} & \val{14.59}{0.93} & \val{17.31}{1.05} \\
                          & Top-10 & \val{4.70}{0.81} & \val{10.46}{1.02} & \val{13.60}{0.96} & \val{16.45}{1.07} \\
\rowsep{6}
\multirow{3}{*}{CALM}     & Top-1  & \val{6.76}{0.26} & \val{10.02}{0.26} & \val{11.75}{0.27} & \vals{13.47}{0.31} \\
                          & Top-3  & \val{6.55}{0.23} & \val{9.68}{0.55}  & \val{11.34}{0.51} & \val{13.33}{0.47} \\
                          & Top-10 & \val{6.19}{0.54} & \val{9.31}{0.80}  & \val{11.04}{0.67} & \val{13.30}{0.91} \\
\rowsep{6}
\multirow{3}{*}{\shortstack[l]{\textbf{EPOCH}\\(GPT-5-nano)}} & Top-1  & \vals{6.19}{0.30} & \vals{9.61}{0.25} & \vals{11.63}{0.21} & \val{13.57}{0.36} \\
                                    & Top-3  & \vals{4.31}{0.30} & \vals{7.62}{0.23} & \vals{9.85}{0.36}  & \vals{11.98}{0.49} \\
                                    & Top-10 & \valsb{2.81}{0.22} & \vals{5.73}{0.15}  & \valsb{8.15}{0.19}  & \val{10.74}{0.38} \\
\rowsep{6}
\multirow{3}{*}{\shortstack[l]{\textbf{EPOCH}\\(GPT-4o-mini)}} & Top-1  & \val{6.92}{0.40} & \val{10.07}{0.21} & \val{11.97}{0.26} & \val{13.94}{0.35} \\
                                    & Top-3  & \val{4.60}{0.38} & \val{7.69}{0.35} & \val{9.97}{0.32}  & \val{12.10}{0.26} \\
                                    & Top-10 & \val{2.87}{0.14} & \val{5.89}{0.36} & \val{8.36}{0.34}  & \val{10.90}{0.28} \\
\rowsep{6}
\multirow{3}{*}{\shortstack[l]{\textbf{EPOCH}\\(own backbone)}} & Top-1  & \val{6.43}{0.35} & \val{9.84}{0.51} & \vals{11.63}{0.63} & \val{13.69}{0.82} \\
                                    & Top-3  & \val{4.45}{0.14} & \val{8.01}{0.62} & \val{10.17}{0.86} & \val{12.37}{0.87} \\
                                    & Top-10 & \val{2.89}{0.19} & \val{5.92}{0.37} & \val{8.22}{0.45}  & \valsb{10.60}{0.56} \\
\bottomrule
\end{tabular}
}
\caption{Teacher LLM ablation on TSP. Optimality gap (\%) to LKH-3 optima for EPOCH with three teachers, named in parentheses, with the EoH-S and CALM rows of Table~\ref{tab:tsp-full} for reference; each cell is the mean over runs with standard deviation, lower is better. Size $n{=}50$ is ID (underlined); $n\in\{20,100,200\}$ are OOD. Within each backbone block the best method per metric and test set is \colorbox{shadegray}{shaded}; each test set's overall best result is shown in \textbf{bold}.}
\label{tab:teacher-tsp}
\end{table*}

\begin{table*}[tp]
\centering
\scriptsize
\setlength{\tabcolsep}{6pt}
\setlength{\extrarowheight}{0pt}
\renewcommand{\arraystretch}{0.95}
\resizebox{\textwidth}{!}{
\begin{tabular}{ll cccccc|c}
\toprule
Method & & \underline{1k\_100} & \underline{1k\_500} & \underline{5k\_100} & \underline{5k\_500} & 10k\_100 & 10k\_500 & Avg. \\
\midrule
\multicolumn{8}{c|}{\textit{LLM-based AHD: Qwen2.5-7B}} & \multicolumn{1}{c}{} \\
\multirow{2}{*}{EoH-S}    & Top-1 & \val{3.70}{1.10} & \val{0.49}{0.29} & \val{2.89}{1.21} & \val{0.52}{0.12} & \val{2.71}{1.21} & \val{0.44}{0.12} & \val{1.79}{0.57} \\
                          & Top-4 & \val{3.37}{1.10} & \val{0.25}{0.25} & \val{2.31}{1.35} & \val{0.39}{0.12} & \val{2.04}{1.43} & \val{0.39}{0.12} & \val{1.46}{0.64} \\
\rowsep{9}
\multirow{2}{*}{CALM}     & Top-1 & \val{4.17}{0.54} & \val{0.35}{0.12} & \val{2.89}{1.18} & \val{0.45}{0.15} & \val{2.55}{1.33} & \val{0.35}{0.15} & \val{1.79}{0.51} \\
                          & Top-4 & \val{4.09}{0.55} & \val{0.30}{0.10} & \val{2.82}{1.28} & \val{0.45}{0.15} & \val{2.50}{1.41} & \val{0.35}{0.16} & \val{1.75}{0.56} \\
\rowsep{9}
\multirow{2}{*}{\textbf{EPOCH (GPT-5-nano)}} & Top-1 & \val{3.91}{0.66} & \val{0.25}{0.22} & \val{1.54}{0.46} & \val{0.17}{0.09} & \val{1.27}{0.66} & \val{0.19}{0.14} & \val{1.22}{0.20} \\
                          & Top-4 & \val{3.19}{0.10} & \val{0.15}{0.20} & \val{0.79}{0.03} & \val{0.17}{0.09} & \valb{0.30}{0.02} & \val{0.15}{0.07} & \vals{0.79}{0.05} \\
\rowsep{9}
\multirow{2}{*}{\textbf{EPOCH (GPT-4o-mini)}} & Top-1 & \val{3.94}{0.71} & \val{1.32}{1.19} & \val{2.08}{1.49} & \val{0.53}{0.06} & \val{1.75}{1.59} & \val{0.42}{0.07} & \val{1.67}{0.50} \\
                          & Top-4 & \val{3.22}{0.15} & \val{0.33}{0.31} & \val{0.78}{0.03} & \val{0.36}{0.13} & \val{0.36}{0.08} & \val{0.27}{0.15} & \val{0.89}{0.05} \\
\rowsep{9}
\multirow{2}{*}{\textbf{EPOCH (own backbone)}} & Top-1 & \val{3.26}{0.33} & \val{0.58}{0.23} & \val{0.96}{0.17} & \val{0.60}{0.04} & \val{1.27}{1.24} & \val{0.52}{0.05} & \vals{1.20}{0.13} \\
                          & Top-4 & \val{3.18}{0.28} & \val{0.16}{0.12} & \val{0.82}{0.04} & \val{0.28}{0.05} & \val{1.22}{1.30} & \val{0.18}{0.06} & \val{0.97}{0.18} \\
\midrule
\multicolumn{8}{c|}{\textit{LLM-based AHD: Llama-3.1-8B}} & \multicolumn{1}{c}{} \\
\multirow{2}{*}{EoH-S}    & Top-1 & \val{3.92}{0.78} & \val{0.21}{0.09} & \val{2.89}{0.87} & \val{0.51}{0.07} & \val{2.59}{1.00} & \val{0.45}{0.05} & \val{1.76}{0.38} \\
                          & Top-4 & \val{3.56}{0.73} & \valb{0.08}{0.12} & \val{1.46}{0.57} & \val{0.16}{0.09} & \val{1.14}{0.69} & \val{0.14}{0.10} & \val{1.09}{0.27} \\
\rowsep{9}
\multirow{2}{*}{CALM}     & Top-1 & \val{3.71}{0.52} & \val{0.59}{0.82} & \val{1.32}{0.40} & \val{0.30}{0.14} & \val{0.82}{0.41} & \val{0.27}{0.14} & \val{1.17}{0.22} \\
                          & Top-4 & \val{3.69}{0.51} & \val{0.59}{0.82} & \val{1.29}{0.43} & \val{0.29}{0.15} & \val{0.80}{0.43} & \val{0.26}{0.13} & \val{1.15}{0.23} \\
\rowsep{9}
\multirow{2}{*}{\textbf{EPOCH (GPT-5-nano)}} & Top-1 & \val{2.99}{0.70} & \val{0.30}{0.36} & \val{1.66}{0.90} & \val{0.32}{0.21} & \val{1.33}{0.97} & \val{0.28}{0.24} & \val{1.15}{0.39} \\
                          & Top-4 & \valb{2.34}{0.43} & \val{0.10}{0.12} & \valb{0.74}{0.06} & \val{0.20}{0.07} & \val{0.36}{0.06} & \val{0.13}{0.04} & \valsb{0.64}{0.07} \\
\rowsep{9}
\multirow{2}{*}{\textbf{EPOCH (GPT-4o-mini)}} & Top-1 & \val{3.20}{0.63} & \val{0.55}{0.43} & \val{1.11}{0.20} & \val{0.30}{0.19} & \val{0.64}{0.23} & \val{0.58}{0.65} & \vals{1.06}{0.15} \\
                          & Top-4 & \val{2.86}{0.52} & \val{0.15}{0.20} & \val{0.80}{0.08} & \valb{0.11}{0.07} & \val{0.35}{0.03} & \valb{0.10}{0.05} & \val{0.73}{0.06} \\
\rowsep{9}
\multirow{2}{*}{\textbf{EPOCH (own backbone)}} & Top-1 & \val{3.65}{0.54} & \val{0.45}{0.29} & \val{1.31}{0.54} & \val{0.38}{0.16} & \val{0.87}{0.56} & \val{0.33}{0.18} & \val{1.16}{0.19} \\
                          & Top-4 & \val{3.04}{0.60} & \val{0.25}{0.16} & \val{0.79}{0.04} & \val{0.21}{0.12} & \val{0.32}{0.02} & \val{0.15}{0.05} & \val{0.79}{0.10} \\
\bottomrule
\end{tabular}
}
\caption{Teacher LLM ablation on OBP. Optimality gap (\%) to the $L_1$ lower bound on the six held-out Weibull test sets for EPOCH with three teachers, named in parentheses, with the EoH-S and CALM rows of Table~\ref{tab:obp-full} for reference; each cell is the mean over runs with standard deviation, lower is better. Columns are $(\text{items},\text{capacity})$ and \texttt{Avg.}\ is the mean over the six. Item scales $1\mathrm{k},5\mathrm{k}$ are ID (underlined); $10\mathrm{k}$ is OOD. Within each backbone block the best \texttt{Avg.}\ per metric is \colorbox{shadegray}{shaded}; each test set's overall best result is shown in \textbf{bold}.}
\label{tab:teacher-obp}
\end{table*}

\section{Implementation Details and Hyperparameters}
\label{app:impl}

\paragraph{Backbones and training.}
We use \texttt{Qwen2.5-7B-Instruct}~\cite{qwen2025qwen25} and \texttt{Meta-Llama-3.1-8B-Instruct}~\cite{grattafiori2024llama3} as policy backbones, loaded in 4-bit and fine-tuned with LoRA (rank $32$)~\cite{hu2022lora}; the DPO reference policy is the LoRA-disabled pass of the current model. The teacher is an external LLM (\texttt{GPT-5-nano} in our main runs; Appendix~\ref{app:teacher} compares other teachers) queried once per iteration. EPOCH runs $125$ iterations with $4{+}4$ generations per iteration (Round~1 ${+}$ Round~2), a budget of $1{,}000$ heuristic generations, sampling at temperature $0.8$ and top-$p\,0.95$. A preference-optimization burst runs every $T_{\mathrm{dpo}}{=}20$ iterations, optimizing the DPO loss of Eq.~\eqref{eq:dpo} with $\beta{=}0.1$ and learning rate $5\times10^{-6}$. The RSD term of Eq.~\eqref{eq:rsd} is applied to the cross-round pairs with weight $\lambda{=}0.1$ and length normalization; $\lambda{=}0$ reduces the burst to plain DPO and isolates the contribution of RSD as a single-coefficient ablation.

\paragraph{Computational cost.}
Every method consumes the same budget of $1{,}000$ heuristic generations, and the teacher LLM adds only a single \texttt{GPT-5-nano} call per iteration, at most $125$ per run. The generations within each phase are independent rollouts and are sampled in parallel, in contrast to the sequential tree expansion of MCTS-AHD. All runs used a workstation with two NVIDIA RTX 6000 Ada GPUs (48\,GB each).

\paragraph{Auxiliary LLM calls and API cost.}
\label{app:cost}
An \emph{auxiliary LLM call} returns text that is not itself a candidate heuristic, such as a reflection on evaluated heuristics or a description of a generated one; the teacher critique of EPOCH is one. Auxiliary calls are not counted in the budget of $1{,}000$ heuristic generations. For a fair comparison, where an open-backbone baseline makes auxiliary calls, they are served by \texttt{GPT-5-nano}, the teacher model of EPOCH. Table~\ref{tab:cost} reports the API cost per run at list prices. The only API calls of EPOCH are the teacher's, which cost \$0.08 per run with \texttt{GPT-5-nano}, at most \$0.05 with \texttt{GPT-4o-mini}, and nothing with the generator's own backbone as the teacher (Appendix~\ref{app:teacher}), the lowest API cost among the open-backbone methods that make auxiliary calls. An inference-only baseline run on a closed-source model costs $4$--$8\times$ as much with \texttt{GPT-4o-mini} and $14$--$29\times$ as much with \texttt{GPT-5-nano}, as each of its $1{,}000$ heuristic generations is an API call.

\begin{table}[t]
\centering
\footnotesize
\setlength{\tabcolsep}{14pt}
\renewcommand{\arraystretch}{1.05}
\resizebox{\columnwidth}{!}{
\begin{tabular}{l cc}
\toprule
 & TSP & OBP \\
\midrule
\multicolumn{3}{l}{\textit{LLM-based AHD: GPT-4o-mini (closed-source)}} \\
\quad{}ReEvo    & 0.60 & 0.52 \\
\quad{}HSEvo    & 0.38 & 0.29 \\
\quad{}MCTS-AHD & 0.42 & 0.36 \\
\quad{}EoH-S    & 0.53 & 0.29 \\
\midrule
\multicolumn{3}{l}{\textit{LLM-based AHD: GPT-5-nano (closed-source)}} \\
\quad{}ReEvo    & 2.2 & 1.7 \\
\quad{}HSEvo    & 2.3 & 1.3 \\
\quad{}MCTS-AHD & 2.1 & 1.5 \\
\quad{}EoH-S    & 1.9 & 1.1 \\
\midrule
\multicolumn{3}{l}{\textit{LLM-based AHD: Qwen2.5-7B / Llama-3.1-8B}} \\
\quad{}ReEvo    & 0.50 & 0.50 \\
\quad{}HSEvo    & 0.12 & 0.12 \\
\quad{}MCTS-AHD & 0.43 & 0.43 \\
\quad{}\textbf{EPOCH} (GPT-5-nano teacher)   & 0.08 & 0.08 \\
\quad{}\textbf{EPOCH} (GPT-4o-mini teacher)  & 0.05 & 0.04 \\
\quad{}\textbf{EPOCH} (own-backbone teacher) & 0.00 & 0.00 \\
\bottomrule
\end{tabular}
}
\caption{API cost per run in US dollars at list prices, grouped by the model that generates the heuristics. With an open backbone, the cost is that of the auxiliary LLM calls, averaged over the two backbones.}
\label{tab:cost}
\end{table}

\paragraph{Hyperparameters.}
Every selection, bandit, and recovery hyperparameter is stated at its first use in Section~\ref{sec:method} and consolidated in Table~\ref{tab:hparams}. Invalid responses carry only a failure-type label, namely no idea, no code block, no parseable function, or runtime failure; the label selects the repair and diagnostic critiques of Section~\ref{sec:method-reflection} and marks the valid-over-invalid pairs, and no numeric score is attached to invalid responses. Training uses standard preference-optimization tooling.\footnote{Implementation built on the Unsloth and TRL libraries.}

\begin{table*}[tp]
\centering
\small
\setlength{\tabcolsep}{12pt}
\resizebox{0.6\textwidth}{!}{
\begin{tabular}{ll}
\toprule
Component & Setting \\
\midrule
Iterations / heuristic generations & $125$ / $1{,}000$ \\
Generations per iteration & $4 + 4$ (Round 1 $+$ Round 2) \\
Sampling & temperature $0.8$, top-$p$ $0.95$ \\
Population size $N_p$ & $10$ \\
LoRA rank & $32$ (4-bit base) \\
DPO every / $\beta$ / lr & $20$ iterations / $0.1$ / $5\times10^{-6}$ \\
DPO filter & $30$th percentile of rewards since the previous burst \\
Buffer & accumulating (unbounded) \\
RSD $\lambda$ (Eq.~\eqref{eq:rsd}) & $0.1$, length-norm., reflective pairs \\
$p_{\mathrm{exp}}$, $p_{\mathrm{comp}}$, $\tau$ & $0.5$, $0.5$, $0.5$ \\
Bandit priors $w^0$ & $3$ (crossover), $1$ (others) \\
Bandit $\phi,\eta,\kappa,(w_e,w_p,w_v)$ & $0.05$, $0.3$, $6$, $(0.6,0.25,0.15)$ \\
Recovery $T_\rho$, $\epsilon_\rho$, $T_{\mathrm{cap}}$ & $10$, $10^{-4}$, $40$ \\
Ensemble save $K$ & $10$ (TSP, KP); $4$ (OBP, all experts) \\
\bottomrule
\end{tabular}
}
\caption{EPOCH hyperparameters (shared across problems and backbones unless noted).}
\label{tab:hparams}
\end{table*}

\section{Prompt Templates}
\label{app:prompts}

Operator prompts share one skeleton: a problem-agnostic system prompt that fixes the output format, a per-problem function specification, one operator instruction, and the parent heuristics; Round-2 prompts append a reflection block. The six operator instructions are listed in Table~\ref{tab:operators} and the per-problem function descriptions in Table~\ref{tab:funcspec}, while the signatures appear in Appendix~\ref{app:problems}. In the boxes below, braces mark fields populated at run time, and lines beginning with \texttt{\#} are explanatory comments rather than prompt text. The function-specification strings follow the conventions of~\citet{ye2024reevo}, and the system prompt and operator instructions adapt those of~\citet{huang2025calmcoevolutionalgorithmslanguage}, with the component memory of injection removed, and complementary crossover and refine added as new operators.

\begin{table*}[!t]
\centering\small
\setlength{\tabcolsep}{10pt}
\resizebox{\textwidth}{!}{
\begin{tabular}{@{}p{0.19\textwidth}p{0.73\textwidth}@{}}
\toprule
Operator & Instruction sentence \\
\midrule
Exploitative crossover & Please generate a new algorithm that is motivated by the following algorithms but performs better on any same instance. \\
\addlinespace
Complementary crossover & The two algorithms below each perform well on different problem instances. Please generate a new algorithm that unites their strengths so that it performs well across all the instances that either of them already handles well. \\
\addlinespace
Simplification & Please create a simplified and more elegant version of an algorithm by distilling and refining the core ideas from the following. \\
\addlinespace
Injection & Inject a novel, meaningful component into the following algorithm. The component may be self-devised or inspired by ideas from other domains or problems. \\
\addlinespace
Replacement & For the following algorithm, identify a fragment that assigns equal or near-equal credits to multiple elements and rewrite it to a fragment where credits are deterministically and reasonably differentiated. \\
\addlinespace
Refine & Identify the single most important weakness or failure mode of the following algorithm, then write an improved version that fixes specifically that weakness while preserving the parts that already work well. \\
\bottomrule
\end{tabular}
}
\caption{Operator instruction sentences. Each is placed before the parent heuristics and fills the \texttt{\{operator\_instruction\}} field of the Round-1 user prompt.}
\label{tab:operators}
\end{table*}

\begin{table*}[!t]
\centering\small
\setlength{\tabcolsep}{10pt}
\resizebox{\textwidth}{!}{
\begin{tabular}{@{}p{0.1\textwidth}p{0.82\textwidth}@{}}
\toprule
Problem & Description and efficiency note \\
\midrule
TSP & Takes the current node, the destination node, the unordered set of unvisited nodes, and a distance matrix where \texttt{distance\_matrix[i][j]} is the distance from node $i$ to $j$; returns the next node, which must be unvisited. Efficiency: called once per step, so avoid nested loops and recomputation. \\
\addlinespace
OBP & Takes the next item's size and an array of remaining bin capacities; returns a priority score per bin, and the highest-priority bin is chosen. \\
\addlinespace
KP & Takes the remaining capacity and two aligned arrays with the weights and values of the items not yet packed; returns the index of the next item to pack, which must fit. Efficiency: called once per packed item, so score all remaining items with array operations rather than loops over items. \\
\bottomrule
\end{tabular}
}
\caption{Per-problem function description used in the operator prompts; it fills the \texttt{\{function\_spec\}} field together with the signature from Appendix~\ref{app:problems}.}
\label{tab:funcspec}
\end{table*}

\begin{table*}[!t]
\centering\small
\setlength{\tabcolsep}{10pt}
\resizebox{\textwidth}{!}{
\begin{tabular}{@{}p{0.1\textwidth}p{0.38\textwidth}p{0.42\textwidth}@{}}
\toprule
Critique & \texttt{\{candidate\_intro\}} & \texttt{\{task\_direction\}} \\
\midrule
Contrastive & Below are the best and worst valid algorithms it produced for that request, ordered from best to worst: & Identify which design choices appear to have helped reduce the optimality gap, and give a concrete change of direction that would lead to an even better next attempt. \\
\addlinespace
Repair & Below are two algorithms it produced for that request: one valid algorithm that ran successfully, and one invalid algorithm that failed to run (its ``Why it failed'' line explains the reason). & One candidate ran to completion and one failed at runtime. Identify the design choices that supported successful execution and the code-level pitfalls that caused the failure, then give a concrete next-attempt direction that combines both lessons. \\
\addlinespace
Diagnostic & Below are two algorithms it produced for that request. Both of them are invalid - neither ran successfully. Each one's ``Why it failed'' line explains why it was rejected. & Both candidates failed to run, so there is no working example: diagnose the shared failure modes and give a concrete next-attempt direction that produces a runnable function. \\
\bottomrule
\end{tabular}
}
\caption{Critique-mode fields of the teacher user prompt. Each row fills \texttt{\{candidate\_intro\}} and \texttt{\{task\_direction\}}; the contrastive critique additionally receives the \texttt{\{instance\_split\}} sentence.}
\label{tab:critique-modes}
\end{table*}

\paragraph{System prompt.}
The system prompt is shared by all operators and both rounds.

\begin{promptbox}{System prompt}
\scriptsize
\begin{verbatim}
# {problem_name} and {problem_description}:
# the task name and one-line description.
Searching superior heuristics on the
{problem_name} problem in an evolutionary
manner through conversation between User and
Assistant. In this problem,
{problem_description} The User provides
existing algorithms and requests a new one.

## Your Task
You should first present a concise conceptual
description, followed by a complete code
implementation.

* The description must:
    * Be enclosed with a double brace {{...}}
      on plain text lines and start with "The
      idea of the algorithm is to".
    * Appear BEFORE the Python code block;
      never put the code block first.
    * Never be wrapped inside a ```python```
      fence - it is plain text, not code.
    * Be self-contained, insightful, and
      creatively original.
    * Not reference or rely on any prior ideas
      or existing code.
* The code must:
    * Conform exactly to the function
      signature given in the user message -
      same name, parameters, and return type.
    * Be a single Python function in a Python
      code block.
    * Exclude any usage examples.
    * Be deterministic; do not use random
      number generation such as np.random or
      the random module.
    * Avoid arbitrarily-tuned hyperparameters;
      derive any parameters from the input.
    * Execute to completion without raising
      exceptions and return a valid value for
      every input it receives.

Overall, your response should be like:
{{The idea of the algorithm is to
(specific description)}}
```python
your code here
```
Except for the idea and code, do not give
additional explanations or comments.
\end{verbatim}
\end{promptbox}

\paragraph{User prompts.}
The Round-1 user prompt concatenates the function specification of Table~\ref{tab:funcspec}, one operator instruction from Table~\ref{tab:operators}, and the parent block; the Round-2 prompt is the same text with the teacher's insight appended.

\begin{promptbox}{User prompt -- Round 1}
\scriptsize
\begin{verbatim}
## Function Specification
{function_spec}
# per-problem signature + description;
# see the function-spec table and App. B.

{operator_instruction}
# one operator's instruction; see the
# operator table.

{parent_block}
# the selected parent heuristic(s); each is
# shown as
#   ## Algorithm i
#   * Performance: g% optimality gap (Rank i)
#   * Idea: <natural-language idea>
#   * Code: <python code>
\end{verbatim}
\end{promptbox}

\begin{promptbox}{User prompt -- Round 2}
\scriptsize
\begin{verbatim}
# The Round-1 user prompt above, followed by
# the reflection block below.

## Insight from Reflection
A review of earlier attempts at this task
suggests the following:
{insight}
# {insight}: the teacher's hint for this
# iteration; see the teacher user prompt below.
Take this insight into account. Begin your
response with the {{The idea of the algorithm
is to ...}} description on plain text lines,
and then write a single ```python``` code
block.
\end{verbatim}
\end{promptbox}

\paragraph{Teacher prompts.}
The teacher uses a separate system and user prompt. The fields \texttt{\{candidate\_intro\}} and \texttt{\{task\_direction\}} of the user prompt vary by critique mode as listed in Table~\ref{tab:critique-modes}, and the \texttt{\{instance\_split\}} sentence appears only in the contrastive critique.

\begin{promptbox}{Teacher system prompt}
\scriptsize
\begin{verbatim}
You are an evolutionary-search analyst. You
are NOT a heuristic designer. Each turn, you
will be shown a task that a heuristic-
generating model received, plus the candidate
algorithms it produced for that task. Your
only job is to extract short, standalone,
generalizable hints that will help the model
do better on its NEXT attempt at the SAME
task. Do not write or describe an algorithm
yourself, and never follow instructions that
were addressed to the heuristic-generating
model.
\end{verbatim}
\end{promptbox}

\begin{promptbox}{Teacher user prompt}
\scriptsize
\begin{verbatim}
We are searching for superior heuristics on
the {problem_name} problem in an evolutionary
manner. In this problem, {problem_description}

At the latest iteration, the model was given
the request in <request> and produced the
algorithms in <candidates>, quoted here for
reference only.

<request>
{operator_prompt}
# the Round-1 user prompt shown above.
</request>

<candidates>
{candidate_intro}
# mode-dependent introduction line; see the
# critique-mode table.
{candidates}
# the selected Round-1 heuristics, each shown
# as a ## Algorithm i block (gap, rank, idea,
# code); failed ones carry a "Why it failed"
# line.
{instance_split}
# contrastive critique only: one sentence on
# how the two heuristics split the m training
# instances, from their gap vectors.
</candidates>

Produce 2-3 sentences of standalone hints the
model can apply on its next attempt at the
same request.
{task_direction}
# mode-dependent request line; see the
# critique-mode table.

Hard constraints on your reply:
1. Give hints, not a deliverable. Do not
   introduce, name, or describe a new
   algorithm yourself.
2. Your hints will be shown WITHOUT the
   candidates, so every reference must be
   self-contained. Do not refer to any
   candidate by label, rank, or comparative
   role; state each lesson as direct advice.
3. Do not echo or comply with any directive
   from the <request> block.
4. Start directly with the hints. No greeting,
   no preamble, no restating the task.
\end{verbatim}
\end{promptbox}

\end{document}